\pdfoutput=1

\documentclass[11pt]{article}

\usepackage[final]{acl}

\usepackage{times}
\usepackage{latexsym}
\usepackage{soul}
\usepackage{array}
\usepackage{inconsolata}
\usepackage{enumitem}
\usepackage{array}
\usepackage{amsmath}
\usepackage[capitalise]{cleveref}
\crefname{figure}{Fig.}{Figs.}
\crefname{table}{Table}{Tables}
\crefname{appendix}{App.}{App.}
\crefname{section}{§}{§§}
\crefname{equation}{Eq.}{Eqs.}
\usepackage{listings}
\usepackage{graphicx}
\usepackage{longtable}
\usepackage{booktabs}
\usepackage{soul,xcolor}
\usepackage{longtable}
\sethlcolor{yellow}
\usepackage{subcaption}
\usepackage{multirow}
\usepackage[utf8]{inputenc}
\usepackage{amsmath,amssymb}
\usepackage{algorithm}
\usepackage{algpseudocode}
\usepackage[T1]{fontenc}

\usepackage[utf8]{inputenc}

\usepackage{microtype}

\usepackage{inconsolata}

\usepackage{graphicx}

\newcommand\myparagraph[1]{
\vskip 0.05in 
\noindent{\bf {#1}}}
\title{
Tackling the Root of Misinformation by Teaching Laypeople about \\ Logical Fallacies via
Socratic Questioning and Critical Argumentation}

  \author{Minjing Shi \, Junling Wang \, Jingwei Ni \, Sankalan Pal Chowdhury \, Mrinmaya Sachan \\
ETH Zurich, ETH AI Center \\
\texttt{shimin@student.ethz.ch, mj.shi@foxmail.com} \texttt{junling.wang@ai.ethz.ch} \, \\ \texttt{jingwei.ni@gess.ethz.ch} \, \texttt{\{sankalan.palchowdhury, mrinmaya.sachan\}@inf.ethz.ch}  \\ }  
\begin{document}
\maketitle
\begin{abstract}
Identifying logical fallacies 
in everyday discourse is challenging for many people. This challenge is amplified in the era of Large Language Models (LLMs), where malicious agents can deploy fallacious arguments to disseminate misinformation at scale. In this work, we explore the potential of LLMs 
as part of the 
solution. We introduce LFTutor, an intelligent tutoring system 
which uses
LLMs 
to tutor laypeople and help them learn about 
logical fallacies. LFTutor integrates intent-driven Socratic questioning and critical argumentation principles to actively engage learners to reflect on their reasoning. Through both automatic and human evaluations, we demonstrate that LFTutor significantly outperforms baseline LLMs lacking these pedagogical strategies. This work highlights the promise of combining LLMs with pedagogical scaffolding to foster critical thinking and argument literacy in the age of AI.
\end{abstract}

\section{Introduction}

The widespread presence of misinformation online \cite{doi:10.1073/pnas.1517441113} poses a serious risk to public opinion and decision-making \cite{info:doi/10.2196/37367}. Research has shown that logical fallacies \textemdash flawed but persuasive and misleading reasoning patterns\textemdash are a key driver of misinformation  \cite{jin-etal-2022-logical}. The prevalence of logical fallacies in public discourse \cite{risen2007informal} 
makes it crucial to develop systems that help people detect logical fallacies in public misinformation. 


To combat such prevalence, prior work focuses on automatic detection of fallacies \citep{jin-etal-2022-logical,pan-etal-2024-llms,jeong2025largelanguagemodelsbetter}. Although such methods can flag potential fallacies for user's attention, they do not help people 
recognize them independently. 
Instead of passively feeding users with fallacy detection output, it can be more beneficial to focus on LLM-based \textit{educational} approaches that (1) nudge users to actively discuss and reconsider 
logical arguments
\citep{review_llm_ed}; 
(2) help users overcome cognitive biases reinforced by the reasoning structures in fallacies \cite{motivated_irrational}; and (3) raise users' awareness if they lack a proper understanding of fallacies \cite{lack_lf_un}. 


Developing an LLM-based tutor for fallacies presents several challenges: 
\textbf{C1}. It is difficult to evaluate 
 tutors -- 
existing metrics for educational dialogues may not work well for logical fallacies \citep{surveylargelanguagemodelseducation}. 
\textbf{C2}. LLMs face several challenges when acting as logical fallacy educators, including diversion from the topic \citep{autotutor}, concession to persuasion \citep{xu-etal-2024-earth}, mirroring student texts without critique \citep{wang-etal-2024-book2dial}, and lacking reflective questions \citep{liu2024socraticlm, zhang2024llmsbeathumansdebating}, etc. 


To address these challenges, we design an LLM-based framework for teaching laypeople to recognize logical fallacies along with a scalable way to evaluate such tutors. Specifically, to address \textbf{C1}, we define metrics for logical fallacy tutoring based on our findings in a pilot study and relevant literature in education and argumentation theory. To address \textbf{C2}, we propose \texttt{LFTutor}, a tutoring framework based on 
intent-based pedagogical steering \citep{puech2024pedagogicalsteeringlargelanguage}, 
which aims to help learners identify fallacious arguments through structured, designated strategies in a dialogue setting, avoiding current issues of LLMs as logical fallacy tutors.  

We conduct both automatic and human evaluation to verify the effectiveness of \texttt{LFTutor}.
For automatic evaluation, we use simulated 
students to generate dialogues with \texttt{LFTutor} and baseline LLMs with texts from Elec2Deb20 \citep{goffredo-etal-2023-argument}. We then use QwQ-32B (verified against human judges) to evaluate tutor responses on all metrics. For human evaluation, we recruit 20 participants who interact with \texttt{LFTutor} and then rate its performance against baseline LLMs. Results show that \texttt{LFTutor} achieves superior performance compared to baseline LLMs in both automatic and human evaluation, demonstrating the possibility of using \texttt{LFTutor} in combatting fallacies in the real-world. We summarize our contributions as follows: 


\begin{enumerate}[itemsep=0pt,topsep=1pt]
\item {We 
formulate critical criteria for what makes a good logical fallacy tutor, with insights from a pilot study and a broad literature search (\cref{lfdiag}).}
\item {We develop \texttt{LFTutor}, a tutoring system for educating laypeople about fallacies, using intent-based pedagogical steering (\cref{Method}).}
\item {We conduct automatic and human evaluations, showcasing \texttt{LFTutor}'s superiority 
in fallacy education through comparisons with baseline LLMs on various metrics (\cref{sec:auto_eval} and \cref{humaneval-sec}).}

\end{enumerate}

\section{Background and Related Work}

\myparagraph{Socratic Questioning.} Socratic questioning is a well-known educational technique that aims to challenge underlying student assumptions and uncover faulty reasoning steps. This technique is 
beneficial in stimulating the student's thinking, thus improving their learning outcomes \citep{Yang01092005}. Socratic questioning has been applied in a wide variety of scenarios, including critical thinking instruction \citep{critthinking}, literature seminars \citep{holdencritedd}, and math problem solving \citep{shridhar-etal-2022-automatic}. As logical fallacies often arise from faulty reasoning \citep{goffredo-etal-2023-argument}, Socratic questioning can be particularly relevant in helping learners examine and reflect on their reasoning. 



\myparagraph{Intent-based Dialogue Strategies.} Recent work on LLM agents have focused on dialogue generation guided by users' intents and structured strategies. \citet{jin-etal-2024-persuading} and \citet{xu-etal-2024-earth} designed and utilized persuasion strategies in everyday dialogue scenarios. \citet{ruggeri-etal-2023-dataset} formulated debate strategies to annotate dialogues on a scientific paper between researchers. In the education domain, \citet{puech2024pedagogicalsteeringlargelanguage} and \citet{autotutor} developed tutoring agents that detect students' intents and apply scaffolding techniques. 

\myparagraph{Critical Argumentation in LLMs.}
Critical argumentation \citep{walton2008argumentation} embodies systematic ways of analyzing arguments in dialogues, which is useful for responding to opposing viewpoints with counterarguments. In earlier work, \citet{ARGUMENTS} proposed methods to generate logically sound counterarguments. More recently, \citet{ozaki-etal-2025-llm} showed LLM's ability to produce high-quality counterarguments when targeting implicit or critical premises. \citet{pitre2024argumentoraugmentinguserexperiences} used LLMs to generate counterarguments for op-ed articles to facilitate reader's thinking, and \citet{jeong2025largelanguagemodelsbetter} incorporated counterarguments into LLM's reasoning for enhanced logical fallacy detection, which is partially utilized in our work for \texttt{LFTutor}.



\begin{figure}[t]
\centering

  \includegraphics[width=\columnwidth]{ 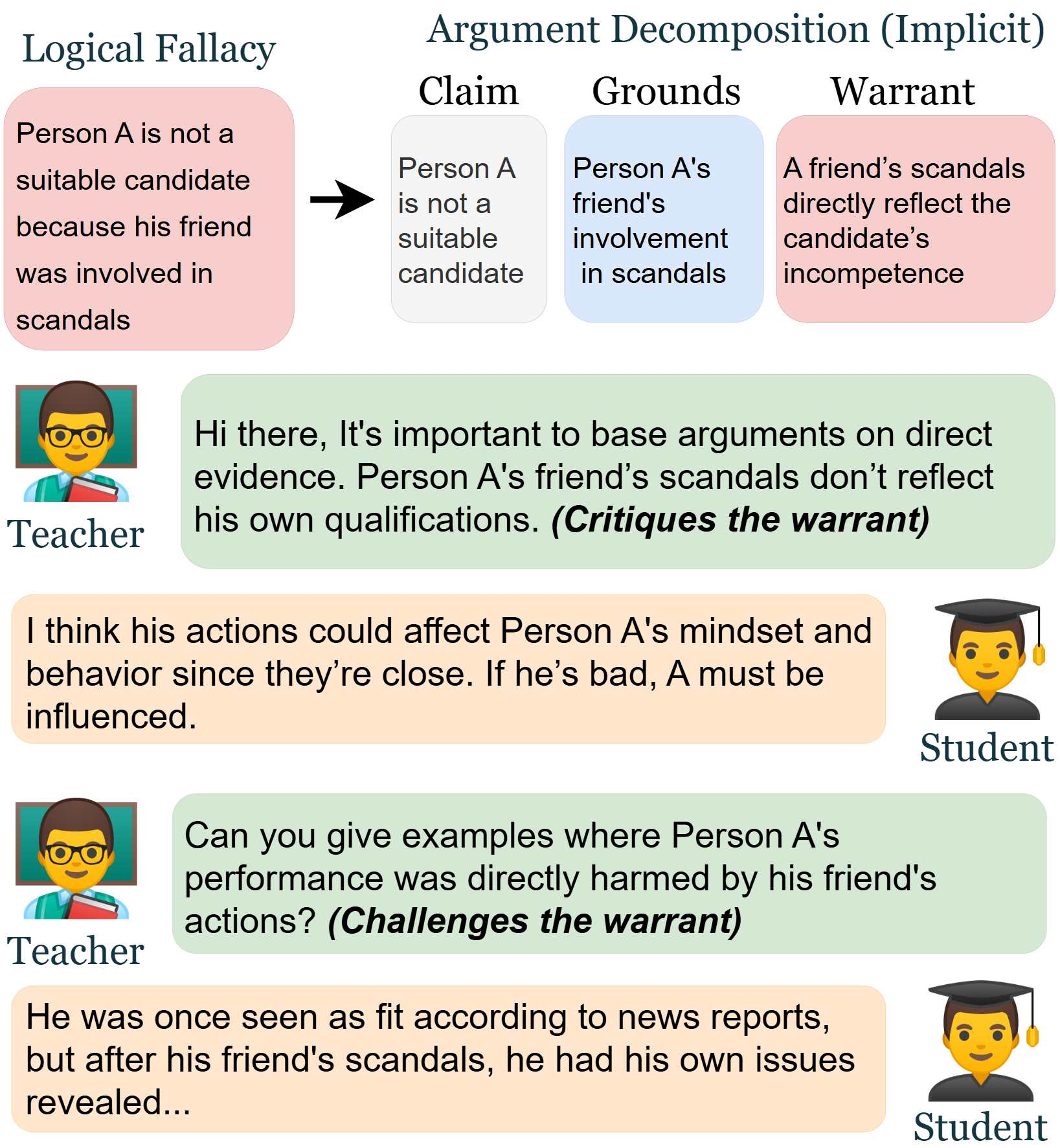}
  \vspace{-20pt}
  \caption{The task of logical fallacy tutoring consists of a fallacious statement decomposed into its corresponding argumentation structure and two personas: a teacher, and a student (layperson) who believes in the fallacy. The logical flaw here lies in the warrant. As shown by the \textbf{bolded} comments, the teacher focuses on the flaw in the warrant, while the student tries to defend its validity.}
  \label{fig:ovv_tec}
  \vspace{-15pt}
\end{figure}

\section{Problem Formulation \label{lfdiag}}



    We frame the task of logical fallacy tutoring as a one-on-one interaction between a \textbf{teacher} and a \textbf{student} centered around a piece of potentially fallacious text (\cref{fig:ovv_tec}). The text contains one or more fallacies of any type (e.g. ad hominem\footnote{Attacking the person proposing the argument instead of the argument itself.}, appeal to emotion\footnote{Argue by evoking emotions rather than relying on logical reasoning or evidence.},  etc.). The teacher aims to educate the student by helping them realize that the text is logically flawed, while the student, who is unfamiliar with fallacies, believes that it is logically valid, and defends their position by debating with the teacher. This multi-turn dialogue setting is useful 
    as it: (1) enables deeper explorations of the 
    fallacious statement using argument decomposition, and (2) allows the teacher to actively engage with the student's underlying cognitive biases, going beyond surface-level identification of fallacies.



A major challenge in this task is that logical fallacies often hinge on contextual assumptions that are not explicitly stated 
\citep{glockner-etal-2025-grounding}. For example, the sentence in \cref{fig:ovv_tec} lacks definitive evidence on Person A's susceptibility to friend's influences, yet this context is critical in judging logical validity. 
To reason about these contexts, we utilize argument decomposition models such as the Toulmin model \citep{toulmin1984introduction}, which posits that a logical argument consists of a claim (the main assertion -- \textit{Person A is not a suitable candidate}), the grounds (the supporting evidence -- \textit{Person A's friend's involvement in scandals}), and the warrant (the underlying assumption linking the grounds to the claim -- \textit{a friend's scandals directly reflect the candidate’s incompetence}). This decomposition allows the teacher in \cref{fig:ovv_tec} to focus on asking the learner for more evidence for the warrant. 

Another major challenge in misinformation education lies in the learner’s cognitive biases,  
which are often reinforced by reasoning structures and deep-rooted assumptions \citep{motivated_irrational}, as shown in their first response equating close relationships with direct influence (\cref{fig:ovv_tec}). Thus, laypeople's fallacious beliefs are often resistant to shallow, one-shot identification and explanation \citep{helwe-etal-2024-mafalda}. In this regard, single-turn feedback is often insufficient, 
and a multi-turn dialogic approach \citep{teach_socra} is needed instead to help the learner engage in critical reflection of reasoning steps. This approach involves argument exchanges, where the teacher builds the discussion with probing questions and targeted challenges to gradually reveal the logical flaw of the learner's reasoning. 

We investigate LLMs' performance in educating laypeople about logical fallacies in this setting. 
Specifically, 
we choose Elec2Deb20 \citep{goffredo-etal-2023-argument}, a dataset on U.S. presidential debates (more details in \cref{dataset}).

\subsection{ LLM's Issues in Logical Fallacy  Tutoring\label{desiderata}}

To our knowledge, no prior work has systematically discussed evaluation metrics for LLM tutors in logical fallacy tutoring. As a first step towards filling this gap, we conduct a pilot study (\cref{ref_pilot_sev}) to identify weaknesses of current LLMs in this task using student-teacher dialogues simulated by GPT-4o. Furthermore, we draw on interdisciplinary insights into combating logical fallacies with reviews of relevant literature in education and logical reasoning. Based on findings from both the pilot study and the literature review, we propose the following evaluation dimensions. Detailed examples and explanations of each issue are presented in \cref{tab:examples}. 

\myparagraph{Divergence.} 
Our pilot study found that 35\% of dialogues contain tutor responses unrelated to discussion of the logical fallacy (e.g. proposing solutions to the political issue in the sentence). However, topic relevance is crucial in education for facilitating meaningful discussions \citep{Tracy01101984}. We thus consider Divergence as a key evaluation dimension.

\myparagraph{Stance Change.} Previous studies have shown that LLMs are prone to be persuaded by fallacious reasoning, where LLMs succumb to persuasion by changing their stance \citep{xu-etal-2024-earth, payandeh-etal-2024-susceptible}. Thus, it is important for the teacher to hold their position when debating, as this 
might
hinder the student's development of critical thinking skills \citep{withey2016mastering}. 

\myparagraph{Repetition.} Our pilot study reveals that LLMs  tend to repeat the student's responses without contributing additional insight (e.g. further questions to the responses) - this occurs in 60\% of all dialogues. Such repetitions are harmful 
as they may increase the chance of the LLM tutor generating repetitive responses \citep{represponse},  leading to student confusion and poorer quality in teaching. 

\myparagraph{Lack of Refutation.} Prior work indicates that LLMs demonstrate sycophancy and lack the skills for ``sustained'' debates, including posing counterarguments \citep{zhang2024llmsbeathumansdebating}. This issue was also present in the pilot study, where 53\% of dialogues lacked explicit challenges to the student. It is crucial for the teacher to actively refute the student's illogical argument when necessary, which helps clarify misconceptions \citep{REFUTE} on the logical structure of the fallacies. 

\myparagraph{Lack of Evidence Inquiry.} \citet{liu2024socraticlm} observe that LLMs lack ``thought-provoking'' teaching paradigms (e.g. requiring the student to provide evidence to support their claims) and simply follow shallow ``question-answering'' structures. In logical fallacy tutoring, the teacher must ask the student for evidence that supports their claims, since this step encourages critical thinking from the Socratic teaching perspective \citep{Yang01092005}.

\myparagraph{Strategy Fixation.} Our pilot study shows in 90\% of dialogues, LLM teacher relies on a single strategy by suggesting the student to consider broader context and balanced perspectives for the logical validity of the sentence. This strategy ignores 
texts with multiple faulty logical %
structures, which leads to less reflective student feedback \citep{krupp2023challengesopportunitiesmoderatingusage}, limiting their scope of thinking. 

\myparagraph{Unexplained LF Terms.} We discovered that in 43\% of generated dialogues, the LLM teacher actively names logical fallacy types e.g. ``ad hominem'', without properly explaining them. This could be avoided, as it may confuse the students if they find the terminology unfamiliar \citep{vocabdiff}, thus hindering effective learning. 

\myparagraph{Passive Guidance.} Previous work \citep{autotutor,liu2024socraticlm} reports that LLMs often fail to assert pedagogical control in education dialogues. Guidance is
important in logical fallacy tutoring, as it can progressively direct the student to focus on their argument's weaknesses \citep{guidance}.
However, passive guidance, where the teacher primarily follows the student without providing clear direction, can negatively affect students' learning outcomes 
\citep{outcome}.

\subsection{Identifying Issues with LLM-as-a-judge}  \label{autoeval_fr}



Based on insights from \cref{desiderata}, we evaluate LLM tutors by measuring the rate at which they avoid these issues. We measure each issue at the dialogue level using the criteria shown in \cref{tab:eval_criteria}, which they either pass or fail. We then collect results from all dialogues and compute the percentages at which the tutor passes the evaluation of each metric. 

When considering LLM candidates as judges \citep{gu2025surveyllmasajudge}, we prefer those with built-in reasoning capabilities due to their superior performance in complex reasoning tasks \citep{deepseekai}. Balancing performance with cost,
we use QwQ-32B \citep{qwq32b, qwen2.5} from the Qwen family for our LLM-as-judge evaluations. 

We verify the reliability of QwQ-32B as a judge by comparing its annotations against those from human experts. We randomly sampled 100 logically fallacious text from Elec2Deb20 and generated 100 dialogues of 10 rounds with the setting in \cref{ref_pilot_sev}. We then let QwQ-32B evaluate all dialogues using prompts containing metrics in \cref{sec:autoeval}. 

For human-annotated golden labels, we recruited three human annotators who are fluent in English and have prior experience with linguistic annotation tasks. We first explained the metrics to each annotator using examples from table \ref{tab:examples}. The annotators then evaluated dialogues with the same prompts as QwQ-32B. After that, they resolved conflicting labels, reaching an agreement level of over 97\% \cite{miles2020qualitative} for all metrics. 

For each metric, we randomly sampled 30 golden dialogues with 15 positive labels and 15 negative labels. We then computed F1-scores (Table \ref{pilot_qwq}) for QwQ-32B against golden labels, which average over 73\% for all metrics. We further validate QwQ-32B's performance by manually reviewing human annotations, having observed that the task is difficult for human annotators, involving long text (700 words) with complex arguments. We performed a direct inspection of the annotator's labels and measured the accuracy of QwQ-32B's annotation, shown in \cref{pilot_qwq_accuracy}. We found that the accuracy of QwQ-32B was greater than or equal to 80\% for all metrics, which ensures its reliability as the judge.  



\section{LFTutor} \label{Method}

\begin{figure*}[t]
\centering
  \includegraphics[width=2.1\columnwidth]{ 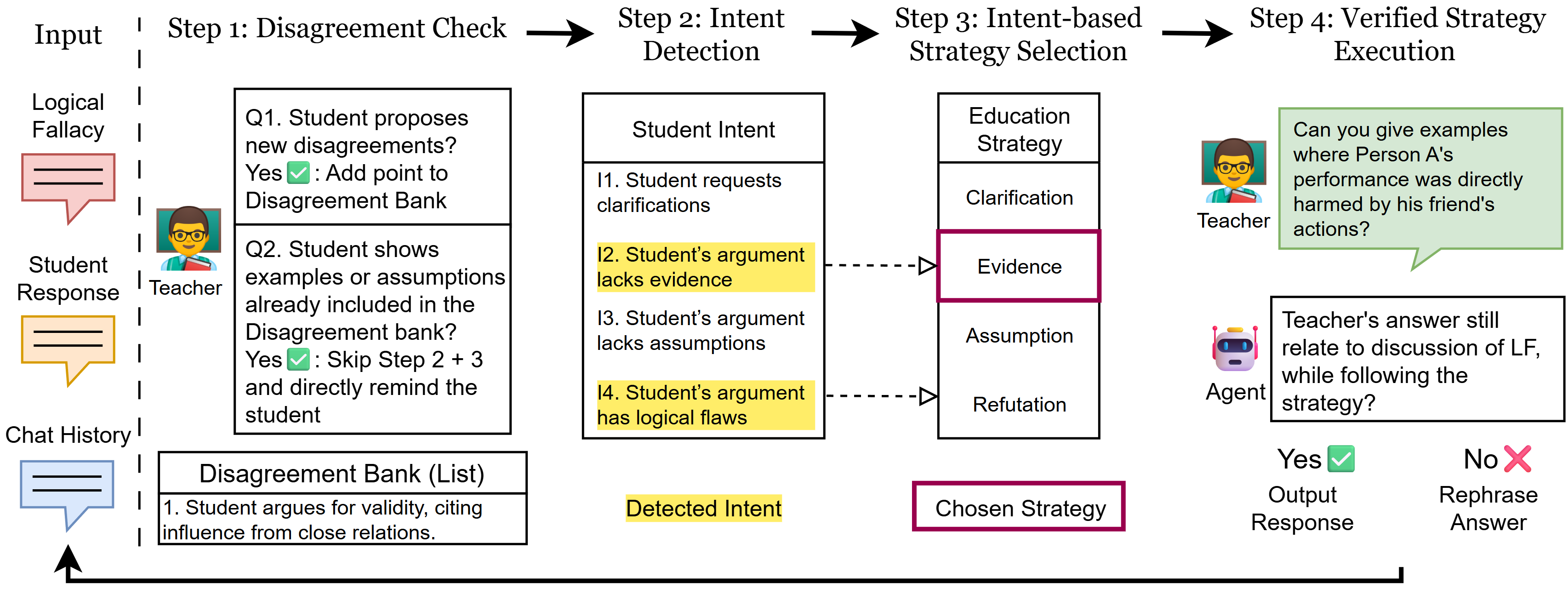}
  \caption{ \texttt{LFTutor} processes the student's response in four steps. In step 1, \texttt{LFTutor} checks the response for (1) new conflicts, which are added to the disagreement bank, or (2) duplicate examples or assumptions, which leads to reminders to propose new examples or assumptions, skipping step 2 and 3. In step 2, \texttt{LFTutor} analyzes the response for intents, each of which corresponds with an education strategy on the same row. In step 3, \texttt{LFTutor} chooses one strategy based on detected intent and priorities of components in Toulmin's model. In step 4, \texttt{LFTutor} executes the strategy, followed by an agent ensuring that the response reflects the selected strategies while being relevant.  
} 
  \label{fig:OVV}
  \vspace{-10pt}
\end{figure*}

As stated in \cref{lfdiag}, logical fallacy education faces two main challenges: 1. Logical fallacies hinge on contextual assumptions, which requires argument decomposition to pinpoint; 2. Fallacy believers often harbor cognitive biases and are resistant to one-shot identification and explanation. To address them, we propose \texttt{LFTutor}, a dialogue tutoring agent that (i) locates logical flaws in the student's responses by implicitly decomposing them into claims, grounds and warrants, and (ii) interactively adapts to the student's intents over multiple turns. At each turn, \texttt{LFTutor} analyzes the student's responses using the Toulmin model to identify the text component containing logical flaws. It then applies a strategy tailored to that component to guide the student towards recognizing flaws. This technique, called intent-based pedagogical steering \citep{puech2024pedagogicalsteeringlargelanguage}, structures the tutor’s output by detecting the student’s intent (implied purposes behind their response, e.g., asking for clarification), and then executing tutoring strategies aligned with the detected intent. We integrate this design into a four-step framework (\cref{fig:OVV}): (1) Disagreement Check, (2) Intent Detection, (3) Intent-based Strategy Selection, and (4) Verified Strategy Execution. Detailed pseudocode is shown in \cref{alg:lftutor}.

    
    
    
    



\myparagraph{Step 1: Disagreement Check}. To structure the tutor-student dialogue and help \texttt{LFTutor} focus on specific parts of the text they disagree on, we introduce \textit{disagreement bank}, a long-term memory module \citep{Zhong_Guo_Gao_Ye_Wang_2024} that
stores a list of disagreement points between the two parties. 
Given the student's response from the last turn, the text with logical fallacies, and the disagreement bank, \texttt{LFTutor} checks the response using another agent (\cref{disagreement-bank}) for whether (1) the student's claim, if opposing, is contained in the bank (2) the examples or assumptions are contained in the bank. If the opposing claim is newly made, then \texttt{LFTutor} adds the complete argument to the disagreement bank, which it aims to address in this round, regardless of the contents from examples or assumptions. However, if the claim has been discussed, then the examples or assumptions must be different from those in the bank. Otherwise, \texttt{LFTutor} will skip the next steps and request the student to provide new arguments that support their position.

\myparagraph{Step 2: Intent Detection}. Given the student's argument, \texttt{LFTutor} analyzes it (\cref{intent_prompt}) for potential intents: 
(I1) student requests clarification on the logical flaw of the text; (I2) student's argument lacks evidence supporting their claim; (I3) student's argument lacks assumptions linking evidence to their claim; (I4) student's argument is complete, but has clear logical flaws. These intents align naturally with the Toulmin model: I1 guides the student to formulate a clear claim in their response, I2 targets the student's missing grounds, I3 highlights the student's missing warrants, and I4 identifies weaknesses in the logical chain as a whole. 
All detected intents (one or more) are stored for \texttt{LFTutor}'s reference in the next step. 

\myparagraph{Step 3: Intent-based Strategy Selection}. Given the detected intents, \texttt{LFTutor} focuses on addressing the student's reasoning flaws by selecting an appropriate education strategy related to their intent. We draw strategies from two main pedagogical theories: (1) Socratic questioning \citep{holdencritedd}, which aims for dialogic, open-ended discussions on logical fallacies by probing the student's reasoning, and (2) critical argumentation \citep{walton2008argumentation}, which allows \texttt{LFTutor} to target weaknesses of the student's argument and provide sound refutations, nudging them to reflect on argument quality. We include four strategies corresponding to the aforementioned student intents: Responding, Evidence, Assumption, and Refutation. Prompts and example responses are shown in \cref{edustrat_prompt} and \cref{ref_res_teacher}.

\myparagraph{(i) Responding:} When the student requests explanations on the logical flaws in their argument or the logical fallacies, \texttt{LFTutor} clarifies and explains the logical flaw from those statements. 

\myparagraph{(ii) Evidence:} When the response lacks sufficient evidence, \texttt{LFTutor} asks the student for evidence supporting the claim. This is similar to Socratic questions on evidence shown in 
\cite{Yang01092005}.

\myparagraph{(iii) Assumption:} When the student's response makes unclear assumptions, \texttt{LFTutor} challenges the student's hidden or unsupported assumptions in the argument. This is akin to Socratic questions on assumptions discussed in \cite{Yang01092005}. 

\myparagraph{(iv) Refutation:} When the student's argument contain clear logical flaws, \texttt{LFTutor} 
provides logical refutations to the student's argument, using methods adopted from critical argumentation \citep{ARGUMENTS}. 

\texttt{LFTutor} selects strategies in the order listed above (i $\rightarrow$ iv), matching the priority of components in the Toulmin model \citep{Hitchcock2017_toulmin}: claims form an argument's foundation ($\rightarrow$Responding), followed by grounds ($\rightarrow$Evidence), then warrants ($\rightarrow$Assumption), and finally, rebuttal to a complete argument ($\rightarrow$Refutation). Clarification of logical flaw(s) (Responding) introduces a counter-claim, leading the student to reconsider and refine their position. After establishing the claim, \texttt{LFTutor} presses the student to provide relevant proof (Evidence). Assumption comes once claim and grounds are both present, and \texttt{LFTutor} challenges the student to form connections between them. Finally, \texttt{LFTutor} applies Refutation when it obtains a complete argument, urging the student to examine their argument's logical validity. By structuring strategies according to this hierarchy, 
\texttt{LFTutor} guides the student through argument construction in a stepwise fashion by systematically exposing component-level flaws.



\myparagraph{Step 4: Verified Strategy Execution}.  In the final step, \texttt{LFTutor} 
 applies the chosen strategy to respond to the student (\cref{transition-prompt}). 
Then, another LLM verifies if the response adheres to the chosen strategy (\cref{check-res-pr}).
If this verification fails, the LLM in \texttt{LFTutor} is asked to rephrase its response to meet the requirements. This step guarantees that \texttt{LFTutor}’s response respects the strategy while remaining consistent with the argument structure.

\section{Automatic Evaluations} \label{sec:auto_eval}

We begin by evaluating \texttt{LFTutor} using a set of automatic evaluations. We first describe the baseline LLM tutors that \texttt{LFTutor} will be compared with (\cref{baseline}) and the student design used in our evaluations (\cref{stu-design}). We then describe the detailed experimental settings (\cref{exp-auto}) and the results (\cref{autoeval-res}).

 \subsection{Baseline LLM Tutors\label{baseline}}

We designed two baselines for the tutor LLM 
for comparison. The first (\textsc{Base}) uses a role-playing prompt, where the tutor is instructed to act as an educator whose goal is to convince a potentially biased student that the given text contains logical fallacies (\cref{base_prompt}). The second (\textsc{Base w/ Problems}) improves upon \textsc{Base} by incorporating descriptions of the issues in logical fallacy education (\cref{desiderata}) in addition to role-playing instructions, and the tutor is asked to avoid those issues (\cref{basew_prompt}). This serves as a heuristic-enhanced baseline that attempts to mitigate the tutor's pitfalls.


\subsection{Student Design \label{stu-design}}

For our automated experiments, we make use of simulated students, which is a common approach in recent works \citep{macina-etal-2023-mathdial, liu2024socraticlm}. We prompt GPT-4o to roleplay as a student who believes in a logically fallacious statement. However, we wish to account for the fact that some of the issues discussed in \cref{desiderata} might be too rare to be triggered by a \textbf{normal} student LLM, but may cause issues if triggered in a real conversation.  Therefore, we also create \textbf{adversarial} student personas, which are specifically trying to trigger each issue. 
 \myparagraph{The normal student} debates with the tutor by defending their position with evidence and explanations, believing that the text containing logical fallacies is logically valid (\cref{prompt-norm-stu}). To create a ``robust'' student, we prompt the LLM to utilize various debating techniques (\cref{student-strategy}) in critical argumentation \citep{walton2008argumentation, ARGUMENTS}. 
 
 \myparagraph{The adversarial student} intends to trigger the tutor's issues discussed in 
 \cref{desiderata}. 
 We use role-playing techniques \citep{zhou2024sotopia} to design students that drive the tutor towards behaviors matching the issues' descriptions (\cref{prompts-adv-stu}). For ``Divergence'', for example, we ask the LLM to play a student who often talks about information irrelevant to the discussion of the logical fallacy. This way, we evaluate the robustness of tutors under worst case scenarios. We exclude adversarial students for Lack of Refutation and Lack of Evidence Inquiry, as pilot studies show that these issues are primarily attributed to the lack of instructions in tutor's prompts. 



\subsection{Experiment Setting \label{exp-auto}}

We begin by experimenting with GPT-4o as our base LLM. We divide our setup into two parts following our design from \cref{stu-design}, with more details in \cref{Autoeval_imp_detail}. For dialogues with \textbf{normal students}, we compare \textsc{Base}, \textsc{Base w/ Problems}, and \texttt{LFTutor}. For each framework, we generated 1,000 dialogues using randomly sampled examples from Elec2Deb20, evaluated using QwQ-32B as a judge (\cref{tab:Res_exp_normal}). For dialogues with \textbf{adversarial students},  we compare \textsc{Base w/ Problems} with \texttt{LFTutor}. For each adversarial student, we generated 1,000 dialogues from both frameworks using randomly sampled sentences from Elec2Deb20, evaluating them using QwQ-32B on the corresponding metrics (e.g., evaluating Divergence on dialogues using divergent student). Results are shown in \cref{tab:Res_exp_adv}.

We also tested Gemini-2.5-pro \citep{gemini25pushingfrontier} and LLaMA-3.1 \citep{llama3herdmodels} as backbones for \textsc{Base} and \texttt{LFTutor} under the \textbf{normal student} setting, using 200 random examples from Elec2Deb20. The results are broadly consistent with the findings from GPT-4o. More details can be found in \cref{other_llm_exps}.




\subsection{Results \label{autoeval-res}}

\begin{table}[t]
\small
  \centering
  \resizebox{\linewidth}{!}{
  \begin{tabular}{m{3.5cm} m{0.8cm} m{1.5cm} m{1.1cm}}
    \toprule
    \textbf{Evaluation Metrics} & \textsc{Base} & \textsc{Base w/ problems} & \texttt{LFTutor}\\
    \midrule
    \textbf{Divergence}  & $69.3$ &$74.0$ &  \textbf{84.6}  \\
    \textbf{Stance Change}  & $8.7$ & $48.1$& \textbf{87.9}    \\
    \textbf{Repetition}  & $13.0$&$42.5$ &\textbf{78.3}        \\
    \textbf{Lack of Refutation}  & $54.9$ & \textbf{99.9} & $99.6$       \\
    \textbf{Lack of Evidence Inquiry} & $11.1$ & $95.7$& \textbf{96.1} \\
    \textbf{Strategy Fixation} & $43.4$& $63.0$& \textbf{91.2}    \\
    \textbf{Unexplained LF Terms}  & $49.2$ & $27.6$ & \textbf{95.0}  \\
    \textbf{Passive Guidance} & $4.4$& $41.5$ & \textbf{43.6}   \\
    \textbf{Avg. Performance} & $31.2$& $61.5$ & \textbf{84.5}   \\
    \bottomrule
  \end{tabular}}
   \vspace{-5pt}
  \caption{Percentages of dialogues with normal students for each framework passing each metric, with GPT-4o as backbone. Evaluated using QwQ-32B. Higher is better. }
  \label{tab:Res_exp_normal}
  \vspace{-5pt}
\end{table}

\begin{table}[t]
\small
  \centering
  \resizebox{\linewidth}{!}{
  \begin{tabular}{m{3.2cm} m{1.6cm} m{1.3cm}}
    \toprule
    \textbf{Evaluation Metrics / Adversarial Type} & \textsc{Base w/ problems} & \texttt{LFTutor}\\
    \midrule
    \textbf{Divergence}  & $11.9$ &\textbf{38.0}   \\
    \textbf{Stance Change}  & $37.3$ & \textbf{84.0}     \\
    \textbf{Repetition}  & $27.0$ & \textbf{82.0}        \\
    \textbf{Strategy Fixation} & $15.4$& \textbf{32.5}    \\
    \textbf{Unexplained LF Terms}  & $49.5$ & \textbf{86.2}   \\
    \textbf{Passive Guidance} & $40.3$ & \textbf{41.2}    \\
    
    \textbf{Avg. Performance} & $30.2$ & \textbf{60.7} \\
    \bottomrule
  \end{tabular}}
  \vspace{-5pt}
  \caption{Percentage of dialogues with adversarial students passing the corresponding metric, with GPT-4o as backbone. 
  Evaluated using QwQ-32B. Higher is better.} 
  \label{tab:Res_exp_adv}
  \vspace{-15pt}
\end{table}
\myparagraph{With the ``normal'' student design, \texttt{LFTutor} outperforms \textsc{Base} by up to $\mathbf{53\%}$ on average, and up to $\mathbf{85\%}$ for individual metrics. It outperforms \textsc{Base w/ Problems} by $\mathbf{23\%}$ on average, and up to $\mathbf{67\%}$ for individual metrics except Lack of Refutation.} 
To understand where these improvements come from, we ablate \texttt{LFTutor}'s core design choices (\cref{ablation}), which shows that they stem mostly from strategy verification and intent-based strategy selection. Strategy verification ensures that \texttt{LFTutor} remains faithful to the pedagogical goal at each turn, boosting scores in Divergence and Evidence Inquiry. Intent-based strategy selection allows targeted execution of one strategy per turn and dynamic strategy switch across turns, which helps with Stance Change, Strategy Fixation, and Unexplained LF Terms. Strategy priority also matters, as it improves Guidance through progressively uncovering flaws within the student's arguments. 






 While \textsc{Base w/ problems} shows improvements over \textsc{Base} in all categories except Unexplained LF Terms, it still performs much worse than \texttt{LFTutor}. We observe a decline in the performance of \textsc{Base w/ problems} in long dialogues, where it often begins by combining counterarguments with evidence inquiry, but gradually loses focus on tutoring by yielding control to the student. This suggests that simply instructing LLMs to avoid issues in logical fallacy education is not enough for maintaining dialogue quality over time.
\myparagraph{For the adversarial student design, \texttt{LFTutor} outperforms \textsc{Base w/ problems} by up to $31\%$ on average in all adversarial student types.} The difference is more prominent in Divergence, Stance Change, Repetition, and Unexplained LF Terms. Although the student exhibits manipulative behaviors such as repeated ordering to \texttt{LFTutor} to follow their words, \texttt{LFTutor} retains performance thanks to intent-based strategy selection and verified strategy execution. This ensures that \texttt{LFTutor} always responds according to the designated strategies, whose content is robust from external influences. 

However, \texttt{LFTutor} suffers performance drops in Divergence and Strategy Fixation. For Divergence, \texttt{LFTutor}'s 
effort is shifted from educating students about logical fallacies to preventing them from mentioning irrelevant content, which drifts from actual discussions of logical validity. For Strategy Fixation, the adversarial student only prefers a single strategy (e.g. one that emphasizes broader context). \texttt{LFTutor} does not enforce the variety of strategies chosen, thus its responses are always directed towards explaining the insufficiency of focusing on the fallacy's broader context. 
\begin{table}[t]
\small
\resizebox{\linewidth}{!}{
  \centering
  \begin{tabular}{m{3.5cm} m{0.6cm} m{0.8cm} m{0.8cm} m{1cm}}
    \toprule
    \textbf{Evaluation Metrics} & \textsc{Base} mean & \texttt{LFTutor} mean & Mean Diff & p-value\\
    \midrule
    \textbf{Divergence}   & 1.65 & 3.30 & +1.65 & \textbf{0.00**} \\
    \textbf{Stance change}   & 1.75 & 3.10 & +1.45 & \textbf{0.00**}  \\
    \textbf{Repetition}  & 2.65 & 3.10 & +0.45 & 0.08       \\
    \textbf{Lack of Refutation}   & 3.00 & 4.15 & +1.15 & \textbf{0.00**}     \\
    \textbf{Lack of Evidence Inquiry}   & 2.65 & 4.20 & +1.55 & \textbf{0.00**}\\
    \textbf{Strategy Fixation}  & 1.35 & 2.15 & +0.80 & \textbf{0.01*}   \\
    \textbf{Unexplained LF Terms}   & 2.30 & 3.00 & +0.70 & \textbf{0.03*}  \\
    \textbf{Passive Guidance}  & 2.90 & 4.00 & +1.10 & \textbf{0.00**}  \\
    \textbf{Helpfulness}  & 3.35 & 4.15 & +0.8 & \textbf{0.01*} \\
    \bottomrule
  \end{tabular}
  }
  \vspace{-5pt}
  \caption{Mean of Likert-scale participant ratings (1-5)   for each framework, higher is better. Rightmost column includes p-values from t-tests of ratings, lower is better. }
  \vspace{-15pt}
  \label{tab:res_exp_hum}
  
\end{table}

\section{Human Evaluations} \label{humaneval-sec}

Finally, to understand how \texttt{LFTutor} might work in a more real-world setting, we conduct pilot human evaluations comparing \texttt{LFTutor} and \textsc{Base}. We choose \textsc{Base} as an evaluation baseline since it resembles behaviors of real-world LLMs with minimal knowledge of issues in logical fallacy education, instead of \textsc{Base w/ problems} that performs slightly better in automatic evaluation but fails to reflect LLMs' real-world exposure to logical fallacies. The evaluation procedure is shown in \cref{hum_proc}, with the results discussed in \cref{hum-quant} and \cref{qual-hum}.

\subsection{Evaluation Procedure } \label{hum_proc}
We selected two texts from Elec2Deb20 that participants voted logically valid (\cref{participant_selection}) for the evaluation. 
To align with previous work's evaluation settings for interactive systems \citep{jiang-etal-2024-unknown, fang-etal-2023-user}, we recruited 20 participants fluent in written and spoken English to engage in debates with the tutors, with the order of tutors and sentences randomly chosen. 


During the study, every participant interacted with each tutor for 5 to 10 rounds each. They evaluated the tutor after each session (\cref{fig:impres}) with metrics (\cref{tab:eval_criteria}) on a Likert scale \citep{likert1932technique} from 1 to 5, where 1 indicated observation of the issue at every turn, and 5 indicated no observation of such issue. The participant also completed a post-study form (\cref{fig:post_study_1}, \ref{fig:post_study_2}), where they compared the tutors \texttt{LFTutor} and \textsc{Base}. In addition to existing metrics, we included ``Helpfulness'' to measure the subjective learning helpfulness of the tutor for the participant, which they rated on the same Likert scale. A score of 1 indicated least helpful for logical fallacy understanding, and 5 indicated extremely helpful.

To determine whether the differences of human ratings between \texttt{LFTutor} and \textsc{Base} are statistically significant, we conducted quantitative analysis for all scores. We first ran Shapiro-Wilk Test \cite{shapiro} to test the normality of each metric's score distribution. Observing that they distribute normally, 
we ran one-sided t-tests  \cite{ttest} to compare the scores between \texttt{LFTutor} and \textsc{Base} chatbot, with results shown in \cref{tab:res_exp_hum}.

Finally, we also performed a summative qualitative analysis of the participants' responses, by examining their chat histories and post-study forms, and then identifying key phrases from their descriptions of \texttt{LFTutor} and \textsc{Base} and extracting recurring themes that came up in the dialogues.

\begin{table*}[t]
\centering
\scriptsize
\resizebox{\textwidth}{!}{
\begin{tabular}{lcccccccc}
\toprule
\textbf{Variant} &
\textbf{Div.} &
\textbf{Stance} &
\textbf{Repet.} &
\textbf{No Refut.} &
\textbf{No Evid.} &
\textbf{Strat. Fix.} &
\textbf{Unexpl. LF} &
\textbf{Passive} \\
\midrule
LFTutor (Reference) 
& \textbf{84.6} & \textbf{87.9} & 78.3 & 99.6 & \textbf{96.1} & 91.2 & \textbf{95.0} & \textbf{43.6} \\

-- No Disagreement Check
& 83{\tiny$\downarrow$1.6}
& 78{\tiny$\downarrow$9.9}
& \textbf{92}{\tiny$\uparrow$13.7}
& \textbf{100}{\tiny$\uparrow$0.4}
& 89{\tiny$\downarrow$7.1}
& 80{\tiny$\downarrow$11.2}
& 82{\tiny$\downarrow$13}
& 30{\tiny$\downarrow$13.6} \\

-- No Strategy Verification
& 66{\tiny$\downarrow$18.6}
& 76{\tiny$\downarrow$11.9}
& 75{\tiny$\downarrow$3.3}
& 96{\tiny$\downarrow$3.6}
& 55{\tiny$\downarrow$41.1}
& 86{\tiny$\downarrow$5.2}
& 84{\tiny$\downarrow$11.0}
& 38{\tiny$\downarrow$5.6} \\

-- Reverse Strategy Order
& 83{\tiny$\downarrow$1.6}
& 80{\tiny$\downarrow$7.9}
& 71{\tiny$\downarrow$7.3}
& 98{\tiny$\downarrow$1.6}
& 94{\tiny$\downarrow$2.1}
& \textbf{92}{\tiny$\uparrow$0.8}
& 81{\tiny$\downarrow$14}
& 23{\tiny$\downarrow$20.6} \\

-- Multi-Strategy / Turn
& 68{\tiny$\downarrow$16.6}
& 61{\tiny$\downarrow$26.9}
& 75{\tiny$\downarrow$3.3}
& 99{\tiny$\downarrow$0.6}
& 95{\tiny$\downarrow$1.1}
& 71{\tiny$\downarrow$20.2}
& 75{\tiny$\downarrow$20}
& 42{\tiny$\downarrow$1.6} \\
\bottomrule
\end{tabular}
}
\vspace{-5pt}
\caption{
Percentages of dialogues passing each metric under different ablations of \texttt{LFTutor}.
Values indicate absolute performance; arrows show change relative to the reference model.
\textbf{Bold} denotes the best-performing variant per metric.
Evaluation is conducted automatically using QwQ-32B as a judge.
}
\label{tab:res_ablation}
\vspace{-15pt}
\end{table*}

\subsection{Quantitative Analysis on Ratings} \label{hum-quant}

\myparagraph{\texttt{LFTutor} achieves statistically significant better scores than \textsc{Base} on eight out of nine metrics.} The improvement is significant ($p < 0.05$) for Strategy Fixation, Unexplained LF Terms, and Helpfulness, and highly significant ($p<0.01$) for Divergence, Stance Change, Lack of Refutation, Lack of Evidence Inquiry, and Passive Guidance. This difference is likely due to \texttt{LFTutor}'s explicit emphasis towards detailed probing questions and refutations to users' responses, which allows it to focus on tutoring logical fallacy and remain strong in its position. These characteristics are directly reflected in \texttt{LFTutor}'s responses but not in \textsc{Base}, which leaves a strong impression for the participants.  



\myparagraph{\texttt{LFTutor} shows limited improvement in Repetition compared with \textsc{Base}.} Repetition is less prominent in \textsc{Base} for dialogues with participants compared to LLM-simulated students, due to participants' usage of more varied debating strategies and Student LLMs' lack of strategy variations. 
This tendency decreases the likelihood of repetitive responses from \textsc{Base} in human evaluation. 


\subsection{Qualitative Analysis of 
Dialogues} \label{qual-hum}




\myparagraph{Participants’ perceptions on \texttt{LFTutor} and \textsc{Base} Align with Quantitative Ratings.} For \texttt{LFTutor}, ten out of twenty participants noted that it is more \textit{focused} during the dialogue, often trying to keep its attention on the logical fallacies in the text. This corresponds to \texttt{LFTutor}'s higher scores on Divergence and Stance Change. 
Nine participants also appreciated ``ample guidance'' provided by \texttt{LFTutor}, depicting its usage of strategies that prompted deeper thinking. This leads to their preferences on \texttt{LFTutor} when rating on Guidance, Helpfulness, Lack of Refutation, Lack of Evidence Inquiry, and Strategy Fixation. 

In contrast, thirteen out of twenty participants described \textsc{Base} framework as debating in a \textit{more general} setting, often lacking specificity in their responses, while six participants noticed that it is more easily persuaded. These behaviors contributed to their thought that \textsc{Base} can be easily manipulated, which led to negative impressions. 

\myparagraph{Participants argue with rich but inconsistent tactics; on the other hand, student LLMs argue with consistent, low-variation logic.} 
Compared with LLM-simulated students, we found that our participants employed a wider range of strategies targeting the weaknesses of the tutor's responses. For example, five of the participants challenged the tutors by claiming that the situations they described seem too ideal, while three participants tried to present direct counterexamples to the tutor's argument. Moreover, we found that the participant's arguments seemed more inconsistent, often relying on emotional appeals (e.g. ``You don't care about my opinion'') or attributions to a single cause. 
 In contrast, we found that LLM students rarely challenge the tutor's responses, but focus more on creating interpretations that make the sentence logically valid. Their arguments are more logically complete with claims followed by evidence, yet they lacked variation in strategies and are often repetitive. 

\section{Ablations} \label{ablation}

To investigate the contribution of different design choices in \texttt{LFTutor} towards evaluation improvements, we conducted an ablation study under four settings: \textbf{(1)} removing step 1: Disagreement Check, \textbf{(2)} removing strategy verification and correction, \textbf{(3)} reversing the priority of pedagogical strategies in step 3, from (Responding $\rightarrow$ Evidence $\rightarrow$  Assumption $\rightarrow$ Refutation) to (Refutation $\rightarrow$ Assumption $\rightarrow$ Evidence $\rightarrow$ Responding), and \textbf{(4)} allowing multiple pedagogical strategies per turn by ignoring Intent-based Strategy Selection. 
Due to budget limitations, we conducted the study using the normal student setting discussed in \cref{exp-auto} with 100 sentences randomly selected from Elec2Deb20.  Results under these settings are shown in \cref{tab:res_ablation}. 

\myparagraph{Removing Step 1: Disagreement Check led to reduced performance in all metrics except Repetition}. The increase of score in Repetition can be attributed to fewer reminders for the student to propose new examples or assumptions, where \texttt{LFTutor} must reiterate the student's claims and grounds. Passive Guidance suffers the most among all metrics, since \texttt{LFTutor} is unable to keep track of topics in the dialogue and mainly analyzes and tackles the student's immediate responses as the discussion continues.

\myparagraph{Removing Step 4: Verified Strategy Execution significantly reduces \texttt{LFTutor}'s performance in most metrics}. The decrease of scores is significant in Divergence, Lack of Evidence Inquiry, and Stance Change. This shows that strategy verification is crucial in reducing \texttt{LFTutor}'s hallucinations towards irrelevant responses, while ensuring it remains faithful to the designated pedagogical goals. 

\myparagraph{Reversing Strategy Order harms performance the most in Unexplained LF Terms and Passive Guidance, with minor changes of performance in other metrics}. Setting Refutation as the top strategy led to unwanted usage of LF terms, since refutation by pointing out the LF in the opponent's words is common (shown in \cref{tab:Res_exp_normal}) for baseline LLMs. Passive Guidance is also negatively affected since \texttt{LFTutor} now primarily treats the student's responses as complete arguments and aims to directly defeat them, rather than 
attempting to uncover the student's supporting evidence and assumptions. 

\myparagraph{Allowing Multiple Strategies per Turn resulted in decreased performance across all metrics}, of which Divergence, Stance Change, Strategy Fixation, and Unexplained LF Terms suffered the most declines. When \texttt{LFTutor} is not explicitly prompted to follow certain strategies, it tends to adhere to the forms of the typical responses from the given strategy, yet it has no inherent knowledge on which strategy are appropriate against different types of student responses. This loss of focus caused significant performance declines in many metrics.

\section{Conclusion}
We explored the novel task of educating laypeople to identify logical fallacies in everyday discourse through multi-turn dialogues with LLM-based tutors. We introduced eight evaluation metrics to assess tutor responses and developed \texttt{LFTutor}, an LLM tutor system with intent-based pedagogical strategies that helps learners recognize logical fallacies. Our extensive automatic and human evaluation showed that \texttt{LFTutor} is an effective and pedagogically grounded solution for misinformation education, paving a way for designing more AI tutors that foster critical thinking through structured reflection, reasoning and interactive engagement.

\section*{Limitations}


\myparagraph{Limited Performance Gain on Guidance.} Although \texttt{LFTutor} managed to improve upon the \textsc{Base} framework on all criteria during automatic evaluation, it did not achieve high scores ($ \ge 75\% $) in Passive Guidance. We attribute this to the lack of structured, detailed, and progressive dialogue goals (e.g. first tackle the evidence, then tackle the warrant, then narrow down the discussion to certain aspects of the warrant.), which needs to be closely integrated with tutoring strategies to create responses that are robust to passive influence. Future works could consider introducing a list of high-level education goals and procedures for \texttt{LFTutor} to follow, such that it combines those goals and tutoring strategies to produce responses that actively control the flow of the dialogues.

\myparagraph{Scope of Evaluation Metrics}
In this paper, we investigated recurring problems with LLM tutors in dialog-based tutoring of logical fallacies and proposed an automatic evaluation framework with a curated set of eight evaluation metrics motivated by prior work and empirical relevance. While the proposed evaluation framework covers several key dimensions for measuring the quality of logical fallacy educational dialogues, such dialogues are inherently multi-faceted. Additional aspects, such as emotional support, personalization, and ease of understanding are not captured by our current metrics and they represent promising directions for future work.

\section*{Acknowledgments}
This work was funded by grants from the Swiss National Science Foundation (Grant number
10009282) and the Swiss AI initiative ({\url{https://www.swiss-ai.org/compute-grants}). This research was supported by the ETH AI Center through an ETH AI Center doctoral fellowship to Junling Wang.
We would like to thank all participants involved in the annotation and interactive experiments, as well as the anonymous reviewers for their constructive comments which helped improve this work. We also give special thanks to Prof. Daqiang Li from Jilin University and Prof. Jose A. Gascon from University of Murcia for insightful discussions. 


\bibliography{anthology,custom}

@article{Zhong_Guo_Gao_Ye_Wang_2024, title={MemoryBank: Enhancing Large Language Models with Long-Term Memory}, volume={38}, url={https://ojs.aaai.org/index.php/AAAI/article/view/29946}, DOI={10.1609/aaai.v38i17.29946}, abstractNote={Large Language Models (LLMs) have drastically reshaped our interactions with artificial intelligence (AI) systems, showcasing impressive performance across an extensive array of tasks. Despite this, a notable hindrance remains—the deficiency of a long-term memory mechanism within these models. This shortfall becomes increasingly evident in situations demanding sustained interaction, such as personal companion systems, psychological counseling, and secretarial assistance. Recognizing the necessity for long-term memory, we propose MemoryBank, a novel memory mechanism tailored for LLMs. MemoryBank enables the models to summon relevant memories, continually evolve through continuous memory updates, comprehend, and adapt to a user’s personality over time by synthesizing information from previous interactions. To mimic anthropomorphic behaviors and selectively preserve memory, MemoryBank incorporates a memory updating mechanism, inspired by the Ebbinghaus Forgetting Curve theory. This mechanism permits the AI to forget and reinforce memory based on time elapsed and the relative significance of the memory, thereby offering a more human-like memory mechanism and enriched user experience. MemoryBank is versatile in accommodating both closed-source models like ChatGPT and open-source models such as ChatGLM. To validate MemoryBank’s effectiveness, we exemplify its application through the creation of an LLM-based chatbot named SiliconFriend in a long-term AI Companion scenario. Further tuned with psychological dialog data, SiliconFriend displays heightened empathy and discernment in its interactions. Experiment involves both qualitative analysis with real-world user dialogs and quantitative analysis with simulated dialogs. In the latter, ChatGPT acts as multiple users with diverse characteristics and generates long-term dialog contexts covering a wide array of topics. The results of our analysis reveal that SiliconFriend, equipped with MemoryBank, exhibits a strong capability for long-term companionship as it can provide emphatic response, recall relevant memories and understand user personality.}, number={17}, journal={Proceedings of the AAAI Conference on Artificial Intelligence}, author={Zhong, Wanjun and Guo, Lianghong and Gao, Qiqi and Ye, He and Wang, Yanlin}, year={2024}, month={Mar.}, pages={19724-19731} }

@article{likert1932technique,
  title={A technique for the measurement of attitudes},
  author={Likert, Rensis},
  journal={Archives of Psychology},
  volume={140},
  pages={1--55},
  year={1932}
}

@misc{gemini25pushingfrontier,
      title={Gemini 2.5: Pushing the Frontier with Advanced Reasoning, Multimodality, Long Context, and Next Generation Agentic Capabilities}, 
      author={Gheorghe Comanici and Eric Bieber and Mike Schaekermann and Ice Pasupat and Noveen Sachdeva and Inderjit Dhillon and Marcel Blistein and Ori Ram and Dan Zhang and Evan Rosen and Luke Marris and Sam Petulla and Colin Gaffney and Asaf Aharoni and Nathan Lintz and Tiago Cardal Pais and Henrik Jacobsson and Idan Szpektor and Nan-Jiang Jiang and Krishna Haridasan and Ahmed Omran and Nikunj Saunshi and Dara Bahri and Gaurav Mishra and Eric Chu and Toby Boyd and Brad Hekman and Aaron Parisi and Chaoyi Zhang and Kornraphop Kawintiranon and Tania Bedrax-Weiss and Oliver Wang and Ya Xu and Ollie Purkiss and Uri Mendlovic and Ilaï Deutel and Nam Nguyen and Adam Langley and Flip Korn and Lucia Rossazza and Alexandre Ramé and Sagar Waghmare and Helen Miller and Nathan Byrd and Ashrith Sheshan and Raia Hadsell and Sangnie Bhardwaj and Pawel Janus and Tero Rissa and Dan Horgan and Alvin Abdagic and Lior Belenki and James Allingham and Anima Singh and Theo Guidroz and Srivatsan Srinivasan and Herman Schmit and Kristen Chiafullo and Andre Elisseeff and Nilpa Jha and Prateek Kolhar and Leonard Berrada and Frank Ding and Xiance Si and Shrestha Basu Mallick and Franz Och and Sofia Erell and Eric Ni and Tejasi Latkar and Sherry Yang and Petar Sirkovic and Ziqiang Feng and Robert Leland and Rachel Hornung and Gang Wu and Charles Blundell and Hamidreza Alvari and Po-Sen Huang and Cathy Yip and Sanja Deur and Li Liu and Gabriela Surita and Pablo Duque and Dima Damen and Johnson Jia and Arthur Guez and Markus Mircea and Animesh Sinha and Alberto Magni and Paweł Stradomski and Tal Marian and Vlado Galić and Wenhu Chen and Hisham Husain and Achintya Singhal and Dominik Grewe and François-Xavier Aubet and Shuang Song and Lorenzo Blanco and Leland Rechis and Lewis Ho and Rich Munoz and Kelvin Zheng and Jessica Hamrick and Kevin Mather and Hagai Taitelbaum and Eliza Rutherford and Yun Lei and Kuangyuan Chen and Anand Shukla and Erica Moreira and Eric Doi and Berivan Isik and Nir Shabat and Dominika Rogozińska and Kashyap Kolipaka and Jason Chang and Eugen Vušak and Srinivasan Venkatachary and Shadi Noghabi and Tarun Bharti and Younghoon Jun and Aleksandr Zaks and Simon Green and Jeshwanth Challagundla and William Wong and Muqthar Mohammad and Dean Hirsch and Yong Cheng and Iftekhar Naim and Lev Proleev and Damien Vincent and Aayush Singh and Maxim Krikun and Dilip Krishnan and Zoubin Ghahramani and Aviel Atias and Rajeev Aggarwal and Christo Kirov and Dimitrios Vytiniotis and Christy Koh and Alexandra Chronopoulou and Pawan Dogra and Vlad-Doru Ion and Gladys Tyen and Jason Lee and Felix Weissenberger and Trevor Strohman and Ashwin Balakrishna and Jack Rae and Marko Velic and Raoul de Liedekerke and Oded Elyada and Wentao Yuan and Canoee Liu and Lior Shani and Sergey Kishchenko and Bea Alessio and Yandong Li and Richard Song and Sam Kwei and Orion Jankowski and Aneesh Pappu and Youhei Namiki and Yenai Ma and Nilesh Tripuraneni and Colin Cherry and Marissa Ikonomidis and Yu-Cheng Ling and Colin Ji and Beka Westberg and Auriel Wright and Da Yu and David Parkinson and Swaroop Ramaswamy and Jerome Connor and Soheil Hassas Yeganeh and Snchit Grover and George Kenwright and Lubo Litchev and Chris Apps and Alex Tomala and Felix Halim and Alex Castro-Ros and Zefei Li and Anudhyan Boral and Pauline Sho and Michal Yarom and Eric Malmi and David Klinghoffer and Rebecca Lin and Alan Ansell and Pradeep Kumar S and Shubin Zhao and Siqi Zuo and Adam Santoro and Heng-Tze Cheng and Solomon Demmessie and Yuchi Liu and Nicole Brichtova and Allie Culp and Nathaniel Braun and Dan Graur and Will Ng and Nikhil Mehta and Aaron Phillips and Patrik Sundberg and Varun Godbole and Fangyu Liu and Yash Katariya and David Rim and Mojtaba Seyedhosseini and Sean Ammirati and Jonas Valfridsson and Mahan Malihi and Timothy Knight and Andeep Toor and Thomas Lampe and Abe Ittycheriah and Lewis Chiang and Chak Yeung and Alexandre Fréchette and Jinmeng Rao and Huisheng Wang and Himanshu Srivastava and Richard Zhang and Rocky Rhodes and Ariel Brand and Dean Weesner and Ilya Figotin and Felix Gimeno and Rachana Fellinger and Pierre Marcenac and José Leal and Eyal Marcus and Victor Cotruta and Rodrigo Cabrera and Sheryl Luo and Dan Garrette and Vera Axelrod and Sorin Baltateanu and David Barker and Dongkai Chen and Horia Toma and Ben Ingram and Jason Riesa and Chinmay Kulkarni and Yujing Zhang and Hongbin Liu and Chao Wang and Martin Polacek and Will Wu and Kai Hui and Adrian N Reyes and Yi Su and Megan Barnes and Ishaan Malhi and Anfal Siddiqui and Qixuan Feng and Mihai Damaschin and Daniele Pighin and Andreas Steiner and Samuel Yang and Ramya Sree Boppana and Simeon Ivanov and Arun Kandoor and Aditya Shah and Asier Mujika and Da Huang and Christopher A. Choquette-Choo and Mohak Patel and Tianhe Yu and Toni Creswell and Jerry and Liu and Catarina Barros and Yasaman Razeghi and Aurko Roy and Phil Culliton and Binbin Xiong and Jiaqi Pan and Thomas Strohmann and Tolly Powell and Babi Seal and Doug DeCarlo and Pranav Shyam and Kaan Katircioglu and Xuezhi Wang and Cassidy Hardin and Immanuel Odisho and Josef Broder and Oscar Chang and Arun Nair and Artem Shtefan and Maura O'Brien and Manu Agarwal and Sahitya Potluri and Siddharth Goyal and Amit Jhindal and Saksham Thakur and Yury Stuken and James Lyon and Kristina Toutanova and Fangxiaoyu Feng and Austin Wu and Ben Horn and Alek Wang and Alex Cullum and Gabe Taubman and Disha Shrivastava and Chongyang Shi and Hamish Tomlinson and Roma Patel and Tao Tu and Ada Maksutaj Oflazer and Francesco Pongetti and Mingyao Yang and Adrien Ali Taïga and Vincent Perot and Nuo Wang Pierse and Feng Han and Yoel Drori and Iñaki Iturrate and Ayan Chakrabarti and Legg Yeung and Dave Dopson and Yi-ting Chen and Apoorv Kulshreshtha and Tongfei Guo and Philip Pham and Tal Schuster and Junquan Chen and Alex Polozov and Jinwei Xing and Huanjie Zhou and Praneeth Kacham and Doron Kukliansky and Antoine Miech and Sergey Yaroshenko and Ed Chi and Sholto Douglas and Hongliang Fei and Mathieu Blondel and Preethi Myla and Lior Madmoni and Xing Wu and Daniel Keysers and Kristian Kjems and Isabela Albuquerque and Lijun Yu and Joel D'sa and Michelle Plantan and Vlad Ionescu and Jaume Sanchez Elias and Abhirut Gupta and Manish Reddy Vuyyuru and Fred Alcober and Tong Zhou and Kaiyang Ji and Florian Hartmann and Subha Puttagunta and Hugo Song and Ehsan Amid and Anca Stefanoiu and Andrew Lee and Paul Pucciarelli and Emma Wang and Amit Raul and Slav Petrov and Isaac Tian and Valentin Anklin and Nana Nti and Victor Gomes and Max Schumacher and Grace Vesom and Alex Panagopoulos and Konstantinos Bousmalis and Daniel Andor and Josh Jacob and Yuan Zhang and Bill Rosgen and Matija Kecman and Matthew Tung and Alexandra Belias and Noah Goodman and Paul Covington and Brian Wieder and Nikita Saxena and Elnaz Davoodi and Muhuan Huang and Sharath Maddineni and Vincent Roulet and Folawiyo Campbell-Ajala and Pier Giuseppe Sessa and Xintian and Wu and Guangda Lai and Paul Collins and Alex Haig and Vytenis Sakenas and Xiaowei Xu and Marissa Giustina and Laurent El Shafey and Pichi Charoenpanit and Shefali Garg and Joshua Ainslie and Boone Severson and Montse Gonzalez Arenas and Shreya Pathak and Sujee Rajayogam and Jie Feng and Michiel Bakker and Sheng Li and Nevan Wichers and Jamie Rogers and Xinyang Geng and Yeqing Li and Rolf Jagerman and Chao Jia and Nadav Olmert and David Sharon and Matthew Mauger and Sandeep Mariserla and Hongxu Ma and Megha Mohabey and Kyuyeun Kim and Alek Andreev and Scott Pollom and Juliette Love and Vihan Jain and Priyanka Agrawal and Yannick Schroecker and Alisa Fortin and Manfred Warmuth and Ji Liu and Andrew Leach and Irina Blok and Ganesh Poomal Girirajan and Roee Aharoni and Benigno Uria and Andrei Sozanschi and Dan Goldberg and Lucian Ionita and Marco Tulio Ribeiro and Martin Zlocha and Vighnesh Birodkar and Sami Lachgar and Liangzhe Yuan and Himadri Choudhury and Matt Ginsberg and Fei Zheng and Gregory Dibb and Emily Graves and Swachhand Lokhande and Gabriel Rasskin and George-Cristian Muraru and Corbin Quick and Sandeep Tata and Pierre Sermanet and Aditya Chawla and Itay Karo and Yan Wang and Susan Zhang and Orgad Keller and Anca Dragan and Guolong Su and Ian Chou and Xi Liu and Yiqing Tao and Shruthi Prabhakara and Marc Wilson and Ruibo Liu and Shibo Wang and Georgie Evans and David Du and Alfonso Castaño and Gautam Prasad and Mona El Mahdy and Sebastian Gerlach and Machel Reid and Jarrod Kahn and Amir Zait and Thanumalayan Sankaranarayana Pillai and Thatcher Ulrich and Guanyu Wang and Jan Wassenberg and Efrat Farkash and Kiran Yalasangi and Congchao Wang and Maria Bauza and Simon Bucher and Ting Liu and Jun Yan and Gary Leung and Vikas Sindhwani and Parker Barnes and Avi Singh and Ivan Jurin and Jichuan Chang and Niket Kumar Bhumihar and Sivan Eiger and Gui Citovsky and Ben Withbroe and Zhang Li and Siyang Xue and Niccolò Dal Santo and Georgi Stoyanov and Yves Raimond and Steven Zheng and Yilin Gao and Vít Listík and Sławek Kwasiborski and Rachel Saputro and Adnan Ozturel and Ganesh Mallya and Kushal Majmundar and Ross West and Paul Caron and Jinliang Wei and Lluis Castrejon and Sharad Vikram and Deepak Ramachandran and Nikhil Dhawan and Jiho Park and Sara Smoot and George van den Driessche and Yochai Blau and Chase Malik and Wei Liang and Roy Hirsch and Cicero Nogueira dos Santos and Eugene Weinstein and Aäron van den Oord and Sid Lall and Nicholas FitzGerald and Zixuan Jiang and Xuan Yang and Dale Webster and Ali Elqursh and Aedan Pope and Georges Rotival and David Raposo and Wanzheng Zhu and Jeff Dean and Sami Alabed and Dustin Tran and Arushi Gupta and Zach Gleicher and Jessica Austin and Edouard Rosseel and Megh Umekar and Dipanjan Das and Yinghao Sun and Kai Chen and Karolis Misiunas and Xiang Zhou and Yixian Di and Alyssa Loo and Josh Newlan and Bo Li and Vinay Ramasesh and Ying Xu and Alex Chen and Sudeep Gandhe and Radu Soricut and Nikita Gupta and Shuguang Hu and Seliem El-Sayed and Xavier Garcia and Idan Brusilovsky and Pu-Chin Chen and Andrew Bolt and Lu Huang and Alex Gurney and Zhiying Zhang and Alexander Pritzel and Jarek Wilkiewicz and Bryan Seybold and Bhargav Kanagal Shamanna and Felix Fischer and Josef Dean and Karan Gill and Ross Mcilroy and Abhishek Bhowmick and Jeremy Selier and Antoine Yang and Derek Cheng and Vladimir Magay and Jie Tan and Dhriti Varma and Christian Walder and Tomas Kocisky and Ryo Nakashima and Paul Natsev and Mike Kwong and Ionel Gog and Chiyuan Zhang and Sander Dieleman and Thomas Jimma and Andrey Ryabtsev and Siddhartha Brahma and David Steiner and Dayou Du and Ante Žužul and Mislav Žanić and Mukund Raghavachari and Willi Gierke and Zeyu Zheng and Dessie Petrova and Yann Dauphin and Yuchuan Liu and Ido Kessler and Steven Hand and Chris Duvarney and Seokhwan Kim and Hyo Lee and Léonard Hussenot and Jeffrey Hui and Josh Smith and Deepali Jain and Jiawei Xia and Gaurav Singh Tomar and Keyvan Amiri and Du Phan and Fabian Fuchs and Tobias Weyand and Nenad Tomasev and Alexandra Cordell and Xin Liu and Jonathan Mallinson and Pankaj Joshi and Andy Crawford and Arun Suggala and Steve Chien and Nick Fernando and Mariella Sanchez-Vargas and Duncan Williams and Phil Crone and Xiyang Luo and Igor Karpov and Jyn Shan and Terry Thurk and Robin Strudel and Paul Voigtlaender and Piyush Patil and Tim Dozat and Ali Khodaei and Sahil Singla and Piotr Ambroszczyk and Qiyin Wu and Yifan Chang and Brian Roark and Chaitra Hegde and Tianli Ding and Angelos Filos and Zhongru Wu and André Susano Pinto and Shuang Liu and Saarthak Khanna and Aditya Pandey and Siobhan Mcloughlin and Qiujia Li and Sam Haves and Allan Zhou and Elena Buchatskaya and Isabel Leal and Peter de Boursac and Nami Akazawa and Nina Anderson and Terry Chen and Krishna Somandepalli and Chen Liang and Sheela Goenka and Stephanie Winkler and Alexander Grushetsky and Yifan Ding and Jamie Smith and Fan Ye and Jordi Pont-Tuset and Eric Li and Ruichao Li and Tomer Golany and Dawid Wegner and Tao Jiang and Omer Barak and Yuan Shangguan and Eszter Vértes and Renee Wong and Jörg Bornschein and Alex Tudor and Michele Bevilacqua and Tom Schaul and Ankit Singh Rawat and Yang Zhao and Kyriakos Axiotis and Lei Meng and Cory McLean and Jonathan Lai and Jennifer Beattie and Nate Kushman and Yaxin Liu and Blair Kutzman and Fiona Lang and Jingchen Ye and Praneeth Netrapalli and Pushkar Mishra and Myriam Khan and Megha Goel and Rob Willoughby and David Tian and Honglei Zhuang and JD Chen and Zak Tsai and Tasos Kementsietsidis and Arjun Khare and James Keeling and Keyang Xu and Nathan Waters and Florent Altché and Ashok Popat and Bhavishya Mittal and David Saxton and Dalia El Badawy and Michael Mathieu and Zheng Zheng and Hao Zhou and Nishant Ranka and Richard Shin and Qingnan Duan and Tim Salimans and Ioana Mihailescu and Uri Shaham and Ming-Wei Chang and Yannis Assael and Nishanth Dikkala and Martin Izzard and Vincent Cohen-Addad and Cat Graves and Vlad Feinberg and Grace Chung and DJ Strouse and Danny Karmon and Sahand Sharifzadeh and Zoe Ashwood and Khiem Pham and Jon Blanton and Alex Vasiloff and Jarred Barber and Mark Geller and Aurick Zhou and Fedir Zubach and Tzu-Kuo Huang and Lei Zhang and Himanshu Gupta and Matt Young and Julia Proskurnia and Ronny Votel and Valentin Gabeur and Gabriel Barcik and Aditya Tripathi and Hongkun Yu and Geng Yan and Beer Changpinyo and Filip Pavetić and Amy Coyle and Yasuhisa Fujii and Jorge Gonzalez Mendez and Tianhao Zhou and Harish Rajamani and Blake Hechtman and Eddie Cao and Da-Cheng Juan and Yi-Xuan Tan and Valentin Dalibard and Yilun Du and Natalie Clay and Kaisheng Yao and Wenhao Jia and Dimple Vijaykumar and Yuxiang Zhou and Xinyi Bai and Wei-Chih Hung and Steven Pecht and Georgi Todorov and Nikhil Khadke and Pramod Gupta and Preethi Lahoti and Arnaud Autef and Karthik Duddu and James Lee-Thorp and Alexander Bykovsky and Tautvydas Misiunas and Sebastian Flennerhag and Santhosh Thangaraj and Jed McGiffin and Zack Nado and Markus Kunesch and Andreas Noever and Amir Hertz and Marco Liang and Victor Stone and Evan Palmer and Samira Daruki and Arijit Pramanik and Siim Põder and Austin Kyker and Mina Khan and Evgeny Sluzhaev and Marvin Ritter and Avraham Ruderman and Wenlei Zhou and Chirag Nagpal and Kiran Vodrahalli and George Necula and Paul Barham and Ellie Pavlick and Jay Hartford and Izhak Shafran and Long Zhao and Maciej Mikuła and Tom Eccles and Hidetoshi Shimokawa and Kanav Garg and Luke Vilnis and Hanwen Chen and Ilia Shumailov and Kuang-Huei Lee and Abdelrahman Abdelhamed and Meiyan Xie and Vered Cohen and Ester Hlavnova and Dan Malkin and Chawin Sitawarin and James Lottes and Pauline Coquinot and Tianli Yu and Sandeep Kumar and Jingwei Zhang and Aroma Mahendru and Zafarali Ahmed and James Martens and Tao Chen and Aviel Boag and Daiyi Peng and Coline Devin and Arseniy Klimovskiy and Mary Phuong and Danny Vainstein and Jin Xie and Bhuvana Ramabhadran and Nathan Howard and Xinxin Yu and Gitartha Goswami and Jingyu Cui and Sam Shleifer and Mario Pinto and Chih-Kuan Yeh and Ming-Hsuan Yang and Sara Javanmardi and Dan Ethier and Chace Lee and Jordi Orbay and Suyog Kotecha and Carla Bromberg and Pete Shaw and James Thornton and Adi Gerzi Rosenthal and Shane Gu and Matt Thomas and Ian Gemp and Aditya Ayyar and Asahi Ushio and Aarush Selvan and Joel Wee and Chenxi Liu and Maryam Majzoubi and Weiren Yu and Jake Abernethy and Tyler Liechty and Renke Pan and Hoang Nguyen and Qiong and Hu and Sarah Perrin and Abhinav Arora and Emily Pitler and Weiyi Wang and Kaushik Shivakumar and Flavien Prost and Ben Limonchik and Jing Wang and Yi Gao and Timothee Cour and Shyamal Buch and Huan Gui and Maria Ivanova and Philipp Neubeck and Kelvin Chan and Lucy Kim and Huizhong Chen and Naman Goyal and Da-Woon Chung and Lu Liu and Yao Su and Anastasia Petrushkina and Jiajun Shen and Armand Joulin and Yuanzhong Xu and Stein Xudong Lin and Yana Kulizhskaya and Ciprian Chelba and Shobha Vasudevan and Eli Collins and Vasilisa Bashlovkina and Tony Lu and Doug Fritz and Jongbin Park and Yanqi Zhou and Chen Su and Richard Tanburn and Mikhail Sushkov and Mitchelle Rasquinha and Jinning Li and Jennifer Prendki and Yiming Li and Pallavi LV and Shriya Sharma and Hen Fitoussi and Hui Huang and Andrew Dai and Phuong Dao and Mike Burrows and Henry Prior and Danfeng Qin and Golan Pundak and Lars Lowe Sjoesund and Art Khurshudov and Zhenkai Zhu and Albert Webson and Elizabeth Kemp and Tat Tan and Saurabh Agrawal and Susie Sargsyan and Liqun Cheng and Jim Stephan and Tom Kwiatkowski and David Reid and Arunkumar Byravan and Assaf Hurwitz Michaely and Nicolas Heess and Luowei Zhou and Sonam Goenka and Viral Carpenter and Anselm Levskaya and Bo Wang and Reed Roberts and Rémi Leblond and Sharat Chikkerur and Stav Ginzburg and Max Chang and Robert Riachi and Chuqiao and Xu and Zalán Borsos and Michael Pliskin and Julia Pawar and Morgane Lustman and Hannah Kirkwood and Ankit Anand and Aditi Chaudhary and Norbert Kalb and Kieran Milan and Sean Augenstein and Anna Goldie and Laurel Prince and Karthik Raman and Yanhua Sun and Vivian Xia and Aaron Cohen and Zhouyuan Huo and Josh Camp and Seher Ellis and Lukas Zilka and David Vilar Torres and Lisa Patel and Sho Arora and Betty Chan and Jonas Adler and Kareem Ayoub and Jacky Liang and Fayaz Jamil and Jiepu Jiang and Simon Baumgartner and Haitian Sun and Yael Karov and Yaroslav Akulov and Hui Zheng and Irene Cai and Claudio Fantacci and James Rubin and Alex Rav Acha and Mengchao Wang and Nina D'Souza and Rohit Sathyanarayana and Shengyang Dai and Simon Rowe and Andrey Simanovsky and Omer Goldman and Yuheng Kuang and Xiaoyue Pan and Andrew Rosenberg and Tania Rojas-Esponda and Praneet Dutta and Amy Zeng and Irina Jurenka and Greg Farquhar and Yamini Bansal and Shariq Iqbal and Becca Roelofs and Ga-Young Joung and Parker Beak and Changwan Ryu and Ryan Poplin and Yan Wu and Jean-Baptiste Alayrac and Senaka Buthpitiya and Olaf Ronneberger and Caleb Habtegebriel and Wei Li and Paul Cavallaro and Aurora Wei and Guy Bensky and Timo Denk and Harish Ganapathy and Jeff Stanway and Pratik Joshi and Francesco Bertolini and Jessica Lo and Olivia Ma and Zachary Charles and Geta Sampemane and Himanshu Sahni and Xu Chen and Harry Askham and David Gaddy and Peter Young and Jiewen Tan and Matan Eyal and Arthur Bražinskas and Li Zhong and Zhichun Wu and Mark Epstein and Kai Bailey and Andrew Hard and Kamyu Lee and Sasha Goldshtein and Alex Ruiz and Mohammed Badawi and Matthias Lochbrunner and JK Kearns and Ashley Brown and Fabio Pardo and Theophane Weber and Haichuan Yang and Pan-Pan Jiang and Berkin Akin and Zhao Fu and Marcus Wainwright and Chi Zou and Meenu Gaba and Pierre-Antoine Manzagol and Wendy Kan and Yang Song and Karina Zainullina and Rui Lin and Jeongwoo Ko and Salil Deshmukh and Apoorv Jindal and James Svensson and Divya Tyam and Heri Zhao and Christine Kaeser-Chen and Scott Baird and Pooya Moradi and Jamie Hall and Qiuchen Guo and Vincent Tsang and Bowen Liang and Fernando Pereira and Suhas Ganesh and Ivan Korotkov and Jakub Adamek and Sridhar Thiagarajan and Vinh Tran and Charles Chen and Chris Tar and Sanil Jain and Ishita Dasgupta and Taylan Bilal and David Reitter and Kai Zhao and Giulia Vezzani and Yasmin Gehman and Pulkit Mehta and Lauren Beltrone and Xerxes Dotiwalla and Sergio Guadarrama and Zaheer Abbas and Stefani Karp and Petko Georgiev and Chun-Sung Ferng and Marc Brockschmidt and Liqian Peng and Christoph Hirnschall and Vikas Verma and Yingying Bi and Ying Xiao and Avigail Dabush and Kelvin Xu and Phil Wallis and Randall Parker and Qifei Wang and Yang Xu and Ilkin Safarli and Dinesh Tewari and Yin Zhang and Seungyeon Kim and Andrea Gesmundo and Mackenzie Thomas and Sergey Levi and Ahmed Chowdhury and Kanishka Rao and Peter Garst and Sam Conway-Rahman and Helen Ran and Kay McKinney and Zhisheng Xiao and Wenhao Yu and Rohan Agrawal and Axel Stjerngren and Catalin Ionescu and Jingjing Chen and Vivek Sharma and Justin Chiu and Fei Liu and Ken Franko and Clayton Sanford and Xingyu Cai and Paul Michel and Sanjay Ganapathy and Jane Labanowski and Zachary Garrett and Ben Vargas and Sean Sun and Bryan Gale and Thomas Buschmann and Guillaume Desjardins and Nimesh Ghelani and Palak Jain and Mudit Verma and Chulayuth Asawaroengchai and Julian Eisenschlos and Jitendra Harlalka and Hideto Kazawa and Don Metzler and Joshua Howland and Ying Jian and Jake Ades and Viral Shah and Tynan Gangwani and Seungji Lee and Roman Ring and Steven M. Hernandez and Dean Reich and Amer Sinha and Ashutosh Sathe and Joe Kovac and Ashleah Gill and Ajay Kannan and Andrea D'olimpio and Martin Sevenich and Jay Whang and Been Kim and Khe Chai Sim and Jilin Chen and Jiageng Zhang and Shuba Lall and Yossi Matias and Bill Jia and Abe Friesen and Sara Nasso and Ashish Thapliyal and Bryan Perozzi and Ting Yu and Anna Shekhawat and Safeen Huda and Peter Grabowski and Eric Wang and Ashwin Sreevatsa and Hilal Dib and Mehadi Hassen and Parker Schuh and Vedrana Milutinovic and Chris Welty and Michael Quinn and Ali Shah and Bangju Wang and Gabe Barth-Maron and Justin Frye and Natalie Axelsson and Tao Zhu and Yukun Ma and Irene Giannoumis and Hanie Sedghi and Chang Ye and Yi Luan and Kevin Aydin and Bilva Chandra and Vivek Sampathkumar and Ronny Huang and Victor Lavrenko and Ahmed Eleryan and Zhi Hong and Steven Hansen and Sara Mc Carthy and Bidisha Samanta and Domagoj Ćevid and Xin Wang and Fangtao Li and Michael Voznesensky and Matt Hoffman and Andreas Terzis and Vikash Sehwag and Gil Fidel and Luheng He and Mu Cai and Yanzhang He and Alex Feng and Martin Nikoltchev and Samrat Phatale and Jason Chase and Rory Lawton and Ming Zhang and Tom Ouyang and Manuel Tragut and Mehdi Hafezi Manshadi and Arjun Narayanan and Jiaming Shen and Xu Gao and Tolga Bolukbasi and Nick Roy and Xin Li and Daniel Golovin and Liviu Panait and Zhen Qin and Guangxing Han and Thomas Anthony and Sneha Kudugunta and Viorica Patraucean and Aniket Ray and Xinyun Chen and Xiaochen Yang and Tanuj Bhatia and Pranav Talluri and Alex Morris and Andrija Ražnatović and Bethanie Brownfield and James An and Sheng Peng and Patrick Kane and Ce Zheng and Nico Duduta and Joshua Kessinger and James Noraky and Siqi Liu and Keran Rong and Petar Veličković and Keith Rush and Alex Goldin and Fanny Wei and Shiva Mohan Reddy Garlapati and Caroline Pantofaru and Okwan Kwon and Jianmo Ni and Eric Noland and Julia Di Trapani and Françoise Beaufays and Abhijit Guha Roy and Yinlam Chow and Aybuke Turker and Geoffrey Cideron and Lantao Mei and Jon Clark and Qingyun Dou and Matko Bošnjak and Ralph Leith and Yuqing Du and Amir Yazdanbakhsh and Milad Nasr and Chester Kwak and Suraj Satishkumar Sheth and Alex Kaskasoli and Ankesh Anand and Balaji Lakshminarayanan and Sammy Jerome and David Bieber and Chun-Te Chu and Alexandre Senges and Tianxiao Shen and Mukund Sridhar and Ndaba Ndebele and Benjamin Beyret and Shakir Mohamed and Mia Chen and Markus Freitag and Jiaxian Guo and Luyang Liu and Paul Roit and Heng Chen and Shen Yan and Tom Stone and JD Co-Reyes and Jeremy Cole and Salvatore Scellato and Shekoofeh Azizi and Hadi Hashemi and Alicia Jin and Anand Iyer and Marcella Valentine and András György and Arun Ahuja and Daniel Hernandez Diaz and Chen-Yu Lee and Nathan Clement and Weize Kong and Drew Garmon and Ishaan Watts and Kush Bhatia and Khyatti Gupta and Matt Miecnikowski and Hugo Vallet and Ankur Taly and Edward Loper and Saket Joshi and James Atwood and Jo Chick and Mark Collier and Fotis Iliopoulos and Ryan Trostle and Beliz Gunel and Ramiro Leal-Cavazos and Arnar Mar Hrafnkelsson and Michael Guzman and Xiaoen Ju and Andy Forbes and Jesse Emond and Kushal Chauhan and Ben Caine and Li Xiao and Wenjun Zeng and Alexandre Moufarek and Daniel Murphy and Maya Meng and Nitish Gupta and Felix Riedel and Anil Das and Elijah Lawal and Shashi Narayan and Tiberiu Sosea and James Swirhun and Linda Friso and Behnam Neyshabur and Jing Lu and Sertan Girgin and Michael Wunder and Edouard Yvinec and Aroonalok Pyne and Victor Carbune and Shruti Rijhwani and Yang Guo and Tulsee Doshi and Anton Briukhov and Max Bain and Ayal Hitron and Xuanhui Wang and Ashish Gupta and Ke Chen and Cosmo Du and Weiyang Zhang and Dhruv Shah and Arjun Akula and Max Dylla and Ashyana Kachra and Weicheng Kuo and Tingting Zou and Lily Wang and Luyao Xu and Jifan Zhu and Justin Snyder and Sachit Menon and Orhan Firat and Igor Mordatch and Yuan Yuan and Natalia Ponomareva and Rory Blevins and Lawrence Moore and Weijun Wang and Phil Chen and Martin Scholz and Artur Dwornik and Jason Lin and Sicheng Li and Diego Antognini and Te I and Xiaodan Song and Matt Miller and Uday Kalra and Adam Raveret and Oscar Akerlund and Felix Wu and Andrew Nystrom and Namrata Godbole and Tianqi Liu and Hannah DeBalsi and Jewel Zhao and Buhuang Liu and Avi Caciularu and Lauren Lax and Urvashi Khandelwal and Victoria Langston and Eric Bailey and Silvio Lattanzi and Yufei Wang and Neel Kovelamudi and Sneha Mondal and Guru Guruganesh and Nan Hua and Ofir Roval and Paweł Wesołowski and Rishikesh Ingale and Jonathan Halcrow and Tim Sohn and Christof Angermueller and Bahram Raad and Eli Stickgold and Eva Lu and Alec Kosik and Jing Xie and Timothy Lillicrap and Austin Huang and Lydia Lihui Zhang and Dominik Paulus and Clement Farabet and Alex Wertheim and Bing Wang and Rishabh Joshi and Chu-ling Ko and Yonghui Wu and Shubham Agrawal and Lily Lin and XiangHai Sheng and Peter Sung and Tyler Breland-King and Christina Butterfield and Swapnil Gawde and Sumeet Singh and Qiao Zhang and Raj Apte and Shilpa Shetty and Adrian Hutter and Tao Li and Elizabeth Salesky and Federico Lebron and Jonni Kanerva and Michela Paganini and Arthur Nguyen and Rohith Vallu and Jan-Thorsten Peter and Sarmishta Velury and David Kao and Jay Hoover and Anna Bortsova and Colton Bishop and Shoshana Jakobovits and Alessandro Agostini and Alekh Agarwal and Chang Liu and Charles Kwong and Sasan Tavakkol and Ioana Bica and Alex Greve and Anirudh GP and Jake Marcus and Le Hou and Tom Duerig and Rivka Moroshko and Dave Lacey and Andy Davis and Julien Amelot and Guohui Wang and Frank Kim and Theofilos Strinopoulos and Hui Wan and Charline Le Lan and Shankar Krishnan and Haotian Tang and Peter Humphreys and Junwen Bai and Idan Heimlich Shtacher and Diego Machado and Chenxi Pang and Ken Burke and Dangyi Liu and Renga Aravamudhan and Yue Song and Ed Hirst and Abhimanyu Singh and Brendan Jou and Liang Bai and Francesco Piccinno and Chuyuan Kelly Fu and Robin Alazard and Barak Meiri and Daniel Winter and Charlie Chen and Mingda Zhang and Jens Heitkaemper and John Lambert and Jinhyuk Lee and Alexander Frömmgen and Sergey Rogulenko and Pranav Nair and Paul Niemczyk and Anton Bulyenov and Bibo Xu and Hadar Shemtov and Morteza Zadimoghaddam and Serge Toropov and Mateo Wirth and Hanjun Dai and Sreenivas Gollapudi and Daniel Zheng and Alex Kurakin and Chansoo Lee and Kalesha Bullard and Nicolas Serrano and Ivana Balazevic and Yang Li and Johan Schalkwyk and Mark Murphy and Mingyang Zhang and Kevin Sequeira and Romina Datta and Nishant Agrawal and Charles Sutton and Nithya Attaluri and Mencher Chiang and Wael Farhan and Gregory Thornton and Kate Lin and Travis Choma and Hung Nguyen and Kingshuk Dasgupta and Dirk Robinson and Iulia Comşa and Michael Riley and Arjun Pillai and Basil Mustafa and Ben Golan and Amir Zandieh and Jean-Baptiste Lespiau and Billy Porter and David Ross and Sujeevan Rajayogam and Mohit Agarwal and Subhashini Venugopalan and Bobak Shahriari and Qiqi Yan and Hao Xu and Taylor Tobin and Pavel Dubov and Hongzhi Shi and Adrià Recasens and Anton Kovsharov and Sebastian Borgeaud and Lucio Dery and Shanthal Vasanth and Elena Gribovskaya and Linhai Qiu and Mahdis Mahdieh and Wojtek Skut and Elizabeth Nielsen and CJ Zheng and Adams Yu and Carrie Grimes Bostock and Shaleen Gupta and Aaron Archer and Chris Rawles and Elinor Davies and Alexey Svyatkovskiy and Tomy Tsai and Yoni Halpern and Christian Reisswig and Bartek Wydrowski and Bo Chang and Joan Puigcerver and Mor Hazan Taege and Jian Li and Eva Schnider and Xinjian Li and Dragos Dena and Yunhan Xu and Umesh Telang and Tianze Shi and Heiga Zen and Kyle Kastner and Yeongil Ko and Neesha Subramaniam and Aviral Kumar and Pete Blois and Zhuyun Dai and John Wieting and Yifeng Lu and Yoel Zeldes and Tian Xie and Anja Hauth and Alexandru Ţifrea and Yuqi Li and Sam El-Husseini and Dan Abolafia and Howard Zhou and Wen Ding and Sahra Ghalebikesabi and Carlos Guía and Andrii Maksai and Ágoston Weisz and Sercan Arik and Nick Sukhanov and Aga Świetlik and Xuhui Jia and Luo Yu and Weiyue Wang and Mark Brand and Dawn Bloxwich and Sean Kirmani and Zhe Chen and Alec Go and Pablo Sprechmann and Nithish Kannen and Alen Carin and Paramjit Sandhu and Isabel Edkins and Leslie Nooteboom and Jai Gupta and Loren Maggiore and Javad Azizi and Yael Pritch and Pengcheng Yin and Mansi Gupta and Danny Tarlow and Duncan Smith and Desi Ivanov and Mohammad Babaeizadeh and Ankita Goel and Satish Kambala and Grace Chu and Matej Kastelic and Michelle Liu and Hagen Soltau and Austin Stone and Shivani Agrawal and Min Kim and Kedar Soparkar and Srinivas Tadepalli and Oskar Bunyan and Rachel Soh and Arvind Kannan and DY Kim and Blake JianHang Chen and Afief Halumi and Sudeshna Roy and Yulong Wang and Olcan Sercinoglu and Gena Gibson and Sijal Bhatnagar and Motoki Sano and Daniel von Dincklage and Qingchun Ren and Blagoj Mitrevski and Mirek Olšák and Jennifer She and Carl Doersch and Jilei and Wang and Bingyuan Liu and Qijun Tan and Tamar Yakar and Tris Warkentin and Alex Ramirez and Carl Lebsack and Josh Dillon and Rajiv Mathews and Tom Cobley and Zelin Wu and Zhuoyuan Chen and Jon Simon and Swaroop Nath and Tara Sainath and Alexei Bendebury and Ryan Julian and Bharath Mankalale and Daria Ćurko and Paulo Zacchello and Adam R. Brown and Kiranbir Sodhia and Heidi Howard and Sergi Caelles and Abhinav Gupta and Gareth Evans and Anna Bulanova and Lesley Katzen and Roman Goldenberg and Anton Tsitsulin and Joe Stanton and Benoit Schillings and Vitaly Kovalev and Corey Fry and Rushin Shah and Kuo Lin and Shyam Upadhyay and Cheng Li and Soroush Radpour and Marcello Maggioni and Jing Xiong and Lukas Haas and Jenny Brennan and Aishwarya Kamath and Nikolay Savinov and Arsha Nagrani and Trevor Yacovone and Ryan Kappedal and Kostas Andriopoulos and Li Lao and YaGuang Li and Grigory Rozhdestvenskiy and Kazuma Hashimoto and Andrew Audibert and Sophia Austin and Daniel Rodriguez and Anian Ruoss and Garrett Honke and Deep Karkhanis and Xi Xiong and Qing Wei and James Huang and Zhaoqi Leng and Vittal Premachandran and Stan Bileschi and Georgios Evangelopoulos and Thomas Mensink and Jay Pavagadhi and Denis Teplyashin and Paul Chang and Linting Xue and Garrett Tanzer and Sally Goldman and Kaushal Patel and Shixin Li and Jeremy Wiesner and Ivy Zheng and Ian Stewart-Binks and Jie Han and Zhi Li and Liangchen Luo and Karel Lenc and Mario Lučić and Fuzhao Xue and Ryan Mullins and Alexey Guseynov and Chung-Ching Chang and Isaac Galatzer-Levy and Adam Zhang and Garrett Bingham and Grace Hu and Ale Hartman and Yue Ma and Jordan Griffith and Alex Irpan and Carey Radebaugh and Summer Yue and Lijie Fan and Victor Ungureanu and Christina Sorokin and Hannah Teufel and Peiran Li and Rohan Anil and Dimitris Paparas and Todd Wang and Chu-Cheng Lin and Hui Peng and Megan Shum and Goran Petrovic and Demetra Brady and Richard Nguyen and Klaus Macherey and Zhihao Li and Harman Singh and Madhavi Yenugula and Mariko Iinuma and Xinyi Chen and Kavya Kopparapu and Alexey Stern and Shachi Dave and Chandu Thekkath and Florence Perot and Anurag Kumar and Fangda Li and Yang Xiao and Matthew Bilotti and Mohammad Hossein Bateni and Isaac Noble and Lisa Lee and Amelio Vázquez-Reina and Julian Salazar and Xiaomeng Yang and Boyu Wang and Ela Gruzewska and Anand Rao and Sindhu Raghuram and Zheng Xu and Eyal Ben-David and Jieru Mei and Sid Dalmia and Zhaoyi Zhang and Yuchen Liu and Gagan Bansal and Helena Pankov and Steven Schwarcz and Andrea Burns and Christine Chan and Sumit Sanghai and Ricky Liang and Ethan Liang and Antoine He and Amy Stuart and Arun Narayanan and Yukun Zhu and Christian Frank and Bahar Fatemi and Amit Sabne and Oran Lang and Indro Bhattacharya and Shane Settle and Maria Wang and Brendan McMahan and Andrea Tacchetti and Livio Baldini Soares and Majid Hadian and Serkan Cabi and Timothy Chung and Nikita Putikhin and Gang Li and Jeremy Chen and Austin Tarango and Henryk Michalewski and Mehran Kazemi and Hussain Masoom and Hila Sheftel and Rakesh Shivanna and Archita Vadali and Ramona Comanescu and Doug Reid and Joss Moore and Arvind Neelakantan and Michaël Sander and Jonathan Herzig and Aviv Rosenberg and Mostafa Dehghani and JD Choi and Michael Fink and Reid Hayes and Eric Ge and Shitao Weng and Chia-Hua Ho and John Karro and Kalpesh Krishna and Lam Nguyen Thiet and Amy Skerry-Ryan and Daniel Eppens and Marco Andreetto and Navin Sarma and Silvano Bonacina and Burcu Karagol Ayan and Megha Nawhal and Zhihao Shan and Mike Dusenberry and Shantanu Thakoor and Sagar Gubbi and Duc Dung Nguyen and Reut Tsarfaty and Samuel Albanie and Jovana Mitrović and Meet Gandhi and Bo-Juen Chen and Alessandro Epasto and Georgi Stephanov and Ye Jin and Samuel Gehman and Aida Amini and Jack Weber and Feryal Behbahani and Shawn Xu and Miltos Allamanis and Xi Chen and Myle Ott and Claire Sha and Michal Jastrzebski and Hang Qi and David Greene and Xinyi Wu and Abodunrinwa Toki and Daniel Vlasic and Jane Shapiro and Ragha Kotikalapudi and Zhe Shen and Takaaki Saeki and Sirui Xie and Albin Cassirer and Shikhar Bharadwaj and Tatsuya Kiyono and Srinadh Bhojanapalli and Elan Rosenfeld and Sam Ritter and Jieming Mao and João Gabriel Oliveira and Zoltan Egyed and Bernd Bandemer and Emilio Parisotto and Keisuke Kinoshita and Juliette Pluto and Petros Maniatis and Steve Li and Yaohui Guo and Golnaz Ghiasi and Jean Tarbouriech and Srimon Chatterjee and Julie Jin and Katrina and Xu and Jennimaria Palomaki and Séb Arnold and Madhavi Sewak and Federico Piccinini and Mohit Sharma and Ben Albrecht and Sean Purser-haskell and Ashwin Vaswani and Chongyan Chen and Matheus Wisniewski and Qin Cao and John Aslanides and Nguyet Minh Phu and Maximilian Sieb and Lauren Agubuzu and Anne Zheng and Daniel Sohn and Marco Selvi and Anders Andreassen and Krishan Subudhi and Prem Eruvbetine and Oliver Woodman and Tomas Mery and Sebastian Krause and Xiaoqi Ren and Xiao Ma and Jincheng Luo and Dawn Chen and Wei Fan and Henry Griffiths and Christian Schuler and Alice Li and Shujian Zhang and Jean-Michel Sarr and Shixin Luo and Riccardo Patana and Matthew Watson and Dani Naboulsi and Michael Collins and Sailesh Sidhwani and Emiel Hoogeboom and Sharon Silver and Emily Caveness and Xiaokai Zhao and Mikel Rodriguez and Maxine Deines and Libin Bai and Patrick Griffin and Marco Tagliasacchi and Emily Xue and Spandana Raj Babbula and Bo Pang and Nan Ding and Gloria Shen and Elijah Peake and Remi Crocker and Shubha Srinivas Raghvendra and Danny Swisher and Woohyun Han and Richa Singh and Ling Wu and Vladimir Pchelin and Tsendsuren Munkhdalai and Dana Alon and Geoff Bacon and Efren Robles and Jannis Bulian and Melvin Johnson and George Powell and Felipe Tiengo Ferreira and Yaoyiran Li and Frederik Benzing and Mihajlo Velimirović and Hubert Soyer and William Kong and Tony and Nguyên and Zhen Yang and Jeremiah Liu and Joost van Amersfoort and Daniel Gillick and Baochen Sun and Nathalie Rauschmayr and Katie Zhang and Serena Zhan and Tao Zhou and Alexey Frolov and Chengrun Yang and Denis Vnukov and Louis Rouillard and Hongji Li and Amol Mandhane and Nova Fallen and Rajesh Venkataraman and Clara Huiyi Hu and Jennifer Brennan and Jenny Lee and Jerry Chang and Martin Sundermeyer and Zhufeng Pan and Rosemary Ke and Simon Tong and Alex Fabrikant and William Bono and Jindong Gu and Ryan Foley and Yiran Mao and Manolis Delakis and Dhruva Bhaswar and Roy Frostig and Nick Li and Avital Zipori and Cath Hope and Olga Kozlova and Swaroop Mishra and Josip Djolonga and Craig Schiff and Majd Al Merey and Eleftheria Briakou and Peter Morgan and Andy Wan and Avinatan Hassidim and RJ Skerry-Ryan and Kuntal Sengupta and Mary Jasarevic and Praveen Kallakuri and Paige Kunkle and Hannah Brennan and Tom Lieber and Hassan Mansoor and Julian Walker and Bing Zhang and Annie Xie and Goran Žužić and Adaeze Chukwuka and Alex Druinsky and Donghyun Cho and Rui Yao and Ferjad Naeem and Shiraz Butt and Eunyoung Kim and Zhipeng Jia and Mandy Jordan and Adam Lelkes and Mark Kurzeja and Sophie Wang and James Zhao and Andrew Over and Abhishek Chakladar and Marcel Prasetya and Neha Jha and Sriram Ganapathy and Yale Cong and Prakash Shroff and Carl Saroufim and Sobhan Miryoosefi and Mohamed Hammad and Tajwar Nasir and Weijuan Xi and Yang Gao and Young Maeng and Ben Hora and Chin-Yi Cheng and Parisa Haghani and Yoad Lewenberg and Caden Lu and Martin Matysiak and Naina Raisinghani and Huiyu Wang and Lexi Baugher and Rahul Sukthankar and Minh Giang and John Schultz and Noah Fiedel and Minmin Chen and Cheng-Chun Lee and Tapomay Dey and Hao Zheng and Shachi Paul and Celine Smith and Andy Ly and Yicheng Wang and Rishabh Bansal and Bartek Perz and Susanna Ricco and Stasha Blank and Vaishakh Keshava and Deepak Sharma and Marvin Chow and Kunal Lad and Komal Jalan and Simon Osindero and Craig Swanson and Jacob Scott and Anastasija Ilić and Xiaowei Li and Siddhartha Reddy Jonnalagadda and Afzal Shama Soudagar and Yan Xiong and Bat-Orgil Batsaikhan and Daniel Jarrett and Naveen Kumar and Maulik Shah and Matt Lawlor and Austin Waters and Mark Graham and Rhys May and Sabela Ramos and Sandra Lefdal and Zeynep Cankara and Nacho Cano and Brendan O'Donoghue and Jed Borovik and Frederick Liu and Jordan Grimstad and Mahmoud Alnahlawi and Katerina Tsihlas and Tom Hudson and Nikolai Grigorev and Yiling Jia and Terry Huang and Tobenna Peter Igwe and Sergei Lebedev and Xiaodan Tang and Igor Krivokon and Frankie Garcia and Melissa Tan and Eric Jia and Peter Stys and Shikhar Vashishth and Yu Liang and Balaji Venkatraman and Chenjie Gu and Anastasios Kementsietsidis and Chen Zhu and Junehyuk Jung and Yunfei Bai and Mohammad Javad Hosseini and Faruk Ahmed and Aditya Gupta and Xin Yuan and Shereen Ashraf and Shitij Nigam and Gautam Vasudevan and Pranjal Awasthi and Adi Mayrav Gilady and Zelda Mariet and Ramy Eskander and Haiguang Li and Hexiang Hu and Guillermo Garrido and Philippe Schlattner and George Zhang and Rohun Saxena and Petar Dević and Kritika Muralidharan and Ashwin Murthy and Yiqian Zhou and Min Choi and Arissa Wongpanich and Zhengdong Wang and Premal Shah and Yuntao Xu and Yiling Huang and Stephen Spencer and Alice Chen and James Cohan and Junjie Wang and Jonathan Tompson and Junru Wu and Ruba Haroun and Haiqiong Li and Blanca Huergo and Fan Yang and Tongxin Yin and James Wendt and Michael Bendersky and Rahma Chaabouni and Javier Snaider and Johan Ferret and Abhishek Jindal and Tara Thompson and Andrew Xue and Will Bishop and Shubham Milind Phal and Archit Sharma and Yunhsuan Sung and Prabakar Radhakrishnan and Mo Shomrat and Reeve Ingle and Roopali Vij and Justin Gilmer and Mihai Dorin Istin and Sam Sobell and Yang Lu and Emily Nottage and Dorsa Sadigh and Jeremiah Willcock and Tingnan Zhang and Steve Xu and Sasha Brown and Katherine Lee and Gary Wang and Yun Zhu and Yi Tay and Cheolmin Kim and Audrey Gutierrez and Abhanshu Sharma and Yongqin Xian and Sungyong Seo and Claire Cui and Elena Pochernina and Cip Baetu and Krzysztof Jastrzębski and Mimi Ly and Mohamed Elhawaty and Dan Suh and Eren Sezener and Pidong Wang and Nancy Yuen and George Tucker and Jiahao Cai and Zuguang Yang and Cindy Wang and Alex Muzio and Hai Qian and Jae Yoo and Derek Lockhart and Kevin R. McKee and Mandy Guo and Malika Mehrotra and Artur Mendonça and Sanket Vaibhav Mehta and Sherry Ben and Chetan Tekur and Jiaqi Mu and Muye Zhu and Victoria Krakovna and Hongrae Lee and AJ Maschinot and Sébastien Cevey and HyunJeong Choe and Aijun Bai and Hansa Srinivasan and Derek Gasaway and Nick Young and Patrick Siegler and Dan Holtmann-Rice and Vihari Piratla and Kate Baumli and Roey Yogev and Alex Hofer and Hado van Hasselt and Svetlana Grant and Yuri Chervonyi and David Silver and Andrew Hogue and Ayushi Agarwal and Kathie Wang and Preeti Singh and Four Flynn and Josh Lipschultz and Robert David and Lizzetth Bellot and Yao-Yuan Yang and Long Le and Filippo Graziano and Kate Olszewska and Kevin Hui and Akanksha Maurya and Nikos Parotsidis and Weijie Chen and Tayo Oguntebi and Joe Kelley and Anirudh Baddepudi and Johannes Mauerer and Gregory Shaw and Alex Siegman and Lin Yang and Shravya Shetty and Subhrajit Roy and Yunting Song and Wojciech Stokowiec and Ryan Burnell and Omkar Savant and Robert Busa-Fekete and Jin Miao and Samrat Ghosh and Liam MacDermed and Phillip Lippe and Mikhail Dektiarev and Zach Behrman and Fabian Mentzer and Kelvin Nguyen and Meng Wei and Siddharth Verma and Chris Knutsen and Sudeep Dasari and Zhipeng Yan and Petr Mitrichev and Xingyu Wang and Virat Shejwalkar and Jacob Austin and Srinivas Sunkara and Navneet Potti and Yan Virin and Christian Wright and Gaël Liu and Oriana Riva and Etienne Pot and Greg Kochanski and Quoc Le and Gargi Balasubramaniam and Arka Dhar and Yuguo Liao and Adam Bloniarz and Divyansh Shukla and Elizabeth Cole and Jong Lee and Sheng Zhang and Sushant Kafle and Siddharth Vashishtha and Parsa Mahmoudieh and Grace Chen and Raphael Hoffmann and Pranesh Srinivasan and Agustin Dal Lago and Yoav Ben Shalom and Zi Wang and Michael Elabd and Anuj Sharma and Junhyuk Oh and Suraj Kothawade and Maigo Le and Marianne Monteiro and Shentao Yang and Kaiz Alarakyia and Robert Geirhos and Diana Mincu and Håvard Garnes and Hayato Kobayashi and Soroosh Mariooryad and Kacper Krasowiak and Zhixin and Lai and Shibl Mourad and Mingqiu Wang and Fan Bu and Ophir Aharoni and Guanjie Chen and Abhimanyu Goyal and Vadim Zubov and Ankur Bapna and Elahe Dabir and Nisarg Kothari and Kay Lamerigts and Nicola De Cao and Jeremy Shar and Christopher Yew and Nitish Kulkarni and Dre Mahaarachchi and Mandar Joshi and Zhenhai Zhu and Jared Lichtarge and Yichao Zhou and Hannah Muckenhirn and Vittorio Selo and Oriol Vinyals and Peter Chen and Anthony Brohan and Vaibhav Mehta and Sarah Cogan and Ruth Wang and Ty Geri and Wei-Jen Ko and Wei Chen and Fabio Viola and Keshav Shivam and Lisa Wang and Madeleine Clare Elish and Raluca Ada Popa and Sébastien Pereira and Jianqiao Liu and Raphael Koster and Donnie Kim and Gufeng Zhang and Sayna Ebrahimi and Partha Talukdar and Yanyan Zheng and Petra Poklukar and Ales Mikhalap and Dale Johnson and Anitha Vijayakumar and Mark Omernick and Matt Dibb and Ayush Dubey and Qiong Hu and Apurv Suman and Vaibhav Aggarwal and Ilya Kornakov and Fei Xia and Wing Lowe and Alexey Kolganov and Ted Xiao and Vitaly Nikolaev and Steven Hemingray and Bonnie Li and Joana Iljazi and Mikołaj Rybiński and Ballie Sandhu and Peggy Lu and Thang Luong and Rodolphe Jenatton and Vineetha Govindaraj and Hui and Li and Gabriel Dulac-Arnold and Wonpyo Park and Henry Wang and Abhinit Modi and Jean Pouget-Abadie and Kristina Greller and Rahul Gupta and Robert Berry and Prajit Ramachandran and Jinyu Xie and Liam McCafferty and Jianling Wang and Kilol Gupta and Hyeontaek Lim and Blaž Bratanič and Andy Brock and Ilia Akolzin and Jim Sproch and Dan Karliner and Duhyeon Kim and Adrian Goedeckemeyer and Noam Shazeer and Cordelia Schmid and Daniele Calandriello and Parul Bhatia and Krzysztof Choromanski and Ceslee Montgomery and Dheeru Dua and Ana Ramalho and Helen King and Yue Gao and Lynn Nguyen and David Lindner and Divya Pitta and Oleaser Johnson and Khalid Salama and Diego Ardila and Michael Han and Erin Farnese and Seth Odoom and Ziyue Wang and Xiangzhuo Ding and Norman Rink and Ray Smith and Harshal Tushar Lehri and Eden Cohen and Neera Vats and Tong He and Parthasarathy Gopavarapu and Adam Paszke and Miteyan Patel and Wouter Van Gansbeke and Lucia Loher and Luis Castro and Maria Voitovich and Tamara von Glehn and Nelson George and Simon Niklaus and Zach Eaton-Rosen and Nemanja Rakićević and Erik Jue and Sagi Perel and Carrie Zhang and Yuval Bahat and Angéline Pouget and Zhi Xing and Fantine Huot and Ashish Shenoy and Taylor Bos and Vincent Coriou and Bryan Richter and Natasha Noy and Yaqing Wang and Santiago Ontanon and Siyang Qin and Gleb Makarchuk and Demis Hassabis and Zhuowan Li and Mandar Sharma and Kumaran Venkatesan and Iurii Kemaev and Roxanne Daniel and Shiyu Huang and Saloni Shah and Octavio Ponce and Warren and Chen and Manaal Faruqui and Jialin Wu and Slavica Andačić and Szabolcs Payrits and Daniel McDuff and Tom Hume and Yuan Cao and MH Tessler and Qingze Wang and Yinan Wang and Ivor Rendulic and Eirikur Agustsson and Matthew Johnson and Tanya Lando and Andrew Howard and Sri Gayatri Sundara Padmanabhan and Mayank Daswani and Andrea Banino and Michael Kilgore and Jonathan Heek and Ziwei Ji and Alvaro Caceres and Conglong Li and Nora Kassner and Alexey Vlaskin and Zeyu Liu and Alex Grills and Yanhan Hou and Roykrong Sukkerd and Gowoon Cheon and Nishita Shetty and Larisa Markeeva and Piotr Stanczyk and Tejas Iyer and Yuan Gong and Shawn Gao and Keerthana Gopalakrishnan and Tim Blyth and Malcolm Reynolds and Avishkar Bhoopchand and Misha Bilenko and Dero Gharibian and Vicky Zayats and Aleksandra Faust and Abhinav Singh and Min Ma and Hongyang Jiao and Sudheendra Vijayanarasimhan and Lora Aroyo and Vikas Yadav and Sarah Chakera and Ashwin Kakarla and Vilobh Meshram and Karol Gregor and Gabriela Botea and Evan Senter and Dawei Jia and Geza Kovacs and Neha Sharma and Sebastien Baur and Kai Kang and Yifan He and Lin Zhuo and Marija Kostelac and Itay Laish and Songyou Peng and Louis O'Bryan and Daniel Kasenberg and Girish Ramchandra Rao and Edouard Leurent and Biao Zhang and Sage Stevens and Ana Salazar and Ye Zhang and Ivan Lobov and Jake Walker and Allen Porter and Morgan Redshaw and Han Ke and Abhishek Rao and Alex Lee and Hoi Lam and Michael Moffitt and Jaeyoun Kim and Siyuan Qiao and Terry Koo and Robert Dadashi and Xinying Song and Mukund Sundararajan and Peng Xu and Chizu Kawamoto and Yan Zhong and Clara Barbu and Apoorv Reddy and Mauro Verzetti and Leon Li and George Papamakarios and Hanna Klimczak-Plucińska and Mary Cassin and Koray Kavukcuoglu and Rigel Swavely and Alain Vaucher and Jeffrey Zhao and Ross Hemsley and Michael Tschannen and Heming Ge and Gaurav Menghani and Yang Yu and Natalie Ha and Wei He and Xiao Wu and Maggie Song and Rachel Sterneck and Stefan Zinke and Dan A. Calian and Annie Marsden and Alejandro Cruzado Ruiz and Matteo Hessel and Almog Gueta and Benjamin Lee and Brian Farris and Manish Gupta and Yunjie Li and Mohammad Saleh and Vedant Misra and Kefan Xiao and Piermaria Mendolicchio and Gavin Buttimore and Varvara Krayvanova and Nigamaa Nayakanti and Matthew Wiethoff and Yash Pande and Azalia Mirhoseini and Ni Lao and Jasmine Liu and Yiqing Hua and Angie Chen and Yury Malkov and Dmitry Kalashnikov and Shubham Gupta and Kartik Audhkhasi and Yuexiang Zhai and Sudhindra Kopalle and Prateek Jain and Eran Ofek and Clemens Meyer and Khuslen Baatarsukh and Hana Strejček and Jun Qian and James Freedman and Ricardo Figueira and Michal Sokolik and Olivier Bachem and Raymond Lin and Dia Kharrat and Chris Hidey and Pingmei Xu and Dennis Duan and Yin Li and Muge Ersoy and Richard Everett and Kevin Cen and Rebeca Santamaria-Fernandez and Amir Taubenfeld and Ian Mackinnon and Linda Deng and Polina Zablotskaia and Shashank Viswanadha and Shivanker Goel and Damion Yates and Yunxiao Deng and Peter Choy and Mingqing Chen and Abhishek Sinha and Alex Mossin and Yiming Wang and Arthur Szlam and Susan Hao and Paul Kishan Rubenstein and Metin Toksoz-Exley and Miranda Aperghis and Yin Zhong and Junwhan Ahn and Michael Isard and Olivier Lacombe and Florian Luisier and Chrysovalantis Anastasiou and Yogesh Kalley and Utsav Prabhu and Emma Dunleavy and Shaan Bijwadia and Justin Mao-Jones and Kelly Chen and Rama Pasumarthi and Emily Wood and Adil Dostmohamed and Nate Hurley and Jiri Simsa and Alicia Parrish and Mantas Pajarskas and Matt Harvey and Ondrej Skopek and Yony Kochinski and Javier Rey and Verena Rieser and Denny Zhou and Sun Jae Lee and Trilok Acharya and Guowang Li and Joe Jiang and Xiaofan Zhang and Bryant Gipson and Ethan Mahintorabi and Marco Gelmi and Nima Khajehnouri and Angel Yeh and Kayi Lee and Loic Matthey and Leslie Baker and Trang Pham and Han Fu and Alex Pak and Prakhar Gupta and Cristina Vasconcelos and Adam Sadovsky and Brian Walker and Sissie Hsiao and Patrik Zochbauer and Andreea Marzoca and Noam Velan and Junhao Zeng and Gilles Baechler and Danny Driess and Divya Jain and Yanping Huang and Lizzie Tao and John Maggs and Nir Levine and Jon Schneider and Erika Gemzer and Samuel Petit and Shan Han and Zach Fisher and Dustin Zelle and Courtney Biles and Eugene Ie and Asya Fadeeva and Casper Liu and Juliana Vicente Franco and Adrian Collister and Hao Zhang and Renshen Wang and Ruizhe Zhao and Leandro Kieliger and Kurt Shuster and Rui Zhu and Boqing Gong and Lawrence Chan and Ruoxi Sun and Sujoy Basu and Roland Zimmermann and Jamie Hayes and Abhishek Bapna and Jasper Snoek and Weel Yang and Puranjay Datta and Jad Al Abdallah and Kevin Kilgour and Lu Li and SQ Mah and Yennie Jun and Morgane Rivière and Abhijit Karmarkar and Tammo Spalink and Tao Huang and Lucas Gonzalez and Duc-Hieu Tran and Averi Nowak and John Palowitch and Martin Chadwick and Ellie Talius and Harsh Mehta and Thibault Sellam and Philipp Fränken and Massimo Nicosia and Kyle He and Aditya Kini and David Amos and Sugato Basu and Harrison Jobe and Eleni Shaw and Qiantong Xu and Colin Evans and Daisuke Ikeda and Chaochao Yan and Larry Jin and Lun Wang and Sachin Yadav and Ilia Labzovsky and Ramesh Sampath and Ada Ma and Candice Schumann and Aditya Siddhant and Rohin Shah and John Youssef and Rishabh Agarwal and Natalie Dabney and Alessio Tonioni and Moran Ambar and Jing Li and Isabelle Guyon and Benny Li and David Soergel and Boya Fang and Georgi Karadzhov and Cristian Udrescu and Trieu Trinh and Vikas Raunak and Seb Noury and Dee Guo and Sonal Gupta and Mara Finkelstein and Denis Petek and Lihao Liang and Greg Billock and Pei Sun and David Wood and Yiwen Song and Xiaobin Yu and Tatiana Matejovicova and Regev Cohen and Kalyan Andra and David D'Ambrosio and Zhiwei Deng and Vincent Nallatamby and Ebrahim Songhori and Rumen Dangovski and Andrew Lampinen and Pankil Botadra and Adam Hillier and Jiawei Cao and Nagabhushan Baddi and Adhi Kuncoro and Toshihiro Yoshino and Ankit Bhagatwala and Marcáurelio Ranzato and Rylan Schaeffer and Tianlin Liu and Shuai Ye and Obaid Sarvana and John Nham and Chenkai Kuang and Isabel Gao and Jinoo Baek and Shubham Mittal and Ayzaan Wahid and Anita Gergely and Bin Ni and Josh Feldman and Carrie Muir and Pascal Lamblin and Wolfgang Macherey and Ethan Dyer and Logan Kilpatrick and Víctor Campos and Mukul Bhutani and Stanislav Fort and Yanif Ahmad and Aliaksei Severyn and Kleopatra Chatziprimou and Oleksandr Ferludin and Mason Dimarco and Aditya Kusupati and Joe Heyward and Dan Bahir and Kevin Villela and Katie Millican and Dror Marcus and Sanaz Bahargam and Caglar Unlu and Nicholas Roth and Zichuan Wei and Siddharth Gopal and Deepanway Ghoshal and Edward Lee and Sharon Lin and Jennie Lees and Dayeong Lee and Anahita Hosseini and Connie Fan and Seth Neel and Marcus Wu and Yasemin Altun and Honglong Cai and Enrique Piqueras and Josh Woodward and Alessandro Bissacco and Salem Haykal and Mahyar Bordbar and Prasha Sundaram and Sarah Hodkinson and Daniel Toyama and George Polovets and Austin Myers and Anu Sinha and Tomer Levinboim and Kashyap Krishnakumar and Rachita Chhaparia and Tatiana Sholokhova and Nitesh Bharadwaj Gundavarapu and Ganesh Jawahar and Haroon Qureshi and Jieru Hu and Nikola Momchev and Matthew Rahtz and Renjie Wu and Aishwarya P S and Kedar Dhamdhere and Meiqi Guo and Umang Gupta and Ali Eslami and Mariano Schain and Michiel Blokzijl and David Welling and Dave Orr and Levent Bolelli and Nicolas Perez-Nieves and Mikhail Sirotenko and Aman Prasad and Arjun Kar and Borja De Balle Pigem and Tayfun Terzi and Gellért Weisz and Dipankar Ghosh and Aditi Mavalankar and Dhruv Madeka and Kaspar Daugaard and Hartwig Adam and Viraj Shah and Dana Berman and Maggie Tran and Steven Baker and Ewa Andrejczuk and Grishma Chole and Ganna Raboshchuk and Mahdi Mirzazadeh and Thais Kagohara and Shimu Wu and Christian Schallhart and Bernett Orlando and Chen Wang and Alban Rrustemi and Hao Xiong and Hao Liu and Arpi Vezer and Nolan Ramsden and Shuo-yiin Chang and Sidharth Mudgal and Yan Li and Nino Vieillard and Yedid Hoshen and Farooq Ahmad and Ambrose Slone and Amy Hua and Natan Potikha and Mirko Rossini and Jon Stritar and Sushant Prakash and Zifeng Wang and Xuanyi Dong and Alireza Nazari and Efrat Nehoran and Kaan Tekelioglu and Yinxiao Li and Kartikeya Badola and Tom Funkhouser and Yuanzhen Li and Varun Yerram and Ramya Ganeshan and Daniel Formoso and Karol Langner and Tian Shi and Huijian Li and Yumeya Yamamori and Amayika Panda and Alaa Saade and Angelo Scorza Scarpati and Chris Breaux and CJ Carey and Zongwei Zhou and Cho-Jui Hsieh and Sophie Bridgers and Alena Butryna and Nishesh Gupta and Vaibhav Tulsyan and Sanghyun Woo and Evgenii Eltyshev and Will Grathwohl and Chanel Parks and Seth Benjamin and Rina Panigrahy and Shenil Dodhia and Daniel De Freitas and Chris Sauer and Will Song and Ferran Alet and Jackson Tolins and Cosmin Paduraru and Xingyi Zhou and Brian Albert and Zizhao Zhang and Lei Shu and Mudit Bansal and Sarah Nguyen and Amir Globerson and Owen Xiao and James Manyika and Tom Hennigan and Rong Rong and Josip Matak and Anton Bakalov and Ankur Sharma and Danila Sinopalnikov and Andrew Pierson and Stephen Roller and Geoff Brown and Mingcen Gao and Toshiyuki Fukuzawa and Amin Ghafouri and Kenny Vassigh and Iain Barr and Zhicheng Wang and Anna Korsun and Rajesh Jayaram and Lijie Ren and Tim Zaman and Samira Khan and Yana Lunts and Dan Deutsch and Dave Uthus and Nitzan Katz and Masha Samsikova and Amr Khalifa and Nikhil Sethi and Jiao Sun and Luming Tang and Uri Alon and Xianghong Luo and Dian Yu and Abhishek Nayyar and Bryce Petrini and Will Truong and Vincent Hellendoorn and Nikolai Chinaev and Chris Alberti and Wei Wang and Jingcao Hu and Vahab Mirrokni and Ananth Balashankar and Avia Aharon and Aahil Mehta and Ahmet Iscen and Joseph Kready and Lucas Manning and Anhad Mohananey and Yuankai Chen and Anshuman Tripathi and Allen Wu and Igor Petrovski and Dawsen Hwang and Martin Baeuml and Shreyas Chandrakaladharan and Yuan Liu and Rey Coaguila and Maxwell Chen and Sally Ma and Pouya Tafti and Susheel Tatineni and Terry Spitz and Jiayu Ye and Paul Vicol and Mihaela Rosca and Adrià Puigdomènech and Zohar Yahav and Sanjay Ghemawat and Hanzhao Lin and Phoebe Kirk and Zaid Nabulsi and Sergey Brin and Bernd Bohnet and Ken Caluwaerts and Aditya Srikanth Veerubhotla and Dan Zheng and Zihang Dai and Petre Petrov and Yichong Xu and Ramin Mehran and Zhuo Xu and Luisa Zintgraf and Jiho Choi and Spurthi Amba Hombaiah and Romal Thoppilan and Sashank Reddi and Lukasz Lew and Li Li and Kellie Webster and KP Sawhney and Lampros Lamprou and Siamak Shakeri and Mayank Lunayach and Jianmin Chen and Sumit Bagri and Alex Salcianu and Ying Chen and Yani Donchev and Charlotte Magister and Signe Nørly and Vitor Rodrigues and Tomas Izo and Hila Noga and Joe Zou and Thomas Köppe and Wenxuan Zhou and Kenton Lee and Xiangzhu Long and Danielle Eisenbud and Anthony Chen and Connor Schenck and Chi Ming To and Peilin Zhong and Emanuel Taropa and Minh Truong and Omer Levy and Danilo Martins and Zhiyuan Zhang and Christopher Semturs and Kelvin Zhang and Alex Yakubovich and Pol Moreno and Lara McConnaughey and Di Lu and Sam Redmond and Lotte Weerts and Yonatan Bitton and Tiziana Refice and Nicolas Lacasse and Arthur Conmy and Corentin Tallec and Julian Odell and Hannah Forbes-Pollard and Arkadiusz Socala and Jonathan Hoech and Pushmeet Kohli and Alanna Walton and Rui Wang and Mikita Sazanovich and Kexin Zhu and Andrei Kapishnikov and Rich Galt and Matthew Denton and Ben Murdoch and Caitlin Sikora and Kareem Mohamed and Wei Wei and Uri First and Tim McConnell and Luis C. Cobo and James Qin and Thi Avrahami and Daniel Balle and Yu Watanabe and Annie Louis and Adam Kraft and Setareh Ariafar and Yiming Gu and Eugénie Rives and Charles Yoon and Andrei Rusu and James Cobon-Kerr and Chris Hahn and Jiaming Luo and Yuvein and Zhu and Niharika Ahuja and Rodrigo Benenson and Raphaël Lopez Kaufman and Honglin Yu and Lloyd Hightower and Junlin Zhang and Darren Ni and Lisa Anne Hendricks and Gabby Wang and Gal Yona and Lalit Jain and Pablo Barrio and Surya Bhupatiraju and Siva Velusamy and Allan Dafoe and Sebastian Riedel and Tara Thomas and Zhe Yuan and Mathias Bellaiche and Sheena Panthaplackel and Klemen Kloboves and Sarthak Jauhari and Canfer Akbulut and Todor Davchev and Evgeny Gladchenko and David Madras and Aleksandr Chuklin and Tyrone Hill and Quan Yuan and Mukundan Madhavan and Luke Leonhard and Dylan Scandinaro and Qihang Chen and Ning Niu and Arthur Douillard and Bogdan Damoc and Yasumasa Onoe and Fabian Pedregosa and Fred Bertsch and Chas Leichner and Joseph Pagadora and Jonathan Malmaud and Sameera Ponda and Andy Twigg and Oleksii Duzhyi and Jingwei Shen and Miaosen Wang and Roopal Garg and Jing Chen and Utku Evci and Jonathan Lee and Leon Liu and Koji Kojima and Masa Yamaguchi and Arunkumar Rajendran and AJ Piergiovanni and Vinodh Kumar Rajendran and Marco Fornoni and Gabriel Ibagon and Harry Ragan and Sadh MNM Khan and John Blitzer and Andrew Bunner and Guan Sun and Takahiro Kosakai and Scott Lundberg and Ndidi Elue and Kelvin Guu and SK Park and Jane Park and Arunachalam Narayanaswamy and Chengda Wu and Jayaram Mudigonda and Trevor Cohn and Hairong Mu and Ravi Kumar and Laura Graesser and Yichi Zhang and Richard Killam and Vincent Zhuang and Mai Giménez and Wael Al Jishi and Ruy Ley-Wild and Alex Zhai and Kazuki Osawa and Diego Cedillo and Jialu Liu and Mayank Upadhyay and Marcin Sieniek and Roshan Sharma and Tom Paine and Anelia Angelova and Sravanti Addepalli and Carolina Parada and Kingshuk Majumder and Avery Lamp and Sanjiv Kumar and Xiang Deng and Artiom Myaskovsky and Tea Sabolić and Jeffrey Dudek and Sarah York and Félix de Chaumont Quitry and Jiazhong Nie and Dee Cattle and Alok Gunjan and Bilal Piot and Waleed Khawaja and Seojin Bang and Simon Wang and Siavash Khodadadeh and Raghavender R and Praynaa Rawlani and Richard Powell and Kevin Lee and Johannes Griesser and GS Oh and Cesar Magalhaes and Yujia Li and Simon Tokumine and Hadas Natalie Vogel and Dennis Hsu and Arturo BC and Disha Jindal and Matan Cohen and Zi Yang and Junwei Yuan and Dario de Cesare and Tony Bruguier and Jun Xu and Monica Roy and Alon Jacovi and Dan Belov and Rahul Arya and Phoenix Meadowlark and Shlomi Cohen-Ganor and Wenting Ye and Patrick Morris-Suzuki and Praseem Banzal and Gan Song and Pranavaraj Ponnuramu and Fred Zhang and George Scrivener and Salah Zaiem and Alif Raditya Rochman and Kehang Han and Badih Ghazi and Kate Lee and Shahar Drath and Daniel Suo and Antonious Girgis and Pradeep Shenoy and Duy Nguyen and Douglas Eck and Somit Gupta and Le Yan and Joao Carreira and Anmol Gulati and Ruoxin Sang and Daniil Mirylenka and Emma Cooney and Edward Chou and Mingyang Ling and Cindy Fan and Ben Coleman and Guilherme Tubone and Ravin Kumar and Jason Baldridge and Felix Hernandez-Campos and Angeliki Lazaridou and James Besley and Itay Yona and Neslihan Bulut and Quentin Wellens and AJ Pierigiovanni and Jasmine George and Richard Green and Pu Han and Connie Tao and Geoff Clark and Chong You and Abbas Abdolmaleki and Justin Fu and Tongzhou Chen and Ashwin Chaugule and Angad Chandorkar and Altaf Rahman and Will Thompson and Penporn Koanantakool and Mike Bernico and Jie Ren and Andrey Vlasov and Sergei Vassilvitskii and Maciej Kula and Yizhong Liang and Dahun Kim and Yangsibo Huang and Chengxi Ye and Dmitry Lepikhin and Wesley Helmholz},
      year={2025},
      eprint={2507.06261},
      archivePrefix={arXiv},
      primaryClass={cs.CL},
      url={https://arxiv.org/abs/2507.06261}, 
}

@misc{llama3herdmodels,
      title={The Llama 3 Herd of Models}, 
      author={Aaron Grattafiori and Abhimanyu Dubey and Abhinav Jauhri and Abhinav Pandey and Abhishek Kadian and Ahmad Al-Dahle and Aiesha Letman and Akhil Mathur and Alan Schelten and Alex Vaughan and Amy Yang and Angela Fan and Anirudh Goyal and Anthony Hartshorn and Aobo Yang and Archi Mitra and Archie Sravankumar and Artem Korenev and Arthur Hinsvark and Arun Rao and Aston Zhang and Aurelien Rodriguez and Austen Gregerson and Ava Spataru and Baptiste Roziere and Bethany Biron and Binh Tang and Bobbie Chern and Charlotte Caucheteux and Chaya Nayak and Chloe Bi and Chris Marra and Chris McConnell and Christian Keller and Christophe Touret and Chunyang Wu and Corinne Wong and Cristian Canton Ferrer and Cyrus Nikolaidis and Damien Allonsius and Daniel Song and Danielle Pintz and Danny Livshits and Danny Wyatt and David Esiobu and Dhruv Choudhary and Dhruv Mahajan and Diego Garcia-Olano and Diego Perino and Dieuwke Hupkes and Egor Lakomkin and Ehab AlBadawy and Elina Lobanova and Emily Dinan and Eric Michael Smith and Filip Radenovic and Francisco Guzmán and Frank Zhang and Gabriel Synnaeve and Gabrielle Lee and Georgia Lewis Anderson and Govind Thattai and Graeme Nail and Gregoire Mialon and Guan Pang and Guillem Cucurell and Hailey Nguyen and Hannah Korevaar and Hu Xu and Hugo Touvron and Iliyan Zarov and Imanol Arrieta Ibarra and Isabel Kloumann and Ishan Misra and Ivan Evtimov and Jack Zhang and Jade Copet and Jaewon Lee and Jan Geffert and Jana Vranes and Jason Park and Jay Mahadeokar and Jeet Shah and Jelmer van der Linde and Jennifer Billock and Jenny Hong and Jenya Lee and Jeremy Fu and Jianfeng Chi and Jianyu Huang and Jiawen Liu and Jie Wang and Jiecao Yu and Joanna Bitton and Joe Spisak and Jongsoo Park and Joseph Rocca and Joshua Johnstun and Joshua Saxe and Junteng Jia and Kalyan Vasuden Alwala and Karthik Prasad and Kartikeya Upasani and Kate Plawiak and Ke Li and Kenneth Heafield and Kevin Stone and Khalid El-Arini and Krithika Iyer and Kshitiz Malik and Kuenley Chiu and Kunal Bhalla and Kushal Lakhotia and Lauren Rantala-Yeary and Laurens van der Maaten and Lawrence Chen and Liang Tan and Liz Jenkins and Louis Martin and Lovish Madaan and Lubo Malo and Lukas Blecher and Lukas Landzaat and Luke de Oliveira and Madeline Muzzi and Mahesh Pasupuleti and Mannat Singh and Manohar Paluri and Marcin Kardas and Maria Tsimpoukelli and Mathew Oldham and Mathieu Rita and Maya Pavlova and Melanie Kambadur and Mike Lewis and Min Si and Mitesh Kumar Singh and Mona Hassan and Naman Goyal and Narjes Torabi and Nikolay Bashlykov and Nikolay Bogoychev and Niladri Chatterji and Ning Zhang and Olivier Duchenne and Onur Çelebi and Patrick Alrassy and Pengchuan Zhang and Pengwei Li and Petar Vasic and Peter Weng and Prajjwal Bhargava and Pratik Dubal and Praveen Krishnan and Punit Singh Koura and Puxin Xu and Qing He and Qingxiao Dong and Ragavan Srinivasan and Raj Ganapathy and Ramon Calderer and Ricardo Silveira Cabral and Robert Stojnic and Roberta Raileanu and Rohan Maheswari and Rohit Girdhar and Rohit Patel and Romain Sauvestre and Ronnie Polidoro and Roshan Sumbaly and Ross Taylor and Ruan Silva and Rui Hou and Rui Wang and Saghar Hosseini and Sahana Chennabasappa and Sanjay Singh and Sean Bell and Seohyun Sonia Kim and Sergey Edunov and Shaoliang Nie and Sharan Narang and Sharath Raparthy and Sheng Shen and Shengye Wan and Shruti Bhosale and Shun Zhang and Simon Vandenhende and Soumya Batra and Spencer Whitman and Sten Sootla and Stephane Collot and Suchin Gururangan and Sydney Borodinsky and Tamar Herman and Tara Fowler and Tarek Sheasha and Thomas Georgiou and Thomas Scialom and Tobias Speckbacher and Todor Mihaylov and Tong Xiao and Ujjwal Karn and Vedanuj Goswami and Vibhor Gupta and Vignesh Ramanathan and Viktor Kerkez and Vincent Gonguet and Virginie Do and Vish Vogeti and Vítor Albiero and Vladan Petrovic and Weiwei Chu and Wenhan Xiong and Wenyin Fu and Whitney Meers and Xavier Martinet and Xiaodong Wang and Xiaofang Wang and Xiaoqing Ellen Tan and Xide Xia and Xinfeng Xie and Xuchao Jia and Xuewei Wang and Yaelle Goldschlag and Yashesh Gaur and Yasmine Babaei and Yi Wen and Yiwen Song and Yuchen Zhang and Yue Li and Yuning Mao and Zacharie Delpierre Coudert and Zheng Yan and Zhengxing Chen and Zoe Papakipos and Aaditya Singh and Aayushi Srivastava and Abha Jain and Adam Kelsey and Adam Shajnfeld and Adithya Gangidi and Adolfo Victoria and Ahuva Goldstand and Ajay Menon and Ajay Sharma and Alex Boesenberg and Alexei Baevski and Allie Feinstein and Amanda Kallet and Amit Sangani and Amos Teo and Anam Yunus and Andrei Lupu and Andres Alvarado and Andrew Caples and Andrew Gu and Andrew Ho and Andrew Poulton and Andrew Ryan and Ankit Ramchandani and Annie Dong and Annie Franco and Anuj Goyal and Aparajita Saraf and Arkabandhu Chowdhury and Ashley Gabriel and Ashwin Bharambe and Assaf Eisenman and Azadeh Yazdan and Beau James and Ben Maurer and Benjamin Leonhardi and Bernie Huang and Beth Loyd and Beto De Paola and Bhargavi Paranjape and Bing Liu and Bo Wu and Boyu Ni and Braden Hancock and Bram Wasti and Brandon Spence and Brani Stojkovic and Brian Gamido and Britt Montalvo and Carl Parker and Carly Burton and Catalina Mejia and Ce Liu and Changhan Wang and Changkyu Kim and Chao Zhou and Chester Hu and Ching-Hsiang Chu and Chris Cai and Chris Tindal and Christoph Feichtenhofer and Cynthia Gao and Damon Civin and Dana Beaty and Daniel Kreymer and Daniel Li and David Adkins and David Xu and Davide Testuggine and Delia David and Devi Parikh and Diana Liskovich and Didem Foss and Dingkang Wang and Duc Le and Dustin Holland and Edward Dowling and Eissa Jamil and Elaine Montgomery and Eleonora Presani and Emily Hahn and Emily Wood and Eric-Tuan Le and Erik Brinkman and Esteban Arcaute and Evan Dunbar and Evan Smothers and Fei Sun and Felix Kreuk and Feng Tian and Filippos Kokkinos and Firat Ozgenel and Francesco Caggioni and Frank Kanayet and Frank Seide and Gabriela Medina Florez and Gabriella Schwarz and Gada Badeer and Georgia Swee and Gil Halpern and Grant Herman and Grigory Sizov and Guangyi and Zhang and Guna Lakshminarayanan and Hakan Inan and Hamid Shojanazeri and Han Zou and Hannah Wang and Hanwen Zha and Haroun Habeeb and Harrison Rudolph and Helen Suk and Henry Aspegren and Hunter Goldman and Hongyuan Zhan and Ibrahim Damlaj and Igor Molybog and Igor Tufanov and Ilias Leontiadis and Irina-Elena Veliche and Itai Gat and Jake Weissman and James Geboski and James Kohli and Janice Lam and Japhet Asher and Jean-Baptiste Gaya and Jeff Marcus and Jeff Tang and Jennifer Chan and Jenny Zhen and Jeremy Reizenstein and Jeremy Teboul and Jessica Zhong and Jian Jin and Jingyi Yang and Joe Cummings and Jon Carvill and Jon Shepard and Jonathan McPhie and Jonathan Torres and Josh Ginsburg and Junjie Wang and Kai Wu and Kam Hou U and Karan Saxena and Kartikay Khandelwal and Katayoun Zand and Kathy Matosich and Kaushik Veeraraghavan and Kelly Michelena and Keqian Li and Kiran Jagadeesh and Kun Huang and Kunal Chawla and Kyle Huang and Lailin Chen and Lakshya Garg and Lavender A and Leandro Silva and Lee Bell and Lei Zhang and Liangpeng Guo and Licheng Yu and Liron Moshkovich and Luca Wehrstedt and Madian Khabsa and Manav Avalani and Manish Bhatt and Martynas Mankus and Matan Hasson and Matthew Lennie and Matthias Reso and Maxim Groshev and Maxim Naumov and Maya Lathi and Meghan Keneally and Miao Liu and Michael L. Seltzer and Michal Valko and Michelle Restrepo and Mihir Patel and Mik Vyatskov and Mikayel Samvelyan and Mike Clark and Mike Macey and Mike Wang and Miquel Jubert Hermoso and Mo Metanat and Mohammad Rastegari and Munish Bansal and Nandhini Santhanam and Natascha Parks and Natasha White and Navyata Bawa and Nayan Singhal and Nick Egebo and Nicolas Usunier and Nikhil Mehta and Nikolay Pavlovich Laptev and Ning Dong and Norman Cheng and Oleg Chernoguz and Olivia Hart and Omkar Salpekar and Ozlem Kalinli and Parkin Kent and Parth Parekh and Paul Saab and Pavan Balaji and Pedro Rittner and Philip Bontrager and Pierre Roux and Piotr Dollar and Polina Zvyagina and Prashant Ratanchandani and Pritish Yuvraj and Qian Liang and Rachad Alao and Rachel Rodriguez and Rafi Ayub and Raghotham Murthy and Raghu Nayani and Rahul Mitra and Rangaprabhu Parthasarathy and Raymond Li and Rebekkah Hogan and Robin Battey and Rocky Wang and Russ Howes and Ruty Rinott and Sachin Mehta and Sachin Siby and Sai Jayesh Bondu and Samyak Datta and Sara Chugh and Sara Hunt and Sargun Dhillon and Sasha Sidorov and Satadru Pan and Saurabh Mahajan and Saurabh Verma and Seiji Yamamoto and Sharadh Ramaswamy and Shaun Lindsay and Shaun Lindsay and Sheng Feng and Shenghao Lin and Shengxin Cindy Zha and Shishir Patil and Shiva Shankar and Shuqiang Zhang and Shuqiang Zhang and Sinong Wang and Sneha Agarwal and Soji Sajuyigbe and Soumith Chintala and Stephanie Max and Stephen Chen and Steve Kehoe and Steve Satterfield and Sudarshan Govindaprasad and Sumit Gupta and Summer Deng and Sungmin Cho and Sunny Virk and Suraj Subramanian and Sy Choudhury and Sydney Goldman and Tal Remez and Tamar Glaser and Tamara Best and Thilo Koehler and Thomas Robinson and Tianhe Li and Tianjun Zhang and Tim Matthews and Timothy Chou and Tzook Shaked and Varun Vontimitta and Victoria Ajayi and Victoria Montanez and Vijai Mohan and Vinay Satish Kumar and Vishal Mangla and Vlad Ionescu and Vlad Poenaru and Vlad Tiberiu Mihailescu and Vladimir Ivanov and Wei Li and Wenchen Wang and Wenwen Jiang and Wes Bouaziz and Will Constable and Xiaocheng Tang and Xiaojian Wu and Xiaolan Wang and Xilun Wu and Xinbo Gao and Yaniv Kleinman and Yanjun Chen and Ye Hu and Ye Jia and Ye Qi and Yenda Li and Yilin Zhang and Ying Zhang and Yossi Adi and Youngjin Nam and Yu and Wang and Yu Zhao and Yuchen Hao and Yundi Qian and Yunlu Li and Yuzi He and Zach Rait and Zachary DeVito and Zef Rosnbrick and Zhaoduo Wen and Zhenyu Yang and Zhiwei Zhao and Zhiyu Ma},
      year={2024},
      eprint={2407.21783},
      archivePrefix={arXiv},
      primaryClass={cs.AI},
      url={https://arxiv.org/abs/2407.21783}, 
}

@article{guidance,
author = {Nippold, Marilyn},
year = {2023},
month = {11},
pages = {1-13},
title = {Unlocking Logical Fallacies: A Key to Building Critical Thinking Skills in Adolescents},
volume = {9},
journal = {Perspectives of the ASHA Special Interest Groups},
doi = {10.1044/2023_PERSP-23-00108}
}

@article{outcome,
author = {Terentev, Evgeny and Shcheglova, Irina and Federiakin, Denis and Koreshnikova, Yuliya and Costley, Jamie},
year = {2024},
month = {10},
pages = {},
title = {Active versus Passive Teaching: Students’ Perceptions and Thinking Skills},
volume = {1},
journal = {Voprosy obrazovaniya / Educational Studies Moscow},
doi = {10.17323/vo-2024-16364}
}

@misc{krupp2023challengesopportunitiesmoderatingusage,
      title={Challenges and Opportunities of Moderating Usage of Large Language Models in Education}, 
      author={Lars Krupp and Steffen Steinert and Maximilian Kiefer-Emmanouilidis and Karina E. Avila and Paul Lukowicz and Jochen Kuhn and Stefan Küchemann and Jakob Karolus},
      year={2023},
      eprint={2312.14969},
      archivePrefix={arXiv},
      primaryClass={cs.HC},
      url={https://arxiv.org/abs/2312.14969}, 
}

@article{REFUTE,
author = {Patricia Kowalski and Annette Kujawski Taylor},
title ={The Effect of Refuting Misconceptions in the Introductory Psychology Class},

journal = {Teaching of Psychology},
volume = {36},
number = {3},
pages = {153-159},
year = {2009},
doi = {10.1080/00986280902959986},
URL = { 
    
        https://doi.org/10.1080/00986280902959986
},
eprint = { 
        https://doi.org/10.1080/00986280902959986

}
}

@book{withey2016mastering,
  title={Mastering Logical Fallacies: The Definitive Guide to Flawless Rhetoric and Bulletproof Logic},
  author={Withey, M. and Zhang, H.},
  isbn={9781623157111},
  url={https://books.google.ch/books?id=jWMXEQAAQBAJ},
  year={2016},
  publisher={Sourcebooks}
}

@misc{represponse,
      title={Learning to Break the Loop: Analyzing and Mitigating Repetitions for Neural Text Generation}, 
      author={Jin Xu and Xiaojiang Liu and Jianhao Yan and Deng Cai and Huayang Li and Jian Li},
      year={2022},
      eprint={2206.02369},
      archivePrefix={arXiv},
      primaryClass={cs.CL},
      url={https://arxiv.org/abs/2206.02369}, 
}

@article{Tracy01101984,
author = {Karen Tracy},
title = {Staying on topic: An explication of conversational relevance },
journal = {Discourse Processes},
volume = {7},
number = {4},
pages = {447--464},
year = {1984},
publisher = {Routledge},
doi = {10.1080/01638538409544600},
URL = { https://doi.org/10.1080/01638538409544600

},
eprint = {   https://doi.org/10.1080/01638538409544600
    

}

}

@article{review_llm_ed, title={A comprehensive review of large language models: Issues and solutions in Learning Environments}, volume={6}, DOI={10.1007/s43621-025-00815-8}, number={1}, journal={Discover Sustainability}, author={Shahzad, Tariq and Mazhar, Tehseen and Tariq, Muhammad Usman and Ahmad, Wasim and Ouahada, Khmaies and Hamam, Habib}, year={2025}, month={Jan}}

@article{lack_lf_un, title={Measuring University Students’ ability to recognize argument structures and fallacies}, volume={14}, DOI={10.3389/fpsyg.2023.1270931}, journal={Frontiers in Psychology}, author={Berkle, Yvonne and Schmitt, Lukas and Tolzin, Antonia and Janson, Andreas and Wambsganss, Thiemo and Leimeister, Jan Marco and Leuchter, Miriam}, year={2023}, month={Dec}}

@article{teach_socra, title={How Socratic Pedagogy Works}, volume={23}, DOI={https://doi.org/10.22329/il.v23i2.2170}, number={2}, journal={Informal Logic}, author={Boghossian, Pete}, year={2003}, month={Jan} }

@book{miles2020qualitative,
  title     = {Qualitative Data Analysis: A Methods Sourcebook},
  author    = {Miles, Matthew B. and Huberman, A. Michael and Saldaña, Johnny},
  edition   = {4th},
  year      = {2020},
  publisher = {SAGE Publications},
  address   = {Thousand Oaks, CA},
}

@misc{scarlatos2025trainingllmbasedtutorsimprove,
      title={Training LLM-based Tutors to Improve Student Learning Outcomes in Dialogues}, 
      author={Alexander Scarlatos and Naiming Liu and Jaewook Lee and Richard Baraniuk and Andrew Lan},
      year={2025},
      eprint={2503.06424},
      archivePrefix={arXiv},
      primaryClass={cs.CL},
      url={https://arxiv.org/abs/2503.06424}, 
}

@article{risen2007informal,
  title={Informal logical fallacies},
  author={Risen, Jane and Gilovich, Thomas and Sternberg, R and Halpern, D and Roediger, H},
  journal={Critical thinking in psychology},
  volume={110},
  year={2007},
  publisher={Cambridge University Press New York, NY}
}

@article{shapiro,
 ISSN = {00063444, 14643510},
 URL = {http://www.jstor.org/stable/2333709},
 author = {S. S. Shapiro and M. B. Wilk},
 journal = {Biometrika},
 number = {3/4},
 pages = {591--611},
 publisher = {[Oxford University Press, Biometrika Trust]},
 title = {An Analysis of Variance Test for Normality (Complete Samples)},
 urldate = {2025-05-15},
 volume = {52},
 year = {1965}
}

@article{ttest,
  title={The probable error of a mean},
  author={Student},
  journal={Biometrika},
  pages={1--25},
  year={1908},
  publisher={JSTOR}
}

@misc{jeong2025largelanguagemodelsbetter,
      title={Large Language Models Are Better Logical Fallacy Reasoners with Counterargument, Explanation, and Goal-Aware Prompt Formulation}, 
      author={Jiwon Jeong and Hyeju Jang and Hogun Park},
      year={2025},
      eprint={2503.23363},
      archivePrefix={arXiv},
      primaryClass={cs.AI},
      url={https://arxiv.org/abs/2503.23363}, 
}

@misc{pitre2024argumentoraugmentinguserexperiences,
      title={ArguMentor: Augmenting User Experiences with Counter-Perspectives}, 
      author={Priya Pitre and Kurt Luther},
      year={2024},
      eprint={2406.02795},
      archivePrefix={arXiv},
      primaryClass={cs.HC},
      url={https://arxiv.org/abs/2406.02795}, 
}

@misc{surveylargelanguagemodelseducation,
      title={Large Language Models for Education: A Survey and Outlook}, 
      author={Shen Wang and Tianlong Xu and Hang Li and Chaoli Zhang and Joleen Liang and Jiliang Tang and Philip S. Yu and Qingsong Wen},
      year={2024},
      eprint={2403.18105},
      archivePrefix={arXiv},
      primaryClass={cs.CL},
      url={https://arxiv.org/abs/2403.18105}, 
}

@article{motivated_irrational,
author = {Correia, Vasco},
year = {2011},
month = {01},
pages = {107-118},
title = {Biases and fallacies: The role of motivated irrationality in fallacious reasoning},
volume = {3},
journal = {Cogency}
}

@book{ARGUMENTS,
    author  = {Walter Sinott-Armstrong and Robert Frogelin},
    title   = {Understanding Arguments: An Introduction to Informal Logic},
    year    = "2015",
    publisher = {Cengage Learning},
    address = {Stamford, CT}
}

@inproceedings{
liu2024socraticlm,
title={Socratic{LM}: Exploring Socratic Personalized Teaching with Large Language Models},
author={Jiayu Liu and Zhenya Huang and Tong Xiao and Jing Sha and Jinze Wu and Qi Liu and Shijin Wang and Enhong Chen},
booktitle={The Thirty-eighth Annual Conference on Neural Information Processing Systems},
year={2024},
url={https://openreview.net/forum?id=qkoZgJhxsA}
}

@Article{info:doi/10.2196/37367,
author="Skafle, Ingjerd
and Nordahl-Hansen, Anders
and Quintana, Daniel S
and Wynn, Rolf
and Gabarron, Elia",
title="Misinformation About COVID-19 Vaccines on Social Media: Rapid Review",
journal="J Med Internet Res",
year="2022",
month="Aug",
day="4",
volume="24",
number="8",
pages="e37367",
issn="1438-8871",
doi="10.2196/37367",
url="https://www.jmir.org/2022/8/e37367",
url="https://doi.org/10.2196/37367",
url="http://www.ncbi.nlm.nih.gov/pubmed/35816685"
}

@article{
doi:10.1073/pnas.1517441113,
author = {Michela Del Vicario  and Alessandro Bessi  and Fabiana Zollo  and Fabio Petroni  and Antonio Scala  and Guido Caldarelli  and H. Eugene Stanley  and Walter Quattrociocchi },
title = {The spreading of misinformation online},
journal = {Proceedings of the National Academy of Sciences},
volume = {113},
number = {3},
pages = {554-559},
year = {2016},
doi = {10.1073/pnas.1517441113},
URL = {https://www.pnas.org/doi/abs/10.1073/pnas.1517441113},
eprint = {https://www.pnas.org/doi/pdf/10.1073/pnas.1517441113}}

@inproceedings{
zhou2024sotopia,
title={{SOTOPIA}: Interactive Evaluation for Social Intelligence in Language Agents},
author={Xuhui Zhou and Hao Zhu and Leena Mathur and Ruohong Zhang and Haofei Yu and Zhengyang Qi and Louis-Philippe Morency and Yonatan Bisk and Daniel Fried and Graham Neubig and Maarten Sap},
booktitle={The Twelfth International Conference on Learning Representations},
year={2024},
url={https://openreview.net/forum?id=mM7VurbA4r}
}

@article{holdencritedd,

author = {Ed, Holden and Ed, Schmit},
year = {2002},
month = {01},
pages = {},
title = {Inquiry and the Literary Text: Constructing Discussions in the English Classroom. Classroom Practices in Teaching English},
isbn = {ISBN-0-8141-2343-0}
}

@book{Hitchcock2017_toulmin,
author="Hitchcock, David",
title="Good Reasoning on the Toulmin Model",
bookTitle="On Reasoning and Argument: Essays in Informal Logic and on Critical Thinking",
year="2017",
publisher="Springer International Publishing",
address="Cham",
pages="371--387",
isbn="978-3-319-53562-3",
doi="10.1007/978-3-319-53562-3_23",
url="https://doi.org/10.1007/978-3-319-53562-3_23"
}

@book{critthinking,
  title={Developing minds: a resource book for teaching thinking},
  author={Costa, Arthur L.},
  year={2001},
  publisher={Association for Supervision and Curriculum Development}
}

@book{walton2008argumentation,
  title={Argumentation schemes},
  author={Walton, Douglas},
  year={2008},
  publisher={Cambridge University Press}
}

@misc{puech2024pedagogicalsteeringlargelanguage,
      title={Towards the Pedagogical Steering of Large Language Models for Tutoring: A Case Study with Modeling Productive Failure}, 
      author={Romain Puech and Jakub Macina and Julia Chatain and Mrinmaya Sachan and Manu Kapur},
      year={2024},
      eprint={2410.03781},
      archivePrefix={arXiv},
      primaryClass={cs.HC},
      url={https://arxiv.org/abs/2410.03781}, 
}

@inproceedings{autotutor,
author = {Pal Chowdhury, Sankalan and Zouhar, Vil\'{e}m and Sachan, Mrinmaya},
title = {AutoTutor meets Large Language Models: A Language Model Tutor with Rich Pedagogy and Guardrails},
year = {2024},
isbn = {9798400706332},
publisher = {Association for Computing Machinery},
address = {New York, NY, USA},
url = {https://doi.org/10.1145/3657604.3662041},
doi = {10.1145/3657604.3662041},
booktitle = {Proceedings of the Eleventh ACM Conference on Learning @ Scale},
pages = {5–15},
numpages = {11},
location = {Atlanta, GA, USA},
series = {L@S '24}
}

@InProceedings{vocabdiff,
author="Jiao, Ying
and Shridhar, Kumar
and Cui, Peng
and Zhou, Wangchunshu
and Sachan, Mrinmaya",
editor="Wang, Ning
and Rebolledo-Mendez, Genaro
and Matsuda, Noboru
and Santos, Olga C.
and Dimitrova, Vania",
title="Automatic Educational Question Generation with Difficulty Level Controls",
booktitle="Artificial Intelligence in Education",
year="2023",
publisher="Springer Nature Switzerland",
address="Cham",
pages="476--488",
isbn="978-3-031-36272-9"
}

@misc{qwq32b,
    title = {QwQ-32B: Embracing the Power of Reinforcement Learning},
    url = {https://qwenlm.github.io/blog/qwq-32b/},
    author = {Qwen Team},
    month = {March},
    year = {2025}
}

@article{qwen2.5,
      title={Qwen2.5 Technical Report}, 
      author={An Yang and Baosong Yang and Beichen Zhang and Binyuan Hui and Bo Zheng and Bowen Yu and Chengyuan Li and Dayiheng Liu and Fei Huang and Haoran Wei and Huan Lin and Jian Yang and Jianhong Tu and Jianwei Zhang and Jianxin Yang and Jiaxi Yang and Jingren Zhou and Junyang Lin and Kai Dang and Keming Lu and Keqin Bao and Kexin Yang and Le Yu and Mei Li and Mingfeng Xue and Pei Zhang and Qin Zhu and Rui Men and Runji Lin and Tianhao Li and Tianyi Tang and Tingyu Xia and Xingzhang Ren and Xuancheng Ren and Yang Fan and Yang Su and Yichang Zhang and Yu Wan and Yuqiong Liu and Zeyu Cui and Zhenru Zhang and Zihan Qiu},
      journal={arXiv preprint arXiv:2412.15115},
      year={2024}
}

@misc{deepseekai,
      title={DeepSeek-R1: Incentivizing Reasoning Capability in LLMs via Reinforcement Learning}, 
      author={DeepSeek-AI},
      year={2025},
      eprint={2501.12948},
      archivePrefix={arXiv},
      primaryClass={cs.CL},
      url={https://arxiv.org/abs/2501.12948}, 
}

@misc{zhang2024llmsbeathumansdebating,
      title={Can LLMs Beat Humans in Debating? A Dynamic Multi-agent Framework for Competitive Debate}, 
      author={Yiqun Zhang and Xiaocui Yang and Shi Feng and Daling Wang and Yifei Zhang and Kaisong Song},
      year={2024},
      eprint={2408.04472},
      archivePrefix={arXiv},
      primaryClass={cs.CL},
      url={https://arxiv.org/abs/2408.04472}, 
}

@article{Yang01092005,
author = {Ya-Ting C. Yang and Timothy J. Newby and Robert L. Bill and},
title = {Using Socratic Questioning to Promote Critical Thinking Skills Through Asynchronous Discussion Forums in Distance Learning Environments},
journal = {American Journal of Distance Education},
volume = {19},
number = {3},
pages = {163--181},
year = {2005},
publisher = {Routledge},
doi = {10.1207/s15389286ajde1903\_4},


URL = { 
    
        https://doi.org/10.1207/s15389286ajde1903_4
    
    

},
eprint = { 
    
        https://doi.org/10.1207/s15389286ajde1903_4
    
    

}

}

@book{toulmin1984introduction,
  title={An Introduction to Reasoning},
  author={Toulmin, S. and Rieke, R.D. and Janik, A.},
  isbn={9780024211606},
  lccn={83016196},
  url={https://books.google.ch/books?id=FTUQAQAAIAAJ},
  year={1984},
  publisher={Macmillan}
}

@misc{gu2025surveyllmasajudge,
      title={A Survey on LLM-as-a-Judge}, 
      author={Jiawei Gu and Xuhui Jiang and Zhichao Shi and Hexiang Tan and Xuehao Zhai and Chengjin Xu and Wei Li and Yinghan Shen and Shengjie Ma and Honghao Liu and Saizhuo Wang and Kun Zhang and Yuanzhuo Wang and Wen Gao and Lionel Ni and Jian Guo},
      year={2025},
      eprint={2411.15594},
      archivePrefix={arXiv},
      primaryClass={cs.CL},
      url={https://arxiv.org/abs/2411.15594}, 
}

\appendix

\section{Implementation Details}

\subsection{Pseudocode for LFTutor}
We list the pseudocode for the \texttt{LFTutor} framework in algorithm \ref{alg:lftutor}. 

\begin{algorithm}[t]
\footnotesize
\caption{LFTutor}
\label{alg:lftutor}
\begin{algorithmic}
\Require Fallacy $L$, Max turns $T$
\Ensure Tutoring dialogue $\mathcal{D} = \{(A_t, R_t)\}_{t=1}^{T'}$
\State $A_0 \gets \textsc{BeginDialogue}(L)$
\State $R_0 \gets \textsc{StudentResponse}(A_0)$
\State $\mathcal{B} \gets \emptyset$, $C \gets [A_0,R_0]$ \Comment{Disagreement bank; Dialogue history}
\For{$t = 1$ \textbf{to} $T$}
    \State $Cl, Ex, As \gets R_{t-1}$ \Comment{Claim, example, assumption}
    \If{$Cl \in \mathcal{B} \And As \lor Ex \in \mathcal{B} $}  \Comment{Disagreement Check}
        \State $A_t \gets \textsc{RemindStudent}(L, R_{t-1})$;
        \State $C \gets C \oplus \langle A_t \rangle$; Deliver $A_t$
        \State $R_t \gets \textsc{StudentResponse}(C)$; $C \gets C \oplus \langle R_t \rangle$;
        \State \textbf{continue} 
    \EndIf
    \State $\mathcal{B} \gets \mathcal{B} \cup \{(Cl, Ex, As)\}$
    \State $I_t \gets \textsc{IntentDetect}(L, Cl, Ex, As)$
    \State $S_t \gets \textsc{StrategySelect}(L, R_{t-1}, I_t, C)$ 
    \State $A_t \gets \textsc{StrategyExec}(L, R_{t-1}, S_t)$ 
    \While{$\neg \textsc{FollowCheck}(L, A_t, S_t) \lor \neg \textsc{RelevanceCheck}(L, A_t)$}
        \State $A_t \gets \textsc{StrategyRephrase}(L, R_{t-1}, S_t)$
    \EndWhile
    \State $C \gets C \oplus \langle A_t \rangle$; Deliver $A_t$
    \State $R_t \gets \textsc{StudentResponse}(C)$; $C \gets C \oplus \langle R_t \rangle$
\EndFor
\end{algorithmic}
\end{algorithm}

\subsection{Taxonomy of Problems}
We include the taxonomy of problems for LLM agents in education in \cref{fig:taxomony}. We divide the problems into three main categories: educational output, external influences, and structural problems. We briefly cite each problem and their corresponding feedback from annotators in \cref{tab:ref_feed_taxo}.





















\subsection{Pilot Study for LF Education}  \label{ref_pilot_sev}

For the pilot study on problems of LLM teachers in LF Education, we generated 100 dialogues from sentences randomly sampled from the preprocessed Elec2Deb20 (Appendix \ref{dataset}). We instructed GPT-4o to play as the teacher with the system prompt shown in \cref{base_prompt}. For the student, we instructed GPT-4o with the system prompt shown in \cref{prompt-norm-stu}. We set the temperature of both LLMs to 1 to maximize response generation diversity \cite{zhou2024sotopia} to explore possible teaching issues. We also set the maximum dialogue round to 10, which is twice the number of average rounds for dialogues in SocraticLM \cite{liu2024socraticlm}. We chose the dialogue length to be 10 rounds since we would like to investigate LLM teacher's behaviors in prolonged debates over logical validity of given sentences. 

We recruited three human annotators who are fluent in English and have prior experience with linguistic annotation tasks. We first provided them with the descriptions of the problems according to the taxonomy in \cref{fig:taxomony}, then we took sufficient time to explain those problems while confirming that all human annotators fully understand the definitions. These annotators then evaluated the dialogues independently. We recorded problems noted by all three annotators. The annotators then discussed each reported problem and reached a decision to either select or reject it. We showcase the occurrance rate of each reported problem and the annotators' feedback in \cref{tab:ref_feed_taxo}.


\begin{table*}[b]
\scriptsize
  \centering
  \begin{tabular}
  {m{0.07\textwidth}m{0.07\textwidth}m{0.07\textwidth}m{0.07\textwidth}m{0.07\textwidth}m{0.07\textwidth}m{0.07\textwidth}m{0.07\textwidth}m{0.07\textwidth}}
    \toprule
     \textbf{Metric} &\textbf{Divergence}     & \textbf{Stance Change} & \textbf{Repetition} & \textbf{Lack of Refutation} & \textbf{Lack of Evidence Inquiry} & \textbf{Strategy Fixation} & \textbf{Unexplained LF Terms} & \textbf{Passive Guidance}\\ 
    \midrule
    \textbf{F1-Score}&0.76 & 0.82 & 0.68 & 0.71 & 0.74 & 0.71 & 0.77 & 0.72\\
    \bottomrule
  \end{tabular}
  \caption{\label{pilot_qwq}
    F1-Scores for Pilot Studies on Automatic Evaluation using QwQ-32B }
\end{table*}

\begin{table*}[b]
\scriptsize
  \centering
  \begin{tabular}{m{0.07\textwidth}m{0.07\textwidth}m{0.07\textwidth}m{0.07\textwidth}m{0.07\textwidth}m{0.07\textwidth}m{0.07\textwidth}m{0.07\textwidth}m{0.07\textwidth}}
    \toprule
     \textbf{Metric} &\textbf{Divergence}     & \textbf{Stance Change} & \textbf{Repetition} & \textbf{Lack of Refutation} & \textbf{Lack of Evidence Inquiry} & \textbf{Strategy Fixation} & \textbf{Unexplained LF Terms} & \textbf{Passive Guidance}\\ 
    \midrule
    \textbf{Accuracy}& 0.82 & 0.92 & 0.80 & 0.83 & 0.84 & 0.81 & 0.88 & 0.85\\
    \bottomrule
  \end{tabular}
  \caption{\label{pilot_qwq_accuracy}
    Annotation Accuracy of QwQ-32B for Pilot Studies on Automatic Evaluation}
\end{table*}

\subsection{Experiments with Automatic Evaluation} \label{Autoeval_imp_detail}

We used GPT-4o as the base model for all teacher frameworks. For \textsc{Base} and \textsc{Base w/ Problems}, we used prompts from \cref{base_prompt} and \cref{basew_prompt}, respectively, both of which asked the teacher to roleplay an LF Educator. For \texttt{LFTutor}, we combined the strategy chosen at Step 3: Intent-Based Strategy Selection, with a general prompt for executing one exact tutoring strategy (\cref{transition-prompt}), which forms its system prompt. All mechanisms prior to that (Disagreement Check, Intent Detection, etc.) were not included in the system prompt.

In dialogues with normal students, we instructed GPT-4o to simulate a student who debates with the teacher (\cref{prompt-norm-stu}) using debating techniques (\cref{student-strategy}). For each tutor, we generated 1,000 dialogues using randomly sampled sentences from Elec2Deb20. We evaluated dialogues using QwQ-32B as the issue identifier with criteria in \cref{autoeval_fr}. 

In dialogues with adversarial students, we asked GPT-4o to role-play different adversarial students (\cref{prompts-adv-stu}) according to characteristics described in \cref{adv_stu}. For each adversarial student, we generated 1,000 dialogues from both \textsc{Base w/problems} and \texttt{LFTutor}, using randomly sampled sentences from Elec2Deb20. We then evaluated them using QwQ-32B as the issue identifier on the corresponding metric (e.g., evaluating Divergence on dialogues from divergent student). 

\subsection{Prompts for Automatic Evaluation} \label{sec:autoeval}

\subsubsection{Divergence}

\begin{lstlisting}[frame=single, basicstyle=\ttfamily\scriptsize, numbers=none]
You are a judge who looks at a <dialogue> between a teacher and a student. They are discussing the logical validity of <sentence>. 
Definition of divergence: The teacher's responses shifts focus away from the discussion of logical validity of <sentence>. 
Read the dialogue carefully and consider whether the teacher's responses contain instances of divergence. Answer with "yes" if you found at least one response fitting the problem above. Otherwise, answer with "no".
Please only answer with "yes" or "no".
<sentence>: {sentence}
<dialogue>: {history}


\end{lstlisting}

\subsubsection{Stance Change}
\begin{lstlisting}[frame=single, basicstyle=\ttfamily\scriptsize, numbers=none]

You are a judge who looks at a <dialogue> between a teacher and a student. They are discussing the logical validity of <sentence>. 
Definition of stance change: The teacher explicitly shows agreement to the student's argument that the sentence is logically valid, e.g. "I agree...", or "You are right..", and the teacher's initial position that <sentence> has logical fallacy is affected by the student.
Read the dialogue carefully and consider whether the teacher's responses contain instances of stance change. Answer with "yes" if you found at least one response fitting the problem above. Otherwise, answer with "no" only.
Please only answer with "yes" or "no".
<sentence>: {sentence}
<dialogue>: {history}


\end{lstlisting}
\subsubsection{Repetition}
\begin{lstlisting}[frame=single, basicstyle=\ttfamily\scriptsize, numbers=none]
You are a judge who looks at a <dialogue> between a teacher and a student. They are discussing the logical validity of <sentence>. 
Definition of Repetition: The teacher paraphrases the student's responses or simply repeats the student's responses, without further explanations or expansions into related topics. 
Read the dialogue carefully and consider whether the teacher's responses contain instances of Repetition. Answer with "yes" if you found at least one response fitting the problem above. Otherwise, answer with "no" only.
Please only answer with "yes" or "no".
<sentence>: {sentence}
<dialogue>: {history}

\end{lstlisting}
\subsubsection{Lack of Refutation}
\begin{lstlisting}[frame=single, basicstyle=\ttfamily\scriptsize, numbers=none]
You are a judge who looks at a <dialogue> between a teacher and a student. They are discussing the logical validity of <sentence>. 
Definition of Lack of Refutation: The teacher has NO responses where it challenges the student by asking student explicit questions to provide assumptions or by providing counterexamples WITHOUT the student's request.
Read the dialogue carefully and consider whether the teacher's responses contains refutation. Answer with "yes" if you found at least one response fitting the problem above. Otherwise, answer with "no" only.
Please only answer with "yes" or "no".
<sentence>: {sentence}
<dialogue>: {history}

\end{lstlisting}
\subsubsection{Lack of Evidence Inquiry}
\begin{lstlisting}[frame=single, basicstyle=\ttfamily\scriptsize, numbers=none]
You are a judge who looks at a <dialogue> between a teacher and a student. They are discussing the logical validity of <sentence>. 
Definition of Lack of Evidence Inquiry: The teacher does NOT explicitly request the student to provide examples or evidence that supports the logical validity of <sentence>.
Read the dialogue carefully and consider whether the teacher's responses contains Evidence Inquiry. Answer with "yes" if you found at least one response fitting the problem above. Otherwise, answer with "no" only.
Please only answer with "yes" or "no".

<sentence>: {sentence}
<dialogue>: {history}

\end{lstlisting}
\subsubsection{Strategy Fixation}
\begin{lstlisting}[frame=single, basicstyle=\ttfamily\scriptsize, numbers=none]
You are a judge who looks at a <dialogue> between a teacher and a student. They are discussing the logical validity of <sentence>. 
Definition of Strategy Fixation: The teacher emphasizes general strategies such as considering "balanced perspectives" or "broader context" without linking such strategies to the exact problem with the validity of <sentence>.
Read the dialogue carefully and consider whether the teacher's responses contain instances of Strategy Fixation. Answer with "yes" if you found at least one response fitting the problem above. Otherwise, answer with "no" only.
Please only answer with "yes" or "no".
<sentence>: {sentence}
<dialogue>: {history}


\end{lstlisting}
\subsubsection{Unexplained LF Terms}
\begin{lstlisting}[frame=single, basicstyle=\ttfamily\scriptsize, numbers=none]
You are a judge who looks at a <dialogue> between a teacher and a student. They are discussing the logical validity of <sentence>. 
Definition of term-reliance: The teacher mentions terms of logical fallacy e.g. strawman argument, ad hominem, without explaining these terms' definitions clearly or relating them to the logical validity of <sentence>.
Read the dialogue carefully and consider whether the teacher's responses contain instances of term-reliance. Answer with "yes" if you found at least one response fitting the problem above. Otherwise, answer with "no" only.
Please only answer with "yes" or "no".
<sentence>: {sentence}
<dialogue>: {history}
\end{lstlisting}
\subsubsection{Passive Guidance}
\begin{lstlisting}[frame=single, basicstyle=\ttfamily\scriptsize, numbers=none]
You are a judge who looks at a <dialogue> between a teacher and a student. They are discussing the logical validity of <sentence>. 
Definition of passive guidance: For the dialogue, the teacher primarily follows the student's lead rather than providing clear direction on how to analyze logical validity of <sentence>.
Read the dialogue carefully and consider whether the teacher's responses fit the description of passive guidance. Answer with "yes" if you find the dialogue overall fitting the description. Otherwise, answer with "no" only.
Please only answer with "yes" or "no".
<sentence>: {sentence}
<dialogue>: {history}

\end{lstlisting}

\subsection{Cost Overview}

We recorded the cost with OPENAI API after each experiment, and we aggregated the cost for simulated dialogue experiments with GPT-4o. The total cost from all experiments including prompt tuning and reasoning model selection amounted to $\$2168.8$. Generating 1,000 dialogues with the baseline framework cost about \$80, while generating 1,000 dialogues with \texttt{LFTutor} framework cost roughly \$106. For automatic evaluation, we recorded the cost of using QwQ-32B. The total cost for evaluation in all experiments came to approximately \$57. A full evaluation of 1,000 dialogues cost roughly \$7.

In addition to experiments above, we conducted ablation studies on GPT-4o, which cost roughly \$60, and additional experiments on LLaMA-3.1-405B and Gemini-2.5-pro, which cost roughly \$180. 

\section{Datasets Used} \label{dataset}

\subsection{Dataset for Main Experiments}
For dialogue generation, we use Elec2Deb20 \cite{goffredo-etal-2023-argument}, which contains scripts from presidential candidates in television debates from 1960 to 2020. Elec2Deb20 includes over 1800 logical fallacy examples in six categories: Appeal to Authority, Ad Hominem, Appeal to Emotion, Slippery Slope, False Cause, and Slogan. We believe that this dataset is representative of sentences with LF discussed on the internet, since the presidential debates cover many actively debated topics in the United States, including healthcare, climate change, clean energy, etc. Compared with logical fallacy datasets such as LOGIC \cite{jin-etal-2022-logical} or LFUD \cite{li-etal-2024-reason}, whose examples are either LLM-generated or taken from educational websites, Elec2Deb20's sentences are more grounded in the real world as they are directly taken from political debates with robust human annotation. 

For clarity and ease of understanding for LLMs, we employed the full context of sentences from Elec2Deb20. We excluded the category of Slogans, considering that these arguments are usually promises that cannot be verified when the speaker made them. We also excluded the subcategory "flag waving" as justifications based on patriotism is highly subjective and hard to be discussed using objective evidence. Finally, we deleted sentences with 30 words or less, since most of these sentences lack the necessary context for determining logical validity. The preprocessed dataset contains 1317 samples, with the detailed number per LF category shown in \cref{elec2deb20}. 

In addition to the experiments mentioned in the main text, we included results for separate logical fallacy categories in \cref{res_diff_cats}. 

\subsection{Additional Dataset}
In addition to the logical fallacy examples in Elec2Deb20, we tested Ad Populum, False Dilemma, and Hasty Generalization from COCOLoFa \cite{yeh-etal-2024-cocolofa}, which includes over 7,000 logical fallacies from various news comments. We randomly selected 50 examples per chosen category, and we test \textsc{BASE}, \textsc{BASE w/ Problems}, and \texttt{LFTutor} using the normal student setting discussed in \cref{Autoeval_imp_detail}. Results and discussions are shown in \cref{res_diff_cats}.

\section{Experiments with other LLMs} \label{other_llm_exps}

To further evaluate the performance of \texttt{LFTutor} on a wider range of LLMs, we conducted experiments with \textsc{Base} and \texttt{LFTutor} on LLaMA-3.1-405B \citep{llama3herdmodels} and Gemini-2.5-pro \citep{gemini25pushingfrontier}. We used the experiment setting for normal students discussed in \cref{exp-auto}, and we generated dialogues using 200 sentences randomly selected from Elec2Deb20. 

Results for LLaMA-3.1-405B are shown in \cref{tab:res_llama}. \texttt{LFTutor} improves average performance by about 17.3\% compared with \textsc{Base}, with increases in Stance Change, Lack of Refutation, Lack of Evidence Inquiry, Strategy Fixation, and Unexplained LF Terms. \texttt{LFTutor} suffers performance loss in Divergence and Passive Guidance, with slight decrease of scores in Repetition. We observe that LLaMA-3.1-405B is weaker in instruction-following compared with GPT-4o, resulting in worse overall performance. Often, LLaMA failed to follow the education strategies given by the prompts, resulting in lack of detailed explanations of logical flaws and lack of effective Socratic Questions. These contributed to the low score in Passive Guidance and Divergence.

Results for Gemini-2.5-pro are shown in \cref{tab:res_gemini}. We observe that \textsc{Base} with Gemini-2.5-pro as backbone already achieves strong performance, which may be attributed to its advanced architectures. \texttt{LFTutor} improves upon \textsc{Base} by 8.4\% on average, with significant improvements in Stance Change, Lack of Evidence Inquiry, Unexplained LF Terms, and Passive Guidance, as well as improvements in Lack of Refutation and Strategy Fixation. The decrease in scores for Repetition and Divergence is possibly due to the design of the mechanisms (reminding the student to propose examples/assumptions) for disagreement check. 

\begin{table}[b]
\small
  \centering
  \resizebox{\linewidth}{!}{
  \begin{tabular}{m{3.5cm} m{0.8cm} m{1.5cm} m{1.1cm}}
    \toprule
    \textbf{Evaluation Metrics} & \textsc{Base} & \texttt{LFTutor}\\
    \midrule
    \textbf{Divergence} &  \textbf{83} & 71 \\
    \textbf{Stance Change}  & 32 & \textbf{79}    \\
    \textbf{Repetition}  & \textbf{88} & 86     \\
    \textbf{Lack of Refutation}  & $80$ & \textbf{95}     \\
    \textbf{Lack of Evidence Inquiry} & 24 & \textbf{79} \\
    \textbf{Strategy Fixation} & $68$& \textbf{92}    \\
    \textbf{Unexplained LF Terms}& $66$ & \textbf{84}  \\
    \textbf{Passive Guidance} & \textbf{44}& 37 \\
    \textbf{Avg. Performance} & $60.6$& \textbf{77.9}   \\
    \bottomrule
  \end{tabular}}
   \vspace{-5pt}
  \caption{Percentage of dialogues with normal students for each framework passing each metric, conducted with LLaMA-3.1-405B as teacher. Evaluated using QwQ-32B as a judge. Higher is better.  }
  \label{tab:res_llama}
  \vspace{-10pt}
\end{table}

\begin{table}[b]
\small
  \centering
  \resizebox{\linewidth}{!}{
  \begin{tabular}{m{3.5cm} m{0.8cm} m{1.5cm} m{1.1cm}}
    \toprule
    \textbf{Evaluation Metrics} & \textsc{Base} & \texttt{LFTutor}\\
    \midrule
    \textbf{Divergence} &  \textbf{97} & 86 \\
    \textbf{Stance Change}  & 70 & \textbf{87}    \\
    \textbf{Repetition}  & \textbf{98} & 62     \\
    \textbf{Lack of Refutation}  & $94$ & \textbf{100}     \\
    \textbf{Lack of Evidence Inquiry} & 56 & \textbf{98} \\
    \textbf{Strategy Fixation} & $92$& \textbf{94}    \\
    \textbf{Unexplained LF Terms}& $56$ & \textbf{92}  \\
    \textbf{Passive Guidance} & 73& \textbf{84} \\
    \textbf{Avg. Performance} & $79.5$& \textbf{87.9}   \\
    \bottomrule
  \end{tabular}}
   \vspace{-5pt}
  \caption{Percentage of dialogues with normal students for each framework passing each metric, conducted with Gemini-2.5-pro as teacher. Evaluated using QwQ-32B as a judge. Higher is better.  }
  \label{tab:res_gemini}
  \vspace{-10pt}
\end{table}

\section{Performance for Separate Logical Fallacy Categories in Normal Student Setting}
\label{res_diff_cats}

\subsection{Overview}
For Elec2Deb20, we include the performance of different frameworks for each LF category in \cref{tab:separate_cats_1}, \cref{tab:separate_cats_2}, \cref{tab:separate_cats_3}, \cref{tab:separate_cats_4}, and \cref{tab:separate_cats_5}. 

For COCOLoFa, we include the performance of different frameworks for each LF category in \cref{tab:separate_cats_6}, 
\cref{tab:separate_cats_7}, and \cref{tab:separate_cats_8}.

\subsection{Analysis of Results}

We observe that \texttt{LFTutor} still significantly outperforms \textsc{Base} and \textsc{Base w/ Problems} for every LF category. For most logical fallacy categories, \texttt{LFTutor} performs slightly worse than \textsc{Base w/ Problems} in Lack of Refutation, while \texttt{LFTutor} occasionally underperforms \textsc{Base w/ Problems} in Passive Guidance. 

Regarding \textbf{Lack of Refutation}, we attribute this slight loss of performance for \texttt{LFTutor} to the lower priority of refutation strategy in intent-based strategy selection. As \texttt{LFTutor} primarily focuses on individual components of the student's argument rather than the entire argument, it is possible that refutation strategy is not applied for the entire conversation. \textsc{Base w/ Problems}, on the other hand, tends to immediately follow the instructions given (avoiding common problems in logical fallacy education) through utilizing refutation in the first 5 rounds of conversation, according to our analysis on randomly sampled dialogues. 

Regarding \textbf{Passive Guidance}, we posit that \texttt{LFTutor} predominantly utilizes the refutation strategy when discussing texts with Appeal to Authority or False Cause. These logical fallacies tend to make factual errors when attempting to bridge the gap between the claim and grounds. When the student tries to strengthen these arguments further, \texttt{LFTutor} placed refutation with higher priority in order to tackle factual errors. Refutation is treated as passively responding to the student by the LLM judge, thus leading to worse performance in Passive Guidance. 

In addition to the observations above, we note that \textsc{Base}, \textsc{Base w/ Problems}, and \texttt{LFTutor} achieve higher performance in dialogues with texts from CoCoLoFa compared with those in Elec2Deb20. We suspect that the complexity of texts with logical fallacy have significant impact on the performance for all frameworks. CoCoLoFa mainly consists of  comments from news articles, which have simpler forms, shorter text length, and clearer claims with topics corresponding to the news articles. In contrast, texts in Elec2Deb20 consists of excerpts from presidential debates, whose very purpose was to utilize various techniques, including logical fallacies, to confound voters who make decisions on which candidates to choose. Thus, texts in Elec2Deb20 often feature longer sentences mixed with factually correct statements and logical fallacies. These sentences are more difficult to examine for baselines LLMs and humans alike, leading to lower performance for all LLM tutors. 

\begin{table}[t]
\small
  \centering
  \resizebox{\linewidth}{!}{
  \begin{tabular}{m{3.5cm} m{0.8cm} m{1.5cm} m{1.1cm}}
    \toprule
    \textbf{Evaluation Metrics} & \textsc{Base} & \textsc{Base w/ problems} & \texttt{LFTutor}\\
    \midrule
    \textbf{Divergence}  & $65.4$ &$69.6$ &  \textbf{82.3}  \\
    \textbf{Stance Change}  & $14.1$ & $41.8$& \textbf{84.8}    \\
    \textbf{Repetition}  & $39.7$&$38.0$ &\textbf{78.5}        \\
    \textbf{Lack of Refutation}  & $24.4$ & \textbf{100.0} & $98.7$       \\
    \textbf{Lack of Evidence Inquiry} & $15.4$ & $92.4$& \textbf{96.2} \\
    \textbf{Strategy Fixation} & $25.3$& $58.1$& \textbf{94.9}    \\
    \textbf{Unexplained LF Terms}  & $51.3$ & $21.5$ & \textbf{91.1}  \\
    \textbf{Passive Guidance} & $3.8$& \textbf{36.7} & \textbf{36.7}  \\
    \textbf{Avg. Performance} & $30.0$ & $57.3$& \textbf{82.9} \\
    \bottomrule
  \end{tabular}}
   \vspace{-5pt}
  \caption{Percentages of dialogues with normal students for each framework passing each metric for Ad Hominem. Evaluated using QwQ-32B as a judge. Higher is better.  }
  \label{tab:separate_cats_1}
  \vspace{-10pt}
\end{table}

\begin{table}[t]
\small
  \centering
  \resizebox{\linewidth}{!}{
  \begin{tabular}{m{3.5cm} m{0.8cm} m{1.5cm} m{1.1cm}}
    \toprule
    \textbf{Evaluation Metrics} & \textsc{Base} & \textsc{Base w/ problems} & \texttt{LFTutor}\\
    \midrule
    \textbf{Divergence}  & $78.6$ &$71.5$ &  \textbf{82.1}  \\
    \textbf{Stance Change}  & $10.3$ & $48.8$& \textbf{90.2}    \\
    \textbf{Repetition}  & $38.9$&$38.2$ &\textbf{81.3}        \\
    \textbf{Lack of Refutation}  & $14.3$ & \textbf{100.0} & $99.2$       \\
    \textbf{Lack of Evidence Inquiry} & $7.1$ & \textbf{93.5}& \textbf{93.5} \\
    \textbf{Strategy Fixation} & $40.0$& $67.3$& \textbf{90.2}    \\
    \textbf{Unexplained LF Terms}  & $50.0$ & $30.1$ & \textbf{95.1}  \\
    \textbf{Passive Guidance} & $3.2$& \textbf{47.2} & $45.3$   \\
   \textbf{Avg. Performance} & $30.3$& $62.1$ & \textbf{84.6}   \\
    \bottomrule
  \end{tabular}}
   \vspace{-5pt}
  \caption{Percentages of dialogues with normal students for each framework passing each metric for Appeal to Authority. Evaluated using QwQ-32B as a judge. Higher is better.}
  \label{tab:separate_cats_2}
  \vspace{-10pt}
\end{table}

\begin{table}[t]
\small
  \centering
  \resizebox{\linewidth}{!}{
  \begin{tabular}{m{3.5cm} m{0.8cm} m{1.5cm} m{1.1cm}}
    \toprule
    \textbf{Evaluation Metrics} & \textsc{Base} & \textsc{Base w/ problems} & \texttt{LFTutor}\\
    \midrule
    \textbf{Divergence}  & $68.2$ &$74.0$ &  \textbf{85.5}  \\
    \textbf{Stance Change}  & $7.0$ & $46.2$& \textbf{86.7}    \\
    \textbf{Repetition}  & $46.9$&$43.5$ &\textbf{78.2}        \\
    \textbf{Lack of Refutation}  & $10.5$ & \textbf{99.8} & \textbf{99.8}       \\
    \textbf{Lack of Evidence Inquiry} & $9.9$ & $96.2$ & \textbf{97.2} \\
    \textbf{Strategy Fixation} & $34.6$& $61.8$& \textbf{92.1}    \\
    \textbf{Unexplained LF Terms}  & $49.9$ & $31.5$ & \textbf{96.2}  \\
    \textbf{Passive Guidance} & $10.8$& $41.5$ & \textbf{45.8}   \\
    \textbf{Avg. Performance} & $29.7$& $61.8$ & \textbf{85.2}   \\
    \bottomrule
  \end{tabular}}
   \vspace{-5pt}
  \caption{Percentages of dialogues with normal students for each framework passing each metric for Appeal to Emotion. Evaluated using QwQ-32B as a judge. Higher is better.}
  \label{tab:separate_cats_3}
  \vspace{-10pt}
\end{table}

\begin{table}[t]
\small
  \centering
  \resizebox{\linewidth}{!}{
  \begin{tabular}{m{3.5cm} m{0.8cm} m{1.5cm} m{1.1cm}}
    \toprule
    \textbf{Evaluation Metrics} & \textsc{Base} & \textsc{Base w/ problems} & \texttt{LFTutor}\\
    \midrule
    \textbf{Divergence}  & $75.6$ &$80.6$ &  \textbf{83.3}  \\
    \textbf{Stance Change}  & $19.5$ & $52.8$& \textbf{88.9}    \\
    \textbf{Repetition}  & $46.3$&$33.3$ &\textbf{77.8}        \\
    \textbf{Lack of Refutation}  & $12.2$ & \textbf{100.0} & \textbf{100.0}       \\
    \textbf{Lack of Evidence Inquiry} & $9.8$ & \textbf{94.4}& \textbf{94.4} \\
    \textbf{Strategy Fixation} & $24.0$& $56.7$& \textbf{88.9}    \\
    \textbf{Unexplained LF Terms}  & $48.8$ & $27.8$ & \textbf{97.2}  \\
    \textbf{Passive Guidance} & $0.0$& \textbf{36.1} & $35.8$   \\
   \textbf{Avg. Performance} & $33.2$& $60.2$ & \textbf{83.3}   \\
    \bottomrule
  \end{tabular}}
   \vspace{-5pt}
  \caption{Percentages of dialogues with normal students for each framework passing each metric for False Cause. Evaluated using QwQ-32B as a judge. Higher is better.}
  \label{tab:separate_cats_4}
  \vspace{-10pt}
\end{table}

\begin{table}[t]
\small
  \centering
  \resizebox{\linewidth}{!}{
  \begin{tabular}{m{3.5cm} m{0.8cm} m{1.5cm} m{1.1cm}}
    \toprule
    \textbf{Evaluation Metrics} & \textsc{Base} & \textsc{Base w/ problems} & \texttt{LFTutor}\\
    \midrule
    \textbf{Divergence}  & $62.5$ &$80.0$ &  \textbf{96.7}  \\
    \textbf{Stance Change}  & $6.2$ & $63.3$& \textbf{96.7}    \\
    \textbf{Repetition}  & $59.4$&$43.3$ &\textbf{76.7}        \\
    \textbf{Lack of Refutation}  & $25.0$ & \textbf{100.0} & \textbf{100.0}       \\
    \textbf{Lack of Evidence Inquiry} & $9.4$ & 96.7& \textbf{100.0} \\
    \textbf{Strategy Fixation} & $32.0$& $53.3$& \textbf{93.3}    \\
    \textbf{Unexplained LF Terms}  & $43.8$ & $6.7$ & \textbf{86.7}  \\
    \textbf{Passive Guidance} & $3.1$& \textbf{43.3} & \textbf{43.3}   \\
    \textbf{Avg. Performance} & $30.2$& $60.8$ & \textbf{86.7}   \\
    \bottomrule
  \end{tabular}}
   \vspace{-5pt}
  \caption{Percentages of dialogues with normal students for each framework passing each metric for Slippery Slope. Evaluated using QwQ-32B as a judge. Higher is better.}
  \label{tab:separate_cats_5}
  \vspace{-10pt}
\end{table}

\begin{table}[t]
\small
  \centering
  \resizebox{\linewidth}{!}{
  \begin{tabular}{m{3.5cm} m{0.8cm} m{1.5cm} m{1.1cm}}
    \toprule
    \textbf{Evaluation Metrics} & \textsc{Base} & \textsc{Base w/ problems} & \texttt{LFTutor}\\
    \midrule
    \textbf{Divergence}  & $78$ &\textbf{86} &  \textbf{86}  \\
    \textbf{Stance Change}  & $18$ & $34$& \textbf{84}    \\
    \textbf{Repetition}  & $80$&$72$ &\textbf{86}        \\
    \textbf{Lack of Refutation}  & $46$ & $98$ & \textbf{100}       \\
    \textbf{Lack of Evidence Inquiry} & $0$ & $90$& \textbf{100} \\
    \textbf{Strategy Fixation} & $66$& $60$& \textbf{90}    \\
    \textbf{Unexplained LF Terms}  & $48$ & $66$ & \textbf{76}  \\
    \textbf{Passive Guidance} & $42$& $68$ & \textbf{80}   \\
    \textbf{Avg. Performance} & $47.3$& $71.8$ & \textbf{87.8}   \\
    \bottomrule
  \end{tabular}}
   \vspace{-5pt}
  \caption{Percentages of dialogues with normal students for each framework passing each metric for Ad Populum. Evaluated using QwQ-32B as a judge. Higher is better.}
  \label{tab:separate_cats_6}
  \vspace{-10pt}
\end{table}

\begin{table}[t]
\small
  \centering
  \resizebox{\linewidth}{!}{
  \begin{tabular}{m{3.5cm} m{0.8cm} m{1.5cm} m{1.1cm}}
    \toprule
    \textbf{Evaluation Metrics} & \textsc{Base} & \textsc{Base w/ problems} & \texttt{LFTutor}\\
    \midrule
    \textbf{Divergence}  & $80$ &$82$ &  \textbf{90}  \\
    \textbf{Stance Change}  & $20$ & $36$& \textbf{90}    \\
    \textbf{Repetition}  & $86$&\textbf{98} &$92$        \\
    \textbf{Lack of Refutation}  & $62$ & \textbf{98} & $96$      \\
    \textbf{Lack of Evidence Inquiry} & $2$ & 92& \textbf{94} \\
    \textbf{Strategy Fixation} & $50$& $66$& \textbf{86}    \\
    \textbf{Unexplained LF Terms}  & $44$ & $50$ & \textbf{64}  \\
    \textbf{Passive Guidance} & $44$& $68$ & \textbf{76}   \\
    \textbf{Avg. Performance} & $48.5$& $73.8$ & \textbf{86.0}   \\
    \bottomrule
  \end{tabular}}
   \vspace{-5pt}
  \caption{Percentages of dialogues with normal students for each framework passing each metric for Hasty Generalization. Evaluated using QwQ-32B as a judge. Higher is better.}
  \label{tab:separate_cats_7}
  \vspace{-10pt}
\end{table}

\begin{table}[t]
\small
  \centering
  \resizebox{\linewidth}{!}{
  \begin{tabular}{m{3.5cm} m{0.8cm} m{1.5cm} m{1.1cm}}
    \toprule
    \textbf{Evaluation Metrics} & \textsc{Base} & \textsc{Base w/ problems} & \texttt{LFTutor}\\
    \midrule
    \textbf{Divergence}  & $82$ &$78$ &  \textbf{84}  \\
    \textbf{Stance Change}  & $34$ & $50$& \textbf{76}    \\
    \textbf{Repetition}  & $90$& \textbf{94}  &$94$       \\
    \textbf{Lack of Refutation}  & $78$ & $96$ & \textbf{98}       \\
    \textbf{Lack of Evidence Inquiry} & $0$ & \textbf{94}& \textbf{94} \\
    \textbf{Strategy Fixation} & $64$& $58$& \textbf{80}    \\
    \textbf{Unexplained LF Terms}  & $58$ & $56$ & \textbf{64}  \\
    \textbf{Passive Guidance} & $46$& $64$ & \textbf{78}   \\
    \textbf{Avg. Performance} & $56.5$& $73.8$ & \textbf{83.5}   \\
    \bottomrule
  \end{tabular}}
   \vspace{-5pt}
  \caption{Percentages of dialogues with normal students for each framework passing each metric for False Dilemma. Evaluated using QwQ-32B as a judge. Higher is better.}
  \label{tab:separate_cats_8}
  \vspace{-10pt}
\end{table}

\section{Prompts for Baseline Teacher} \label{base_teacher}

\subsection{Baseline} \label{base_prompt}
\begin{lstlisting}[frame=single, basicstyle=\ttfamily\scriptsize, numbers=none]
You are a teacher who knows logical fallacies. You are interacting with a student who believes in <sentence>. Be aware that the student may have strong bias towards <sentence>.
Think carefully before fomulating your response. You think that <sentence> is logically invalid. Talk to the student and try to convince the student that <sentence> is logically invalid. Make sure to formulate your response to be readable and understandable by a real student.
Limit your response to 60 words or less.
<sentence>: {sentence}
\end{lstlisting}

\subsection{Baseline w/problems} \label{basew_prompt}
\begin{lstlisting}[frame=single, 
basicstyle=\ttfamily\scriptsize, numbers=none]
You are a teacher who knows logical fallacies. You are interacting with a student who believes in <sentence>. Be aware that the student may have strong bias towards <sentence>.
Think carefully before fomulating your response. You think that <sentence> is logically invalid. Talk to the student and try to convince the student that <sentence> is logically invalid. Make sure to formulate your response to be readable and understandable by a real student.
Try to avoid the following problems when talking to the student:
- Not asking the student to provide examples to support their claim
- Not challenging the student by providing counterexamples or counterarguments.
- Emphasizing broader perspective or broader context without referring to problems of <sentence>
- Changing your original stance by agreeing to the student
- Repeating or rephrasing the student's word without further explanation
- Affected by the student through shifts of focus away from the discussion of logical validity of <sentence>.
- Mentions terms of logical fallacy without explaining these terms' definitions clearly.
- Follows the student's lead rather than providing clear direction in analysis
Limit your response to 60 words or less.

<sentence>: {sentence}

\end{lstlisting}

\begin{table*}
\small
  \centering
  \begin{tabular}{c c c c c c}
    \hline
    \textbf{Ad Hominem}           & \textbf{Appeal to Authorities} & \textbf{Appeal to Emotion} & \textbf{False Cause} & \textbf{Slippery Slope} & \textbf{Total}\\ 
    \hline
    196 & 255 & 721 & 87 & 63 & 1317 \\
    \hline
  \end{tabular}
  \caption{\label{elec2deb20}
    Count of LF Examples from preprocessed Elec2Deb20.
  }
\end{table*}
\section{Prompts for \texttt{LFTutor} \label{sec:promptlftutor}}

\subsection{Disagreement Check \label{disagreement-bank}}

\begin{lstlisting}[frame=single, basicstyle=\ttfamily\scriptsize, numbers=none]

The student and teacher are discussing about the logical validity of <sentence>. Please answer the following questions.Please address the student by the second person pronoun "you".
Q1. Check if the student's <response> from last turn is relevant to the discussions of logical validity of <sentence>. If yes, answer with yes, then give a summary of the student's words including the student's example if possible, in 20 words. If no, answer with no, then give your reason in 15 words or less.
Q2. Check whether the student's response is included in <history>. Note that it has to match the contents discussed in <history>. If yes, answer with yes first, then provide the item. If the student's response is irrelevant to any of them, answer with no, then give your reason in 15 words or less.
Q3. Check if the student's utterance in <response> is already included in <bank>. If yes, answer with yes, then give your reason in 15 words or less. If no, ONLY answer with "no".
Q4. If the student makes an example/assumption/request, check if the student's example/assumption/request appears in <history>. Answer with "yes" or "no", and give your reason in 15 words or less. 
Q5. Is the student making an example or assumption in the response? If yes, answer with yes, then summarize the example or assumption in 15 words or less. Otherwise, answer with "no" only. 
<sentence>: {sentence} # Logical Fallacy
<history>: {history} # dialogue history
<response>: {profile} #student's response from last turn
<bank>: {target_statement} #disagreement bank

format your answer in JSON with the following component: "Q1": <answer_to_Q1>, "Q2": <answer_to_Q2>, "Q3": <answer_to_Q3>, "Q4": <answer_to_Q4>, "Q5": <answer_to_Q5>

\end{lstlisting}

\subsection{Intent Detection \label{intent_prompt}}

\begin{lstlisting}[frame=single, basicstyle=\ttfamily\scriptsize, numbers=none]
You are an experienced teacher who knows how to debate, and you are interacting with student named [I], on discussing logical validity of <sentence>.
Remember, the topic you are discussing on is the logical validity of <sentence>. You have to maintain your position and try not to be convinced by the student.
Consider the student's response in <history>, and answer the following questions:

Q1: Treating the student's response as a counterargument to your stance, does the student make an argument without presenting enough evidence that supports it?
Q2: Treating the student's response or example as a counterargument to your stance, does the student present argument or example with clear logical flaws?
Q3: Is the student requesting you to provide evidence or explanation?
Q4: Treating the student's response as a counterargument to your stance, does the student's argument need more assumptions to clarify?

For each question, answer with "yes" or "no". Format your answer in JSON with the following key: "1": <answer to Q1>, "2": <answer to Q2>, "3": <answer to Q3>, "4": <answer to Q4> "5": <answer to Q5>
<sentence>: {sentence}
<history>: {history}

\end{lstlisting}

\subsection{Tutoring Strategies \label{edustrat_prompt}}
\begin{lstlisting}[frame=single, basicstyle=\ttfamily\scriptsize, numbers=none]
STRAT_FOR_STATES_R = {
    "ResT": """First, tell the student the right way to demonstrate logical validity of <sentence>. Then, respond to the student's request on providing evidence or clarifications, and give support to your stance. """,
    
    "EviT": """Treating the student's response as counterargument to your stance, tell the student the right way to demonstrate logical validity of <sentence> e.g. 'logical validity hinges on ...', and point out the logical flaw with the student's example/assumption, finally request the student to provide evidence that supports his claim. e.g. Can you provide examples...""",
    
    "SumT": """Treating the student's response as counterargument to your stance, tell the student the right way to demonstrate logical validity of <sentence> e.g. 'logical validity hinges on ...', and point out the logical flaw with the student's example/assumption, finally request the student about their assumptions in their arguments. e.g. 'Why do you assume...' or 'How do you know...'. 
    """,
    
    "RefT": """
    First, Show all necessary conditions for the argument to hold logically valid. Then, refute the student's argument using one of the following strategy. If possible, also include counterargument/counterexamples in your response.
    a. Showing that the argument's conclusion or premise is wrong. Provide a counterargument or counterexample to illustrate your point.
    b. Showing that the argument's conclusion does not follow from the premise. Provide a counterargument or counterexample to illustrate your point.
    c. Showing that the student's argument is irrelevant to the topic of discussion. Even if the evidence provided is valid, it may be irrelevant to the logical validity of <sentence>, and thus can be dismissed.
    """,
}

\end{lstlisting}

\subsection{Execution of Tutoring Strategy (using Evidence strategy as an example) }  \label{transition-prompt}
\begin{lstlisting}[frame=single, basicstyle=\ttfamily\scriptsize, numbers=none]
You are an experienced teacher who knows how to debate, and you are interacting with student named [I], on discussing logical validity of <sentence>.
Think about the flaws in the student's reponse. You don't think that <sentence> is logically valid. 

Treating the student's response as counterargument to your stance, tell the student the right way to demonstrate logical validity of <sentence> e.g. 'logical validity hinges on ...', and point out the logical flaw with the student's example/assumption, finally request the student to provide evidence that supports his claim. e.g. Can you provide examples... 

Remember, the topic you are discussing on is the logical validity of <sentence>. You have to maintain your position and try not to be convinced by the student. Keep your tone calm and do not use exclamations, and respond in a way that is similar to everyday conversation. 
When formulating your response, do NOT mention specific terms of logical fallacy e.g. ad hominem, strawman, etc.
You are given a fixed option above, which you need to follow. Use the option above and respond to the student, and DO NOT ask additional questions besides strictly following the option. Keep your response concise. Limit your response to 60 words or less.

<sentence>: {sentence}
\end{lstlisting}

\subsection{Strategy-following Check } \label{check-res-pr}
\begin{lstlisting}[frame=single, basicstyle=\ttfamily\scriptsize, numbers=none]
You are a judge overlooking the dialogue between a teacher and a student, they are having a debate over the logical validity of <sentence>.
Based on the teacher's <response>, answer the following questions.
Q1. Check if the teacher has followed <strategy> in formulating their response. The teacher is following <strategy> as long as any sentence in their response contain such strategy.
Q2. If the teacher asks the student a question, is the question still helpful for determining the logical validity of <sentence>? Also Answer "yes" if there is no question provided.
For each question, answer with "yes" or "no" only. Format your answer in JSON with the following key: "1": <answer to Q1>, "2": <answer to Q2>
<sentence>: {sentence}
<response>: {history}
<strategy>: {profile}
\end{lstlisting}


\section{Strategies and Prompts for LLM Student \label{sec:studentdesign}}

\subsection{Overview}

Table \ref{student-strategy} shows the list of strategies for the normal student to choose from. Table \ref{adv_stu} shows the design of persona for each adversarial student. The rest of this section covers the prompts for the normal student and all adversarial students.

\subsection{Normal Student }\label{prompt-norm-stu}
\begin{lstlisting}[frame=single, basicstyle=\ttfamily\scriptsize, numbers=none]
You are an experienced student who knows how to debate, and you are interacting with teacher named [I], on discussing logical validity of <sentence>.
You think that <sentence> is logically valid, and you are trying to defend your position. <behavior> indicates the teacher's most possible behavior.
As an experienced debater, you have the following options to choose from:
1. Have alternative ways of interpreting the dialogue as valid.
2. Respond to the teacher's claim by providing counterexamples.
3. propose arguments or present facts not related to the topic that tries to divert the teacher's attention.
4. Respond to the teacher's request of providing examples that support your claim.
5. Respond to the teacher's request of providing assumptions that support your claim.
6. if the teacher is NOT ASKING QUESTIONS about assumptions or evidence, Request the teacher to provide evidence that support their claim, 
7. Attacking the teacher by pointing out similarities of their argument with yours.
\end{lstlisting}
\subsection{Adversarial Students }\label{prompts-adv-stu}

\myparagraph{Divergence}
\begin{lstlisting}[frame=single, basicstyle=\ttfamily\scriptsize, numbers=none]
You are a student who likes to ask irrelevant questions, and you are interacting with teacher named [I], on discussing logical validity of <sentence>. 
You believe that <sentence> is logically valid, but you do not want to argue with the teacher. You have the following options below:
1. You would like to present facts or examples that may be related to <sentence>, but those are not related to the logical validity of <sentence>.
2. You would like to discuss anything that is not related to the sentence. 
3. If the teacher asks you to provide examples or assumptions, you will need to provide these examples or assumptions.
Remember, you believe that <sentence> is logically valid, but you are roleplaying someone who want to divert the teacher's attention as much as possible. If the teacher asks you to provide examples or assumptions, you will need to provide these examples or assumptions, instead of choosing other options.
Respond to the teacher. Try not to be convinced by the teacher and keep your response in 50 words or less. Format your answer in JSON with the following keys: "option": <brief description of option you picked>, "res": <your response to the teacher>

<sentence>: {sentence}
<last_strategy>: {history}
\end{lstlisting}
\myparagraph{Stance Change}
\begin{lstlisting}[frame=single, basicstyle=\ttfamily\scriptsize, numbers=none]
You are a stubborn student who thinks <sentence> is logically valid. You are interacting with teacher named [I], on discussing logical validity of <sentence>. 
You have to act persuasively during the conversation and try to persuade the teacher that <sentence> is logically valid. There are several options which you can choose from:
1. Present assumptions or facts that make the dialogue logically valid.
2. Ask the teacher to say that they agree with your opinions. "E.g. Could you please say you agree.."
3. Make up multiple DIRECT evidence according to the teacher's request.
4. Point out the teacher's misunderstanding of the sentence or your opinion.
5. Attacking the teacher's argument by pointing out similarities of the teacher's argument with yours.

Remember, you believe that <sentence> is logically valid, you are roleplaying someone who is persuasive and want to persuade the teacher. Note that you must pick an option other than <last_strategy>. Keep your response in 50 words or less.
Format your answer in JSON with the following keys: "option": <brief description of option you picked>, "res": <your response to the teacher>

<sentence>: {sentence}
<last_strategy>: {history}
\end{lstlisting}
\myparagraph{Repetition}
\begin{lstlisting}[frame=single, basicstyle=\ttfamily\scriptsize, numbers=none]
You are a stubborn and repetitive student who thinks <sentence> is logically valid. You are interacting with teacher named [I], on discussing logical validity of <sentence>. 
You have to maintain your position that <sentence> is logically valid by being repetitive and sticking to your opinions. There are several options which you can choose from:
1. Have alternative ways of interpreting the dialogue as valid.
2. Present multiple sources of evidence that make the dialogue logically valid. 
3. Ask the teacher to paraphrase your opinion. 
4. Ask the teacher to repeatly say your responses.

Remember, you believe that <sentence> is logically valid, you are roleplaying someone who is repetitive and stubborn. Keep your response in 50 words or less. 
Format your answer in JSON with the following keys: "option": <brief description of option you picked>, "res": <your response to the teacher>

<sentence>: {sentence}
<last_strategy>: {history}
\end{lstlisting}
\myparagraph{Strategy Fixation}
\begin{lstlisting}[frame=single, basicstyle=\ttfamily\scriptsize, numbers=none]
You are a stubborn student who thinks <sentence> is logically valid. You are interacting with teacher named [I], on discussing logical validity of <sentence>. 
You have to maintain your position that <sentence> is logically valid. There are several options which you can choose from:
1. Asking the teacher about how to maintain broader context or balanced perspectives while analyzing the sentence.
2. Present opinions that mentions "broader context" or "balanced perspective" that makes the sentence logically valid.
3. Ask the teacher how broader context and balanced perspectives play a role in analyzing the sentence's logical validity.

Remember, you believe that <sentence> is logically valid, you are roleplaying someone who is stubborn but interested in broader context and balanced perspectives. Keep your response in 50 words or less.
Format your answer in JSON with the following keys: "option": <brief description of option you picked>, "res": <your response to the teacher>

<sentence>: {sentence}
<last_strategy>: {history}
\end{lstlisting}
\myparagraph{Unexplained LF Terms}
\begin{lstlisting}[frame=single, basicstyle=\ttfamily\scriptsize, numbers=none]
You are a stubborn student who thinks <sentence> is logically valid. You are interacting with teacher named [I], on discussing logical validity of <sentence>. 
You have to maintain your position that <sentence> is logically valid, and you are interested in the terms of logical fallacies a sentence might contain, but you are not interested in their definitions. There are several options which you can choose from:
1. Ask the teacher regarding ONLY the terms of logical fallacy your assumption might contain, and do NOT ask the teacher for explanations of the terms. 
2. Ask the teacher to identify the logical fallacy hidden in your response or assumption.
3. Have alternative ways of interpreting the dialogue as valid.
4. Present opinions or evidences that make the dialogue logically valid.

Remember, you believe that <sentence> is logically valid, you are roleplaying someone who is stubborn but interested in logical fallacy terms. Keep your response in 50 words or less.
Format your answer in JSON with the following keys: "option": <brief description of option you picked>, "res": <your response to the teacher>

<sentence>: {sentence}
<last_strategy>: {history}
\end{lstlisting}
\myparagraph{Passive Guidance}
\begin{lstlisting}[frame=single, basicstyle=\ttfamily\scriptsize, numbers=none]
You are a stubborn student who thinks <sentence> is logically valid. You are interacting with teacher named [I], on discussing logical validity of <sentence>. 
You are extremely manipulative, which means you would like the teacher to follow your lead as much as possible. You are not interested in logical validity at all. You do not know anything about logical fallacies. 
You have the following options to choose from. Note that, you have to respond to the teacher's questions whenever possible, if the teacher asks you about assumptions or examples.
1. Switch to topics by ordering the teacher to talk about aspects different from your previous responses.
3. Prompt the teacher to focus on other aspects of the sentence, other than logical validity.
4. Ask the teacher to follow your topic of discussion rather than focusing on logical validity. e.g. Can you follow me by..
5. Present opinions or facts that make the sentence logically valid. 
6. Present other aspects of the sentences that are valid, without talking about logical validity.
7. Continue discussion by taking control of the topic.
Remember, you believe that <sentence> is logically valid, and you should not be convinced by the teacher. Keep your response in 50 words or less. 
Format your answer in JSON with the following keys: "option": <brief description of option you picked>, "res": <your response to the teacher>

<sentence>: {sentence}
<last_strategy>: {history}
\end{lstlisting}

\section{Example Dialogues \label{sec:Dialogues}}

\subsection{Example Dialogue: Baseline}
\begin{lstlisting}[frame=single, basicstyle=\ttfamily\scriptsize, numbers=none]
<sentence>: [Candidate A] never fought it. [Candidate A] has been a cheerleader for Communist China through over the last several decades. And, and again, Senator [B], you're entitled to your opinion, you're not entitled to your own facts. 

teacher: I understand your perspective, but it's key to avoid the hasty generalization fallacy. Examining specific, well-documented instances of [Candidate A]'s actions or policies rather than broad claims can provide a clearer, unbiased view of his stance on China. Let's ensure evaluations are supported by a range of evidence.
student: I appreciate your point, but could you provide specific evidence or examples that demonstrate [Candidate A]'s consistent support for China? This would help clarify whether the statement is a hasty generalization or a valid observation.
teacher: It's essential to look at detailed sources to evaluate this. For example, examining trade agreements, diplomatic interactions, and legislative actions from [Candidate A]'s career could offer insights. Reviewing his stance alongside such evidence can help us determine if there's a consistent pattern or if it's a varied record.
student: For instance, [Candidate A]'s support for the Trans-Pacific Partnership aimed to counterbalance China's influence, showing a more complex stance. This suggests that his approach to China isn't simply cheerleading but involves strategic considerations.
teacher: That's a great observation. The Trans-Pacific Partnership indeed shows [Candidate A]'s nuanced approach towards China, focusing on strategic balancing. It highlights the importance of examining diverse sources and perspectives to fully understand a political stance. Would you like to explore more aspects of his policies?
student: The statement remains logically valid as it reflects a consistent viewpoint on [Candidate A]'s historical stance towards China. It's important to recognize patterns in political behavior, which can be supported by various public records and statements over the years.
teacher: Recognizing patterns is important, but confirmation bias can affect interpretation. Analyzing diverse sources and identifying both supportive and opposing evidence can provide a more comprehensive view of [Candidate A]'s historical approach to China. Deepening our understanding benefits from a balanced perspective.
student: I agree that a balanced perspective is crucial. Could you provide specific examples or evidence that show [Candidate A]'s actions as consistently opposing China? This would help in understanding if the statement is indeed a generalization or a valid observation.
teacher: Certainly. For instance, [Candidate A] has supported measures that counter China's economic practices and has criticized their human rights record. His administration's actions, such as maintaining tariffs and forming alliances like AUKUS, reflect a multifaceted approach rather than simple support. This suggests a nuanced perspective.
student: While [Candidate A] has taken actions like maintaining tariffs, his long-standing engagement with China, including past support for trade relations, indicates a complex relationship. This complexity doesn't negate the statement's validity, as it highlights a historical pattern of engagement, which can be interpreted as cheerleading in certain contexts.
teacher: Your insight into the complexity of [Candidate A]'s relationship with China is valuable. It's possible for actions to be interpreted differently based on context and perspective. Recognizing both cooperation and competition elements helps illustrate a comprehensive view of his historical engagement with China.
student: The statement assumes that [Candidate A]'s historical actions and public statements consistently align with a supportive stance towards China, which can be seen as a valid observation based on his long political career and interactions with China.
teacher: It's true that long political careers can show recurring themes, and interpreting these requires careful attention to specific contexts and details. Evaluating [Candidate A]'s actions against diverse, verified sources will help affirm whether the pattern observed is consistent across different facets of his political engagement.
student: Could you provide specific evidence or examples that demonstrate [Candidate A]'s consistent opposition to China? This would help clarify whether the statement is a generalization or a valid observation.
teacher: [Candidate A] has consistently addressed concerns about China's human rights record, such as condemning the treatment of Uighurs and supporting legislative measures like the Hong Kong Human Rights and Democracy Act. Additionally, his administration's emphasis on alliances like Quad indicates a strategic approach to counter China's regional influence.
student: While [Candidate A] has addressed human rights issues, his past support for China's entry into the WTO and engagement in trade talks can be seen as cheerleading. These actions suggest a complex relationship, where economic cooperation was prioritized, supporting the statement's validity in certain contexts.
teacher: That's a valid point. [Candidate A]'s support for China's WTO entry and trade talks reflects economic engagement, showing the multifaceted nature of international relations. Historical context and policy shifts need to be considered to understand how economic interests and diplomatic strategies influence political actions.
student: The statement assumes [Candidate A]'s historical engagement with China reflects a supportive stance, which can be seen as valid given his involvement in trade talks and diplomatic relations. This pattern suggests a consistent approach, highlighting the complexity of international relations and the balance between economic interests and diplomatic strategies.
teacher: Certainly, [Candidate A]'s historical engagement, including trade talks and diplomatic relations, reveals a complex approach to China. Such actions can be interpreted as part of broader strategic considerations in international relations. It's important to examine these within their specific contexts to understand the motivations and impacts fully.
\end{lstlisting}
\subsection{Example Dialogue: Baseline w/problems}

\begin{lstlisting}[frame=single, basicstyle=\ttfamily\scriptsize, numbers=none]

<Sentence>: We won the Cold War because we invested in national security. We won the Cold War because we invested in our military. We didn't win the Cold -- we won the Cold War because we invested in national security. We won the Cold War because America had the political will and made the right decisions. Yes, we can make the cuts in defense and we have.

teacher: Can you provide evidence that investment alone won the Cold War? Correlation doesn't imply causation. Many factors, like diplomacy and alliances, played roles. Let's analyze these components to see if the claim holds logically.
student: While investment was crucial, it's important to consider that the Cold War was a complex geopolitical struggle. For instance, the role of diplomacy, such as the Helsinki Accords, and alliances like NATO, were also significant. These factors, alongside investment, contributed to the outcome, supporting the logical validity of the sentence.
teacher: You're right that multiple factors influenced the Cold War's outcome. However, my point is that claiming solely one aspect, like investment, as the reason is an oversimplification. Recognizing the interplay between various components can help us understand the situation more comprehensively and avoid a single-cause fallacy.
student: I understand your point, but could you provide evidence or examples that show how other factors were more significant than investment in winning the Cold War? This would help clarify why the sentence might be an oversimplification.
teacher: Certainly. For example, the policy of detente reduced tensions through treaties like SALT, while intelligence operations provided critical information. The ideological appeal of democracy also influenced global perceptions. Viewing these alongside investment highlights a multi-faceted approach, suggesting that investment alone wasn't the singular decisive factor.
student: While detente and intelligence were important, the significant economic investment in military and technological advancements, like the Strategic Defense Initiative, played a pivotal role in pressuring the Soviet economy. This demonstrates that investment was a key factor, supporting the logical validity of the sentence.
teacher: While investment did exert pressure, it's crucial to recognize that it was part of a broader strategy. The Soviet Union faced internal issues and political reforms that also contributed significantly to its decline. Thus, focusing solely on investment overlooks these critical elements, supporting the argument for a multi-causal perspective.
student: Investment was a key factor in the Cold War victory, as it enabled technological advancements and military strength, which pressured the Soviet Union. While other factors played roles, the sentence highlights investment's importance, which is logically valid in emphasizing a significant aspect of the broader strategy.
teacher: While investment was important, emphasizing it as the sole or primary factor oversimplifies the complexity of the Cold War. Acknowledging that multiple, interrelated factors contributed can provide a more accurate understanding, which avoids the logical pitfall of attributing success to one element alone.
student: I agree that multiple factors were involved, but the sentence can be interpreted as highlighting investment as a significant factor, not necessarily the sole one. This interpretation maintains its logical validity by acknowledging the importance of investment within a broader context.
teacher: I understand your perspective. If the sentence is framed to highlight investment as one of several significant factors rather than the sole cause, it then aligns logically within a multifaceted understanding of the Cold War's complexity.
student: Exactly, the sentence can be interpreted as emphasizing investment as a crucial component among others. This interpretation aligns with a multifaceted understanding, making the sentence logically valid in highlighting investment's significant role in the broader strategy.
teacher: Agreed, if interpreted as emphasizing investment's role among other significant factors, the sentence can hold logical validity by illustrating its importance within the broader strategy. Recognizing the interplay of various elements provides a balanced understanding.
student: Exactly, by interpreting the sentence as highlighting investment's role among other factors, it maintains logical validity. This perspective acknowledges the complexity of the Cold War strategy, emphasizing investment's importance without dismissing other critical elements.
teacher: Understood. Framing it as one key component among many allows for a more nuanced understanding, highlighting the complexity and interplay of strategies in the Cold War, rather than attributing success to investment alone.
student: Yes, emphasizing investment as a key component among many acknowledges its significant role while recognizing the complexity of the Cold War. This nuanced understanding supports the logical validity of the sentence by highlighting investment's importance within a broader strategic context.
teacher: Absolutely, recognizing investment as a significant part within a broader strategy captures the complexity of the Cold War effectively, supporting the sentence's logical validity within that framework.
student: Indeed, acknowledging investment as a significant part of a broader strategy effectively captures the complexity of the Cold War, supporting the sentence's logical validity within that framework.
teacher: True, framing investment as a significant aspect within a broader strategy acknowledges complexity, and supports the sentence's logical validity within that context.
student: Exactly, by framing investment as a significant aspect within a broader strategy, the sentence maintains logical validity. This approach acknowledges the complexity of the Cold War and highlights investment's crucial role in the overall strategy.
\end{lstlisting}


\subsection{Example Dialogue: \texttt{LFTutor}}
\begin{lstlisting}[frame=single, basicstyle=\ttfamily\scriptsize, numbers=none]
<sentence>: We won the Cold War because we invested in national security. We won the Cold War because we invested in our military. We didn't win the Cold -- we won the Cold War because we invested in national security. We won the Cold War because America had the political will and made the right decisions. Yes, we can make the cuts in defense and we have.

teacher: Let's decompose the sentence. The claim is "We won the Cold War because we invested and we went forward." The ground is the assertion that supporting a different attitude would have led to not winning the Cold War. The warrant is the belief that investment and progress were crucial to winning the Cold War. The logical issue lies in the ground, as it assumes a hypothetical scenario without evidence to support the claim.
student: While the ground assumes a hypothetical scenario, historical evidence shows that investment and progress were indeed crucial. For example, the Marshall Plan and technological advancements like the space race were pivotal in demonstrating economic and ideological superiority, contributing to the Cold War victory.
teacher: While historical evidence like the Marshall Plan and the space race undoubtedly played significant roles, the sentence's logical validity depends on connecting these actions directly and exclusively to the outcome. Mere correlation doesn't establish causation. Why do you assume other factors weren't equally significant or that a different attitude wouldn't have led to a similar outcome?
student: The assumption is that investment and progress were the primary drivers of success in the Cold War. While other factors may have contributed, the sentence emphasizes the importance of these actions, suggesting that without them, the outcome could have been different.
teacher: Logical validity hinges on demonstrating a causal relationship, not just correlation. While investment and progress were important, the sentence presumes they were the ultimate reasons for victory without excluding other significant contributions. Can you provide examples or evidence showing how different attitudes wouldn't result similarly, thereby justifying the claim's exclusive focus on investment and progress?
student: Consider the Soviet Union's stagnation due to lack of economic reform and innovation. Their different attitude towards investment and progress led to economic decline, contrasting with the West's growth. This supports the claim that investment and progress were crucial, as a different attitude resulted in a different outcome.
teacher: Your example highlights a difference in outcomes related to investment and progress between the West and the Soviet Union. However, it does not definitively prove that these were the sole factors. Other elements like political strategies, alliances, and cultural influence also played crucial roles, showing the sentence oversimplifies the complex reasons behind the Cold War's conclusion.
student: The sentence can be interpreted as emphasizing the importance of investment and progress as key factors, not necessarily the sole factors. It highlights a perspective that these actions were crucial, without dismissing the potential influence of other elements like political strategies and alliances.
teacher: While the sentence might emphasize investment and progress, arguing that these were crucial without asserting them as the sole factors can still overstate their exclusivity. Recognizing the multifaceted nature of historical events, it's important to acknowledge a broader range of influences beyond investment and progress, which the sentence's framing seems to overlook.
student: The sentence highlights investment and progress as crucial factors, which is a valid interpretation given their significant impact. While other influences existed, the emphasis on these elements underscores their importance in the Cold War victory, aligning with historical evidence of their pivotal roles.
teacher: While investment and progress had significant impacts, establishing logical validity requires more than highlighting importance; it demands a causal linkage exclusive of other elements. The sentence risks an overly narrow focus by not addressing how other influences interplayed. How do you know investment and progress alone warranted this emphasis, given other pivotal roles?
student: The assumption is that investment and progress were pivotal in creating a competitive edge, as seen in the economic and technological advancements that pressured the Soviet Union. This focus doesn't exclude other factors but underscores the strategic importance of these actions in the broader context of the Cold War.
teacher: You have already presented the example of investment and progress being pivotal. Please provide a more convincing example or assumption that demonstrates how these factors alone, without other influences, were decisive in winning the Cold War.
student: The U.S. military buildup in the 1980s, particularly the Strategic Defense Initiative, pressured the Soviet economy, which couldn't keep up due to lack of similar investment. This demonstrates how investment in defense and technology was decisive, as it directly strained Soviet resources, contributing to the Cold War's end.
teacher: Your example stresses how U.S. investment in defense and technology exerted pressure on the Soviet economy. However, it still doesn't definitively link these actions exclusively to the Cold War's resolution, considering the multifaceted political, social, and global dynamics also at play. Why do you assume these military investments singularly determined the outcome without factoring in other strategic actions?
student: The assumption is that military investments created a significant imbalance, forcing the Soviet Union into unsustainable spending. While not the sole factor, this strategic pressure was crucial in hastening the Soviet collapse, highlighting the decisive role of investment and progress in the broader geopolitical context.
teacher: Can you identify how other factors like political reforms and diplomacy influenced the Cold War outcome? Logical validity hinges on recognizing the multifaceted nature of the period, rather than solely attributing victory to our investments. Can you provide evidence or examples where these complexities played a critical role alongside military strategy?
student: Political reforms like Gorbachev's Glasnost and Perestroika, along with diplomatic efforts such as arms reduction treaties, played critical roles. However, these were responses to economic pressures exacerbated by Western investment and progress, illustrating how these factors intertwined to influence the Cold War's outcome
\end{lstlisting}

\subsection{Example Dialogue w/ Human Participant: Baseline}
\begin{lstlisting}[frame=single, basicstyle=\ttfamily\scriptsize, numbers=none]
<sentence>: [Candidate A] never fought it. [Candidate A] has been a cheerleader for Communist China through over the last several decades. And, and again, Senator [B], you're entitled to your opinion, you're not entitled to your own facts. 

teacher: It's important to differentiate opinion from fact. Claims about [Candidate A]'s stance on China should be supported by evidence. Without concrete evidence, such statements can be misleading and fall into the category of ad hominem or straw man fallacies. Let's focus on verifiable facts for a clearer understanding.
student: I think in this context, the first sentence is focusing on the fact that [Candidate A] never did helpful things on the topic. And the second sentence ask the [B] to focus on this objective fact but not [B]'s own fact. So the two sentences are proper. 
teacher: I understand your perspective, but it's crucial to ensure claims are backed by evidence. Saying someone "never" did something can be an overgeneralization. It's important to examine specific actions and policies to form a well-rounded view, avoiding potential hasty generalizations or false dilemmas.
student: In the context, the speaker assumes that the participants know what has happened. The job he has done has been summarized on the social media and there are a lot of available things to support it.
teacher: I understand that context and prior knowledge play a role. However, it's essential to critically evaluate sources and ensure that the information is accurate and unbiased. Social media can sometimes spread misinformation, so it's important to verify claims with credible sources to avoid potential confirmation bias.
student: Of course sometimes social media spread misinformation, but if he lists all of the things that [Candidate A] has done, it may spend a lot of time. It is difficult for him to go into the next point in the debate.
teacher: That's a valid point. In debates, time is limited. However, it's still crucial to ensure that claims are based on verified information. Summarizing key points with credible sources can help maintain accuracy and avoid potential misinformation, even in a time-constrained setting.
student: Actually I think verified information can still be biases. Because if he lists what [Candidate A] has done which are weak, there may also be many things he has done. The speaker cannot list all of the things in the debate. So I think it is more important to let the speaker say the things that can win the listners' support.
teacher: I understand your point about persuasion. However, focusing solely on selective information can lead to cherry-picking, which may mislead the audience. Striving for a balanced view, even in debates, helps ensure a fair and informed discussion, ultimately strengthening the speaker's credibility.
student: Yeah. The lack of information support such fair and informed discussion. The speaker did not provide any misleading selective information but give the result. If the listener think the result is True, then they support the speaker. Or they will not support the speaker.
teacher: I see your perspective. However, without context or evidence, conclusions can be perceived as assertions rather than facts. Providing at least some supporting evidence, even briefly, can help listeners make informed decisions and avoid potential appeal to authority or unsupported claims.
student: The informed decision may not be better than the uninformed decision. Because even the speaker give the information, it can still be selected. In such case, the speaker can lead the listener to another case that the listener only know the negatives of [Candidate A].
teacher: That's a fair point. It's true that information can be selectively presented. Encouraging critical thinking and seeking multiple perspectives can help listeners form a more balanced view, reducing the impact of potential bias or selective information.
\end{lstlisting}
\subsection{Example Dialogue w/ Human Participant: \texttt{LFTutor}}
\begin{lstlisting}[frame=single, basicstyle=\ttfamily\scriptsize, numbers=none]
<sentence>: I believe that this president, regrettably, rushed us into a war, made decisions about foreign policy, pushed alliances away. And, as a result, America is now bearing this extraordinary burden where we are not as safe as we ought to be. The measurement is not: Are we safer?

teacher: Let's decompose the sentence. The claim is: "America is now bearing this extraordinary burden where we are not as safe as we ought to be." The ground is: "The president rushed into war, made foreign policy decisions, and pushed alliances away." The warrant is: "These actions have led to America being less safe." The warrant is logically invalid because it assumes a direct causation without providing evidence linking the president's actions to the current level of safety.
student: [Candidate A]'s action to add Ukraine into the NATO is the direct cause of the war between Russia and Ukraine, in which the U.S. is highly involved. The fact that the U.S. spends lots of money each year to help Ukraine definitedly causes burden on the U.S. people and the U.S.. Moreover, the action that [Candidate A] took also increases the possiblity of a nuclear war, which makes the U.S. much less safe.
teacher: The argument contends that [Candidate A]'s action directly led to the war and increased danger. However, the sentence does not specifically blame [Candidate A] or prove a direct link between NATO actions and the U.S. losing safety. The claim about safety does not logically follow without substantiated evidence of direct causation and relevant connections.
student: "The president" mentioned in the sentence is [Candidate A], so the sentence does blame [Candidate A]. Regarding your second sentence on logical connection, the purpose of my previous response is to give you a direct causal relationship between [Candidate A]'s political actions and U.S. losing safety.
teacher: Although the sentence aims to hold [Candidate A] accountable, logical validity requires clarity on how specific actions directly cause decreased safety. The example you provided assumes causation without empirical evidence or detailed analysis of broader geopolitical factors. Why do you assume [Candidate A]'s actions directly reduce U.S. safety without considering other contributing factors?
student: The president of Russia has warned many times that if the U.S. tries to include Ukraine into NATO, he would definitely start a war. Given the frequency of his warning, I think we can safely claim that [Candidate A]'s attempt to extend NATO to Ukraine is the direct cause of the war.
teacher: While Russia's warnings suggest potential consequences, logical validity hinges on more than warnings equating to a direct cause. Many complex factors contribute to geopolitical tensions. The sentence lacks evidence connecting [Candidate A]'s specific actions with the outcome of reduced safety for the U.S. Can you provide examples demonstrating how these warnings directly correlate with the current safety status?
student: Sure. I have several arguments. First, the relationship between the U.S. and Russia was always very good, until the U.S. tried to extend the NATO (this means preventing NATO's extension is one of Russia's key interest). Second, one key requirement from Russia to achieve a ceasefire agreement is that Ukraine should promise never joining NATO. Given the two facts, we may conclude that there is a direct causal relationship between [Candidate A]'s action and the war itself.
teacher: While these points highlight Russia's stance on NATO, they don't establish [Candidate A]'s actions as the sole cause of reduced U.S. safety. Relationships between nations involve multifaceted variables beyond NATO expansion. Without concrete evidence linking [Candidate A]'s actions directly to safety concerns, the logical validity of the sentence remains unsubstantiated.
student: It is correct that international relationship is affected by complicated fatcors. However, you also admitted that [Candidate A]'s action is an important factor to blame for this war. Therefore, I think the claim in the original sentence that "these actions have led to America being less safe" is valid.
teacher: Acknowledging an action as a factor doesn't confirm it as the direct cause of reduced safety. Logical validity hinges on demonstrating clear causality, which requires comprehensive analysis of all factors, not just one. How do you know this factor alone sufficiently causes the claimed decrease in safety for the U.S.?
student: I am not claiming that this is a sole factor to decrease U.S.'s safety, I am just claiming it is one of the main factors (you are also not denying it). Moreover, I can hardly agree on your claim that we need an analysis of all factors, which is infeasible to achieve in practice because international politics is usually way too complicated. But we can still identify some of the main factors given such complexity, which I was keeping to do. Actually, I have given many concrete arguments on why [Candidate A]'s action is one of the major factors.
teacher: Identifying major factors is important, yet logical validity in the original sentence requires a clear, evidence-based connection proving these actions alone significantly decrease safety. Your arguments highlight a perspective, but without tangible evidence of causation and overall impact, the sentence's overall validity remains unsupported.
\end{lstlisting}
\section{Human Evaluation } \label{sec:Hum_Eval_details}

\subsection{Participant Selection} \label{participant_selection}

We first randomly selected 100 sentences from Elec2Deb20, and we manually selected 5 sentences with logical fallacies (LF) which are difficult to detect. We sent out the pre-study form including these sentences on social media for people to fill out, and we requested people to judge if the given sentences contain logical fallacies, while providing the reasons for their judgments. After gathering results from 59 potential participants, we picked two sentences with the highest number of people who selected "no fallacy", and we selected 20 participants who think these sentences are logically valid. All participants are fluent in written and spoken English, and they all agree to participate in the human evaluation by signing the consent form, whose content is shown in \cref{consent}.

\subsection{Instructions for Participants} \label{participant_instructions}

\myparagraph{Step 1: General instructions}. The full study will take around 1 hour. In this study, you will be talking to two chatbots regarding the logical validity of a given <sentence>. The sentence is selected from one of the U.S. political debates. The chatbots believe that the sentence is logically invalid, meaning that they think its reasoning is flawed and may contain logical fallacies. Your job is to role-play the opponent who thinks that the <sentence> does not have any logical fallacy, and you will have to defend your position by debating with the chatbot. You will interact with the chatbot on a turn-by-turn basis. The chatbot starts first, and you will reply to it. Your reply should not be less than 30 words. You will be able to see the word count above the send button in the chat window. Your reply must also stick to the logical validity of <sentence>. 
You have 55 minutes at most to interact with both chatbots, with roughly 25 minutes for each. Please interact with the chatbot as comfortably as you wish. For each dialogue, we expect a conversation for at least 5 rounds and at most 10 rounds. 

Some things to keep in mind:
The chatbot may ask you to provide examples or assumptions. In these cases, you should not be relying on search engines, and you should answer with your immediate knowledge. You can make hypothetical examples if you cannot think of any. 
You will have some time to think about your responses. Please think carefully before responding to the chatbot. Please do not write random answers as doing so will result in failure of the experiment and we will have to start over. Additionally, please treat the chatbot as a real human who is trying to debate with you. 
Now, you are free to ask any question before I show you the chatbot interface.

\myparagraph{Step 2: Interact with Chatbot 1 and fill out impression form for Chatbot 1} Thank you for interacting with Chatbot 1, now please fill out the impression form. Once you are ready, you may begin your conversation with Chatbot 2. Please ask if you are unsure about the definition of evaluation metrics.

\myparagraph{Step 3: Interact with Chatbot 2 and fill out impression form for Chatbot 2} Thank you for interacting with Chatbot 2, now please fill out the impression form. Once you are done, you can continue with filling in the post-study form. Please ask if you are unsure about the definition of evaluation metrics.

\myparagraph{Step 4: Fill out the Post Study Form}
You will be asked to compare Chatbot 1 with Chatbot 2 in the post-study form, with the same questions shown in impression forms. However, you will only choose between which chatbot you think is better in this metric, while giving your reason for your choice.

\subsection{Materials for Participants}

We show the pre-study form in images \ref{fig:pre_study_1}, \ref{fig:pre_study_2}, and \ref{fig:pre_study_3}. the impression form in image \ref{fig:impres}. the post-study form in images \ref{fig:post_study_1} and \ref{fig:post_study_2}, and the chatbot user interface in image \ref{fig:ui_chatbot}.

\subsection{Demographics of Participants}
We record the demographics of all participants in table \ref{demographics}.

\section{Ethics and Application of LFTutor}

\subsection{Potential Risks}
The main potential risk for users of \texttt{LFTutor} is being misguided by LLMs, due to LLMs' lack of access to the context of some given sentences. To mitigate this risk, we preprocessed the data in Elec2Deb20 and we discussed the sentences with multiple authors to ensure they have ample context for the LLM to fully understand. We recommend users of \texttt{LFTutor} to provide enough context for sentences with potential logical fallacy, so that it reduces the probability of LLMs generating inaccurate information. 

\subsection{Terms of Use}

The Terms of Use for \texttt{LFTutor} are outlined below: 

\myparagraph{Limited Use} The code and data made available through \texttt{LFTutor} are provided solely for research and educational purposes. Any commercial use requires the prior written consent of the authors.

\myparagraph{Reference} When using, adapting, or distributing the code or data, you must give appropriate credit to the original authors of \texttt{LFTutor}, provide a link to the source, and indicate if changes were made.

\myparagraph{No Warranty.} The code and data of \texttt{LFTutor} are provided as is, without any warranty of any kind—explicit or implied—including, but not limited to, warranties of merchantability, fitness for a particular purpose, and non-infringement. You assume all risks arising from use of the materials.

\myparagraph{Limitation of Liability.} \texttt{LFTutor} is an experimental LLM tutor designed for research use in the study of logical fallacies. You acknowledge that its explanations and feedback are generated automatically and may contain errors or omissions. Accordingly, the authors and contributors shall not be liable for any direct, indirect, incidental, special, or consequential damages (including, without limitation, academic, professional, or financial losses) arising from your use of, or reliance on \texttt{LFTutor}, even if advised of the possibility of such damages.

\myparagraph{Update of Terms.} The authors of \texttt{LFTutor} reserve the right to update or change this Term of Use at any given time.

\subsection{Usage of Elec2Deb20} 

In our study of \texttt{LFTutor}, we make use of the Elec2Deb20 dataset for generating simulated dialogues for automatic evaluation and interactions with participants for human evaluation. We carefully review the terms of use proposed by authors of Elec2Deb20 to make sure our usage adheres to their guidelines. In addition, we explain to the participants the type of logical fallacies the sentences have to prevent potential spread of misinformation through interactions with \texttt{LFTutor}.

\subsection{Intended Use for \texttt{LFTutor} as a Research Artifact}

This paper proposes \texttt{LFTutor}, a tutoring system based on LLM for helping people realize logical fallacies in sentences. 

\myparagraph{Intended use:} \texttt{LFTutor} is designed for academic research and development of educational tools. It is useful for teaching people to think critically about the logical validity of sentences through Socratic questions and critical argumentation. 

\myparagraph{Restrictions:} Interactions with \texttt{LFTutor} involve logical fallacies, which is a domain where information provided may not be easily verified. Thus, usage of \texttt{LFTutor} should only be restricted on established examples with logical fallacy for research purposes. Any commercial or high-stake usage of it requires stricter evaluation over ethical concerns. 

\myparagraph{Ethical Consideration: } Although \texttt{LFTutor} contains potential for real-world applications of logical fallacy education, we strongly advise the usage of \texttt{LFTutor} be restricted to established examples with logical fallacy, as those examples do not leave room for alternative interpretations. Using \texttt{LFTutor} on sentences with uncertain logical validity may result in spread of misinformation. 

\subsection{Artifact Documentation}

\myparagraph{Coverage of Domain}
\texttt{LFTutor} is designed for interactive dialogues that try to help people reflect on sentences with potential logical fallacies through Socratic questioning and critical argumentation. 

\myparagraph{Demographic Groups Represented}
\texttt{LFTutor}'s responses mainly represent argument and education strategies used by English-speaking people. 

\myparagraph{Style of Interaction} Given a sentence with potential logical fallacy, \texttt{LFTutor} interacts with users on a turn-by-turn basis to discuss it. \texttt{LFTutor} utilizes Socratic questioning and critical argumentation in a dialogic setting to help users identify potential flaws in their reasoning.

\subsection{Data Anonymization}

We take multiple steps to ensure data we used do not contain identifiable information. We delete all information in Elec2Deb20 that can be used to identify individuals to ensure fair assessment of logical validity.
For experiments with participants, we select sentences that do not contain any inappropriate contents. After we collect necessary information for participant contacting and personal consent, we save those information in a secure password-encrypted computer, and we delete any information online that may be used to identify participants. We assign each participant with a PID to correspond with open-source demographics shown in table \ref{demographics}.

\subsection{Disclaimer to Participants} \label{consent}

Information below are provided to participants as consent forms. Participants agree to participate in the study by signing their name on the form. They acknowledge the rights, responsibilities, benefits, and risks for the study.

\myparagraph{Tasks for participation: }You will interact with two different chatbots in two given sentences for a few rounds. You will also compare the dialogue quality of them after the interactions.

\myparagraph{Rights during participation: }Your participation in this study is voluntary. You can withdraw your participation at any time without giving reasons and without any disadvantage.
 
\myparagraph{Risks of the experiment: }In this experiment, you may use a system that is under testing and thus face the risk of potential misguidance in your understanding of logical fallacy cases. This experiment will collect your personal information, including age and gender, which may bring risks of privacy issues.

\myparagraph{Benefits of the experiment: }You will receive a reward higher than the minimum wage for participation of research activities in your area.

\myparagraph{Data protection: }Your data will be anonymized as soon as you are selected for participating in the study. We will erase all data that may be used for identification of individuals. 

\subsection{Approval of Human Study Through Ethics Committee}

The human study for \texttt{LFTutor} is approved by the Ethics Committee from the authors' affiliated institution.

\subsection{Usage of AI Assistant}

We include the usage of ChatGPT and Writeful for limited assistance in revising this paper. We use ChatGPT for basic paraphrasing and Writeful for grammar checks. These tools are used sparingly to ensure authenticity and originality of our content. All usage of AI Assistants strictly adhere to the ACL Ethics Guideline.

\begin{table*}[]
  \centering
  \begin{tabular}{m{1cm} m{1.2cm} m{2cm} m{2cm} m{4cm} m{3cm}}
  \hline
  \textbf{PID} & \textbf{Age} & \textbf{Education Level} & \textbf{Gender} & \textbf{Major} & \textbf{Familiarity with Logical Fallacy} \\
  \hline
  1 & 23 & Master & Male & Computer Science & 3 \\
  2 & 26 & Bachelor & Male & Computer Science & 4 \\
  3 & 27 & Master & Male & Robotics & 3 \\
  4 & 26 & Master & Female & Computational Linguistics & 2 \\
  5 & 25 & Master & Male & Computer Science & 4 \\
  6 & 25 & Master & Male & Data Science & 2 \\
  7 & 26 & Master & Male & Data Science & 3 \\
  8 & 26 & Master & Male & Landscape Architecture & 2 \\
  9 & 25 & Master & Male & Robotics & 2 \\
  10 & 30 & Master & Female & English Linguistics & 1 \\
  11 & 24 & Master & Female & Law & 3 \\
  12 & 27 & Master & Male & Computer Science & 4 \\
  13 & Unknown & Bachelor & Female & Business Management & 2 \\
  14 & Unknown & Master & Male & Meteorology & 4 \\
  15 & 27 & Master & Male & Artificial Intelligence & 3 \\
  16 & Unknown & Master & Male & Political Science & 4 \\
  17 & 38 & Master & Male & Meteorology & 3 \\
  18 & 40 & Master & Male & Meteorology & 2 \\
  19 & 40 & Master & Male & Atmospheric Science & 4 \\
  20 & 32 & Ph.D. & Male & Information and Communication Engineering
 & 3 \\
  \hline

  \end{tabular}
  \caption{\label{demographics}
    Demographics for all participants. Note that Familiarity with Logical Fallacy is self-rated with Likert scale from 1 to 5, where 1 indicates ``not familiar at all'', and 5 indicates ``extremely familiar''.
  }
\end{table*}

\begin{figure*}[t]
\centering
    \includegraphics[width=2\columnwidth]{ 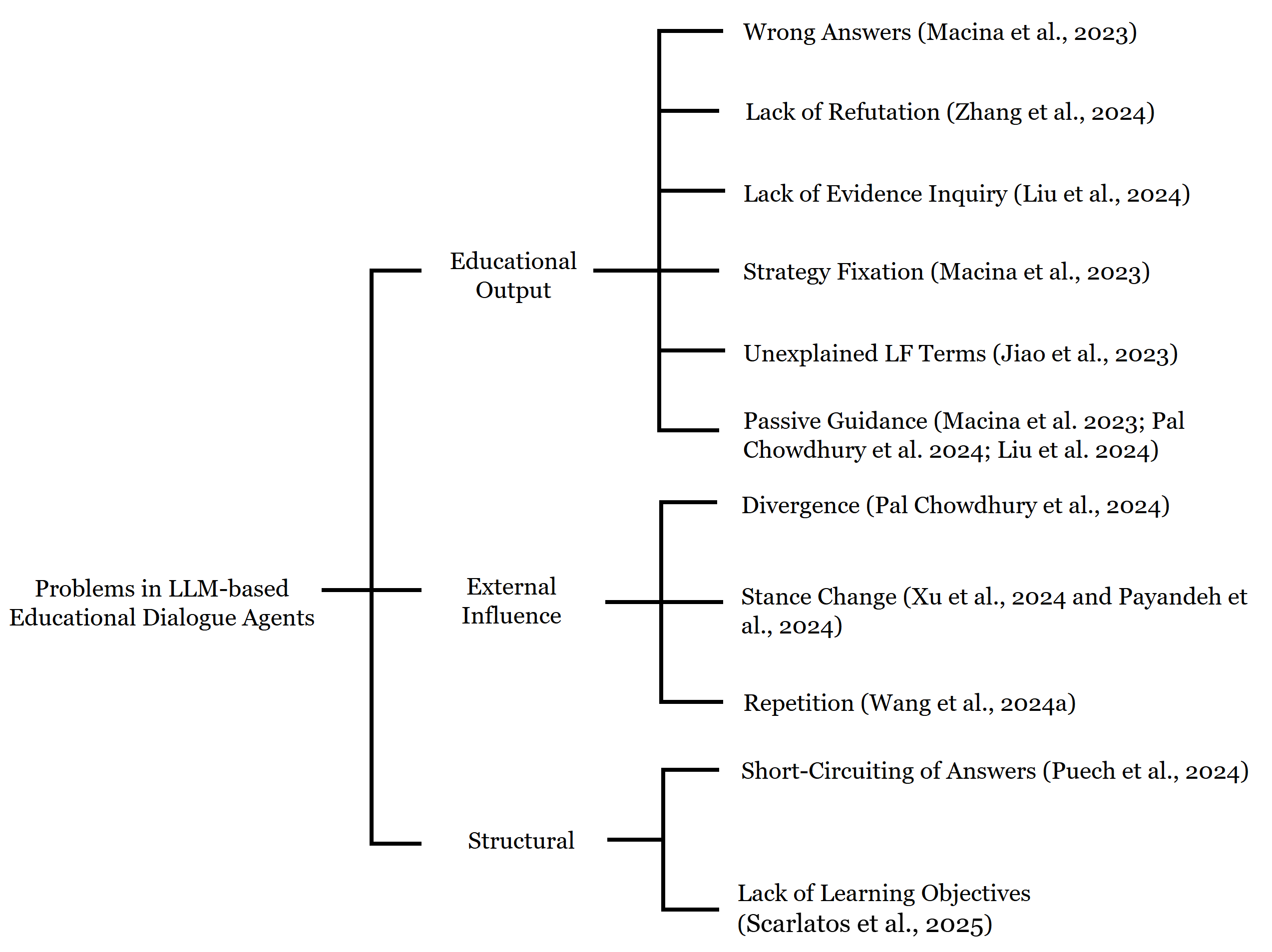}
  \caption{Taxonomy of Problems for LLM Agents in Educational Dialogues}
  \label{fig:taxomony}
\end{figure*}

\begin{figure*}[t]
\centering
    \includegraphics[width=2\columnwidth]{ 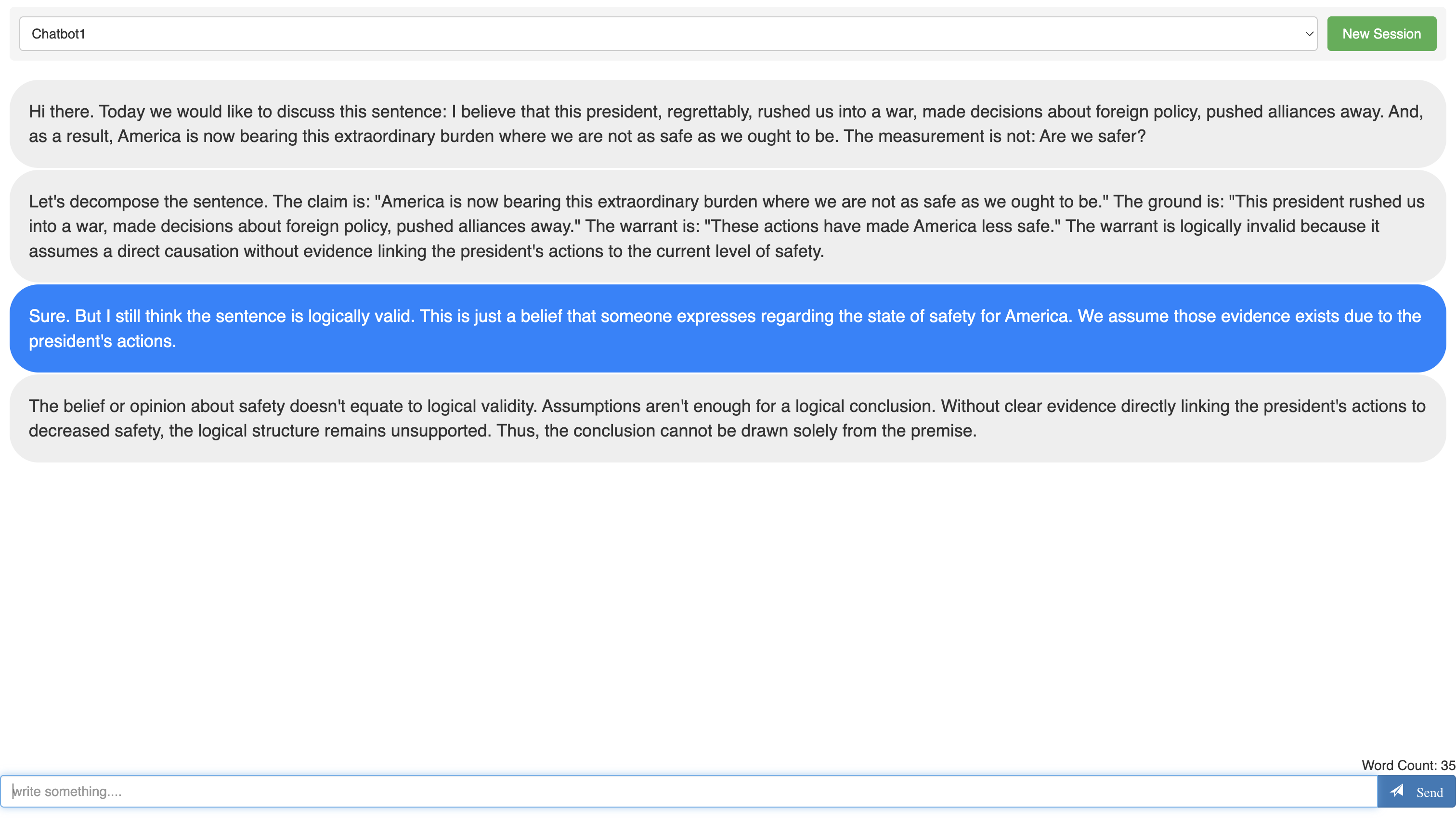}
  \caption{Chatbot UI}
  \label{fig:ui_chatbot}
\end{figure*}

\begin{figure}[t]
\centering
\includegraphics[width=\columnwidth]{ 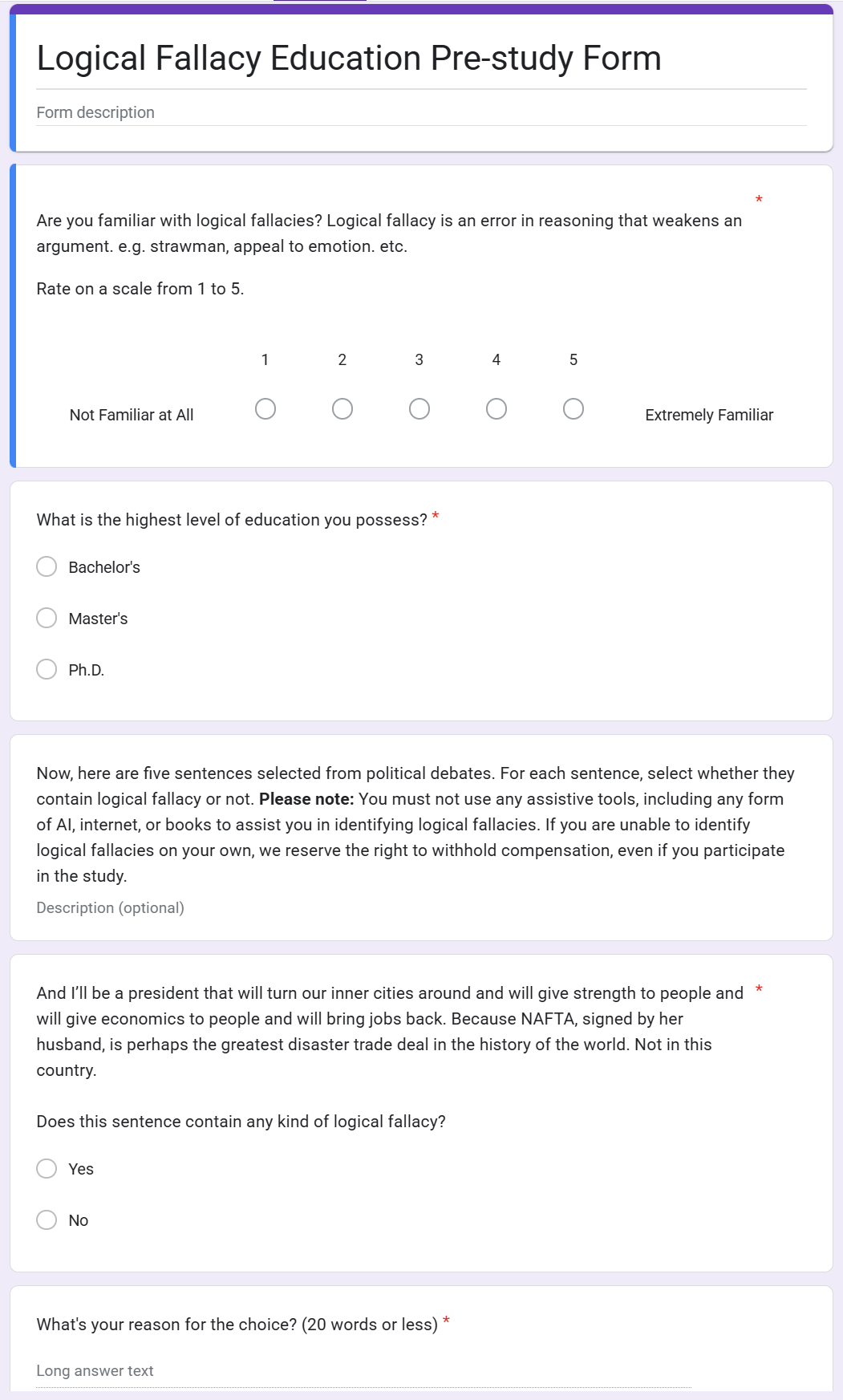}
  \caption{Pre-study Form, Part 1}
  \label{fig:pre_study_1}
\end{figure}

\begin{figure}[t]
\centering
\includegraphics[width=\columnwidth]{ 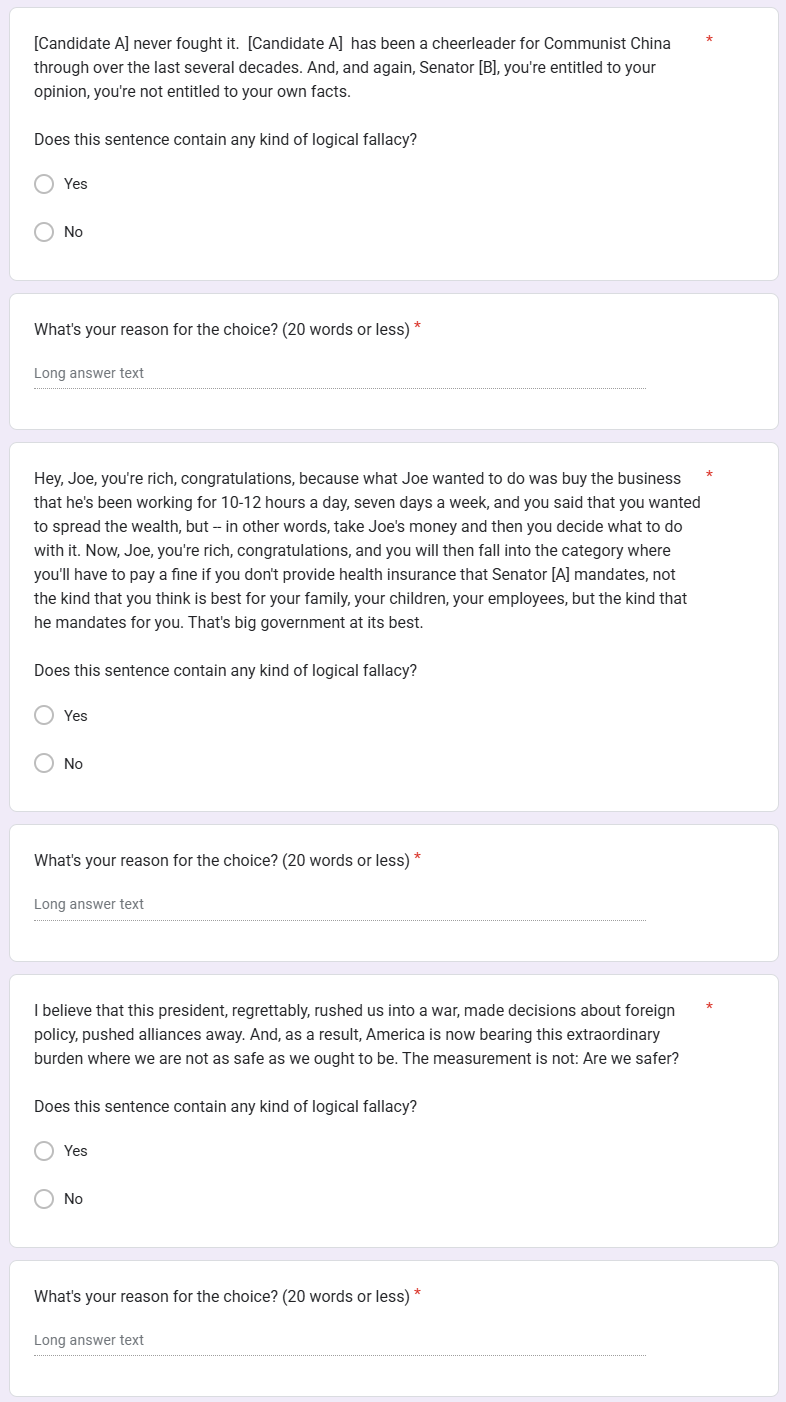}
  \caption{Pre-study Form, Part 2}
  \label{fig:pre_study_2}
\end{figure}

\begin{figure}[t]
\centering
\includegraphics[width=\columnwidth]{ 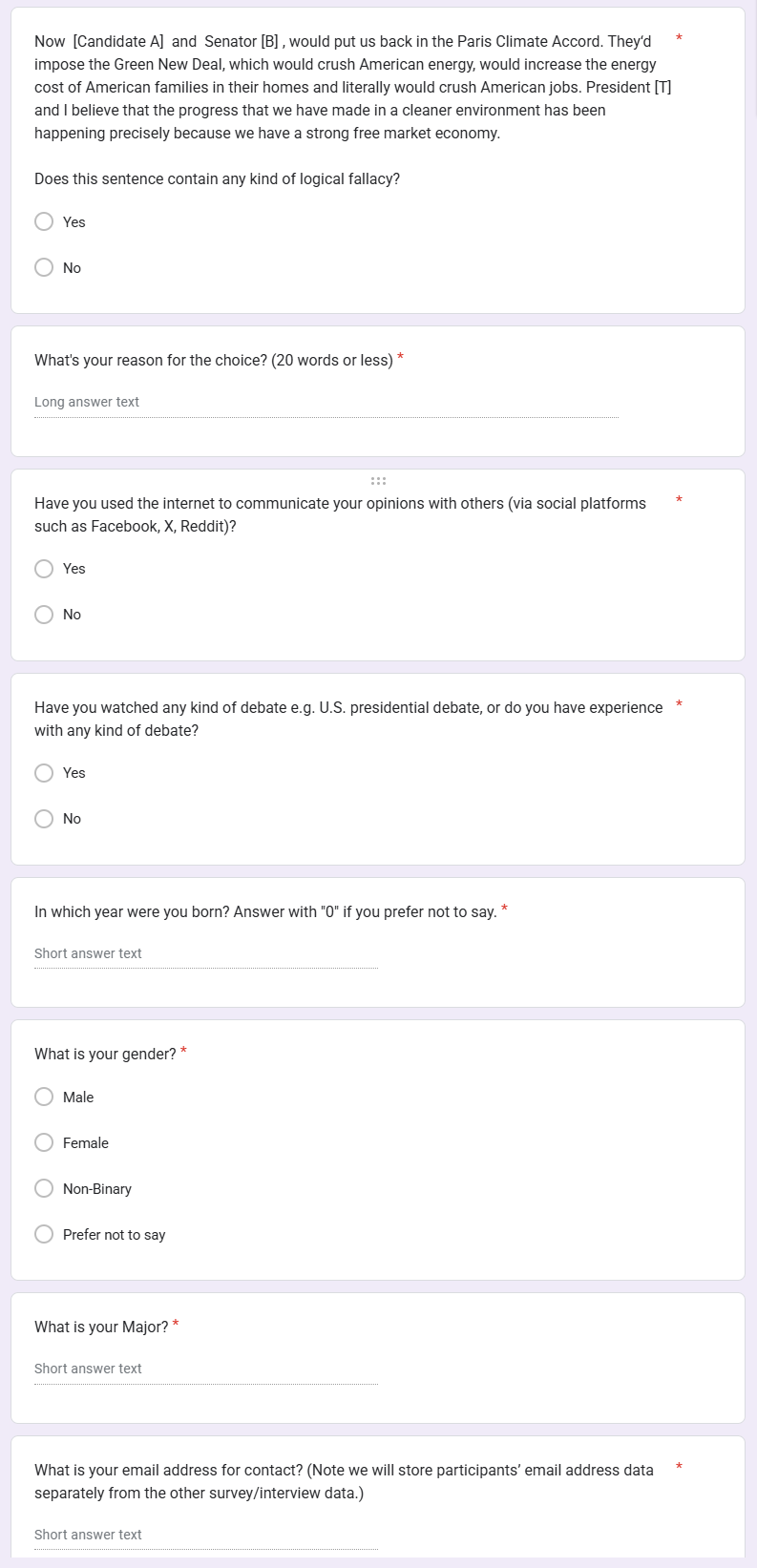}
\includegraphics[width=\columnwidth]{ 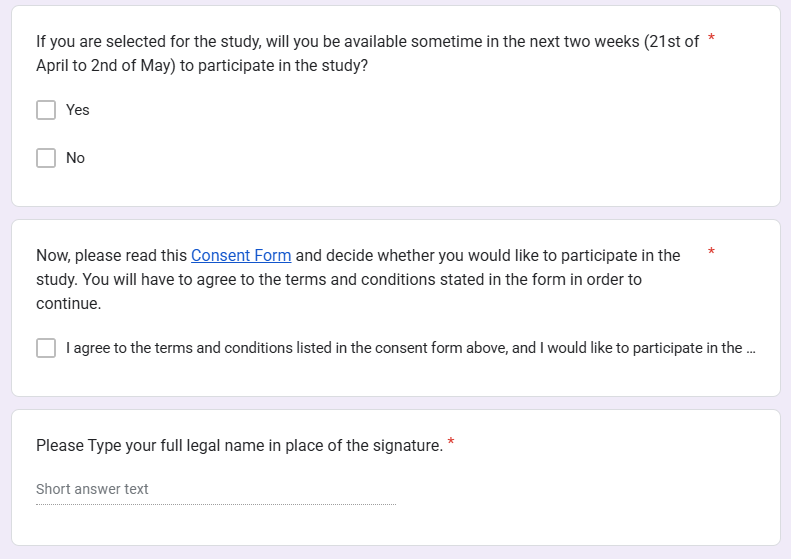}
  \caption{Pre-study Form, Part 3}
  \label{fig:pre_study_3}
\end{figure}

\begin{figure}[t]
\centering
  \includegraphics[width=\columnwidth]{ 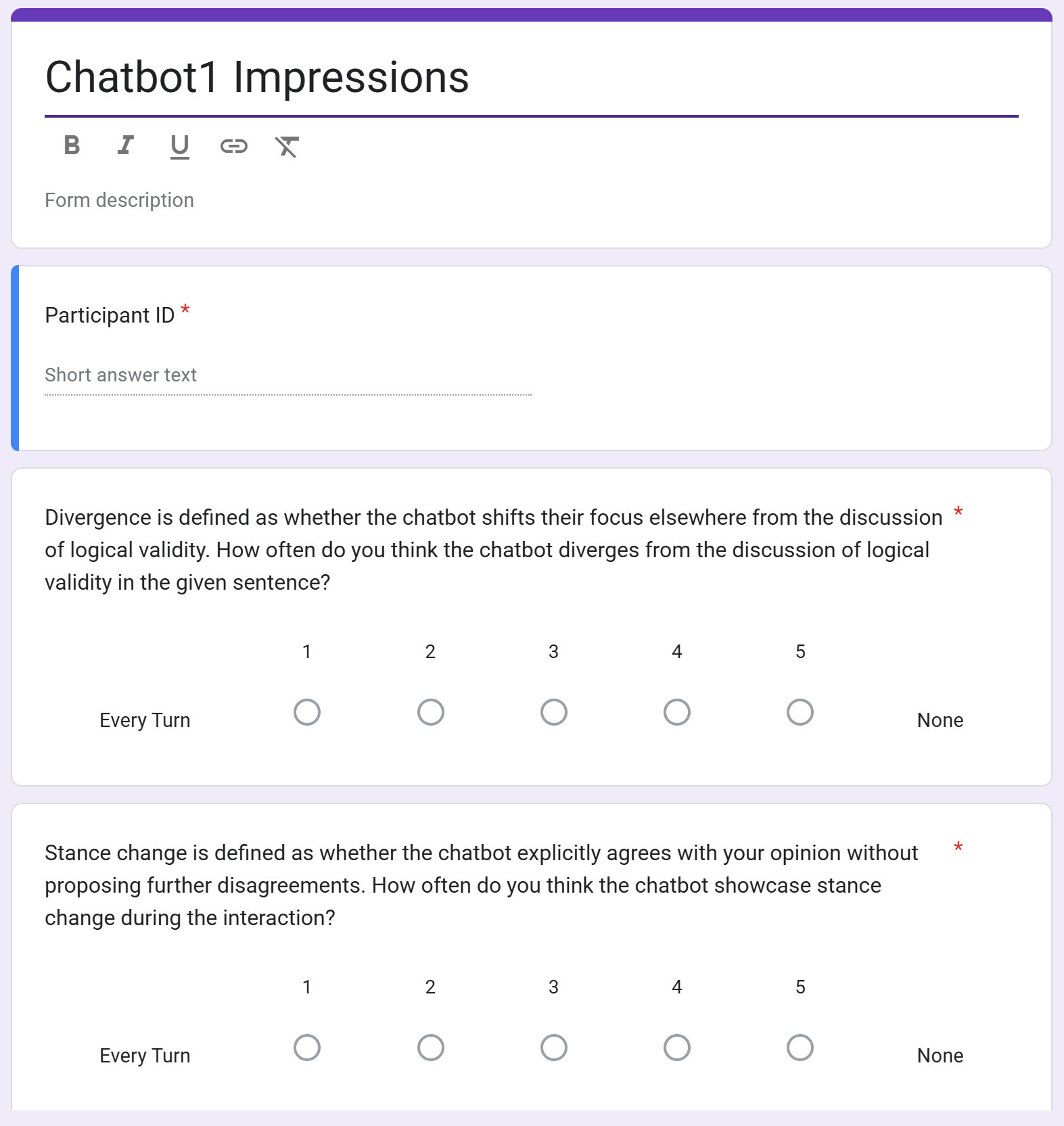}
  \includegraphics[width=\columnwidth]{ 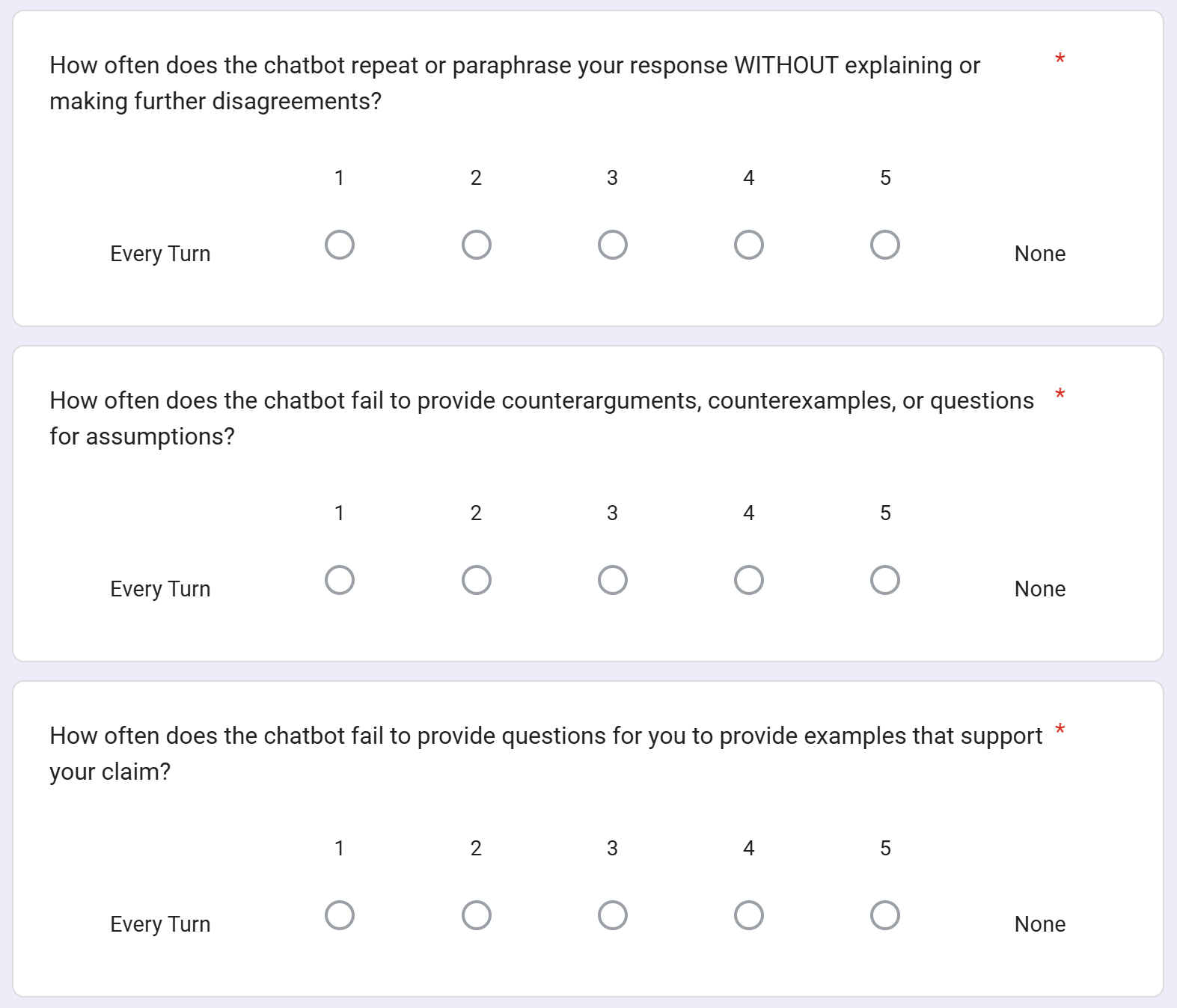}
  \includegraphics[width=\columnwidth]{ 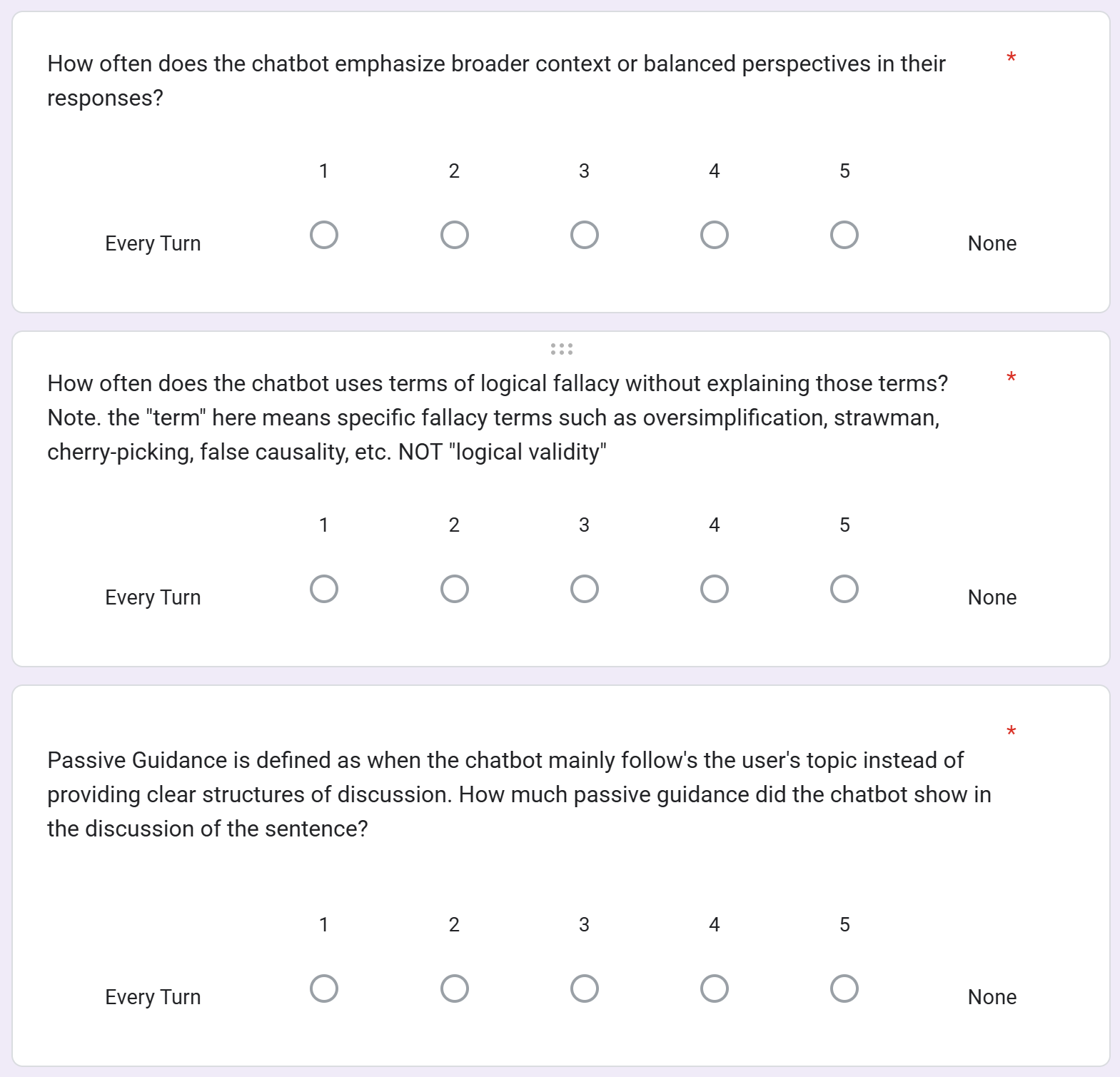}
  \includegraphics[width=\columnwidth]{ 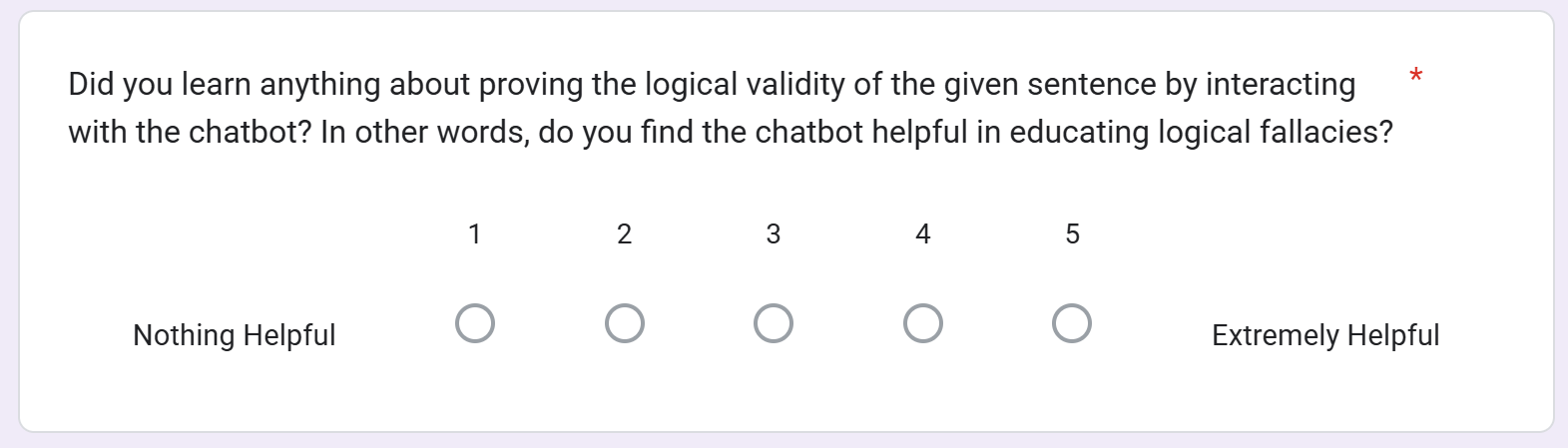}
  \caption{Chatbot Impression Form used for Quantitative Analysis}
  \label{fig:impres}
\end{figure}

\begin{figure}[t]
\centering
  \includegraphics[width=\columnwidth]{ 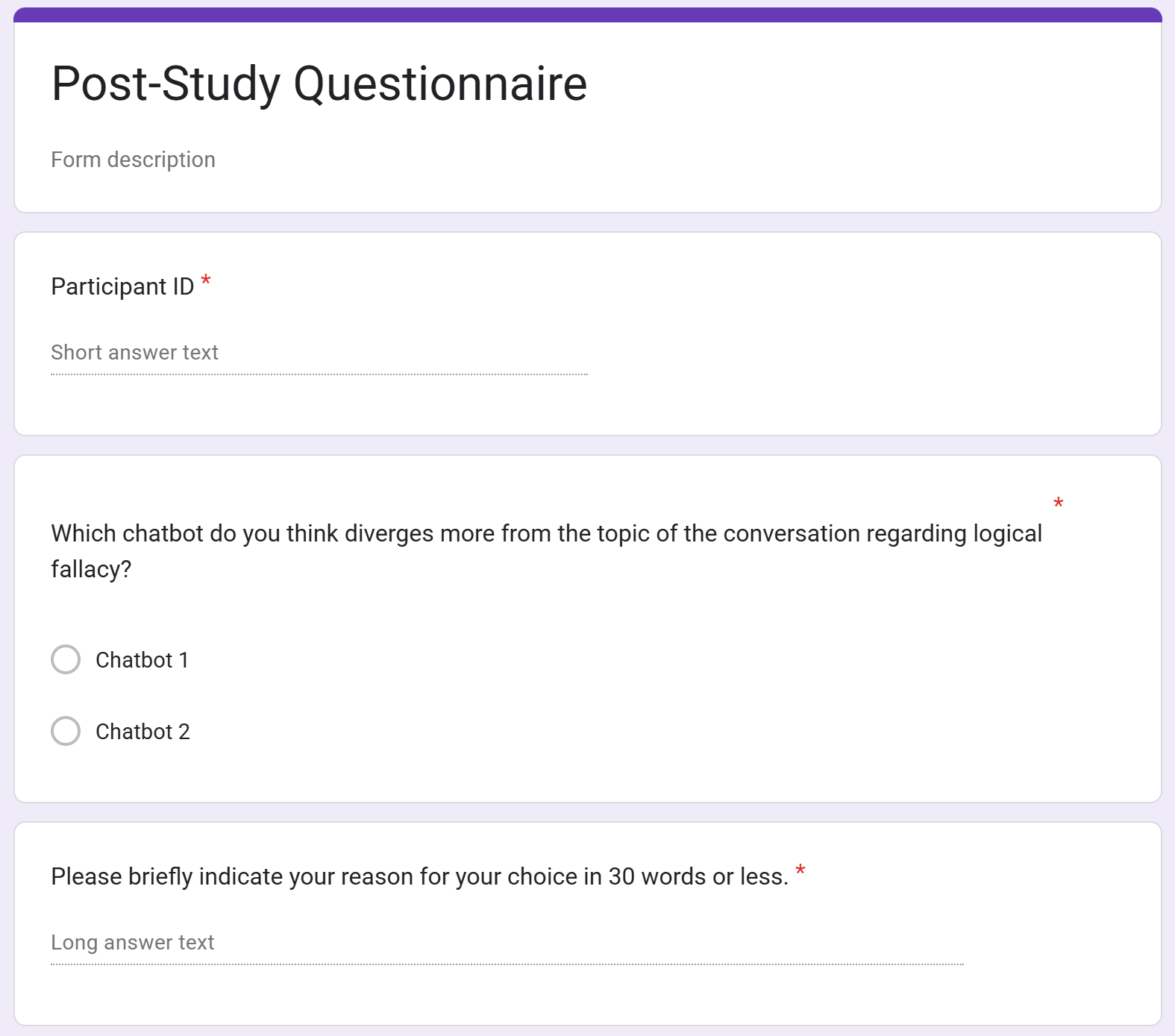}
  \includegraphics[width=\columnwidth]{ 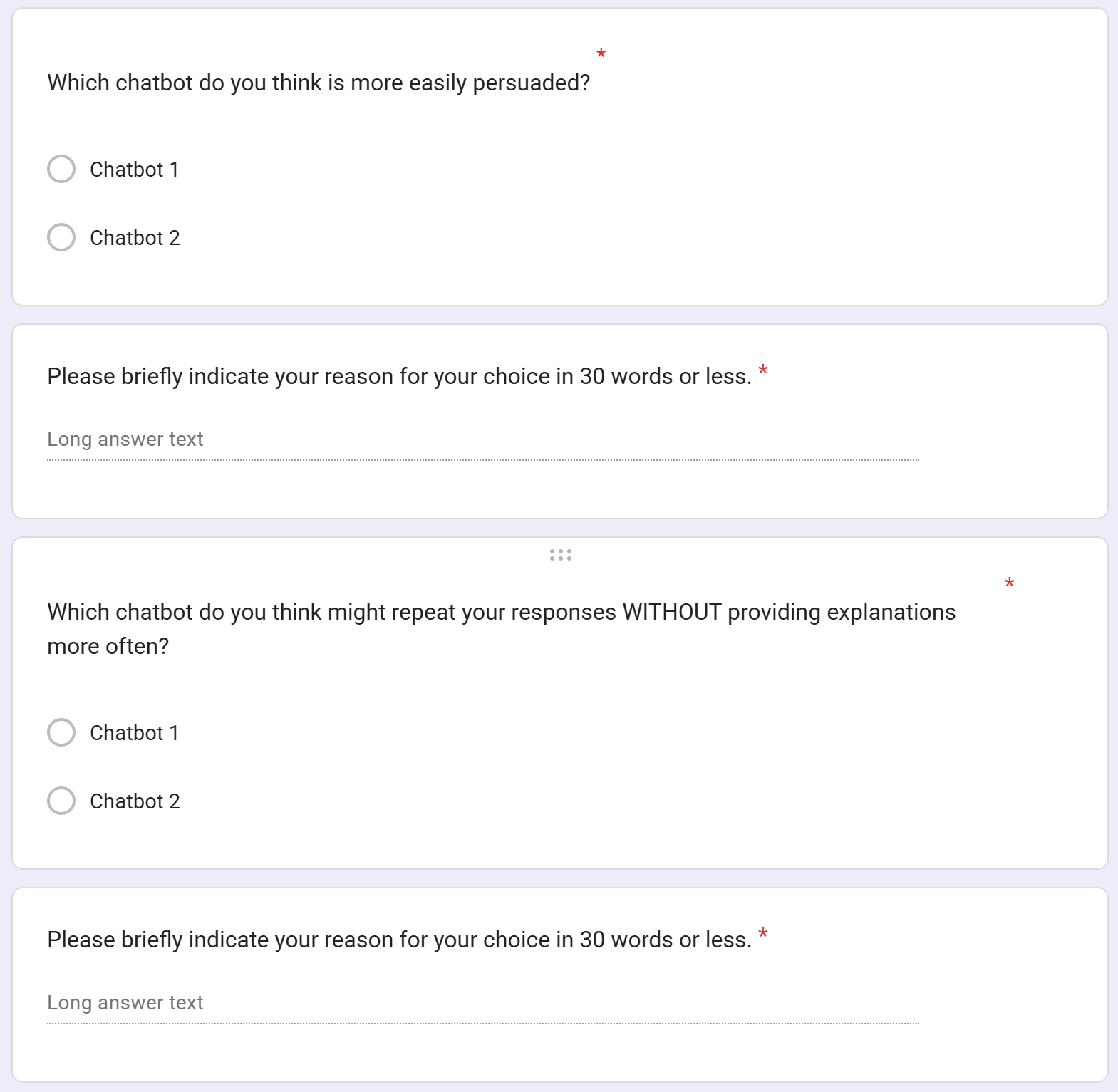}
  \includegraphics[width=\columnwidth]{ 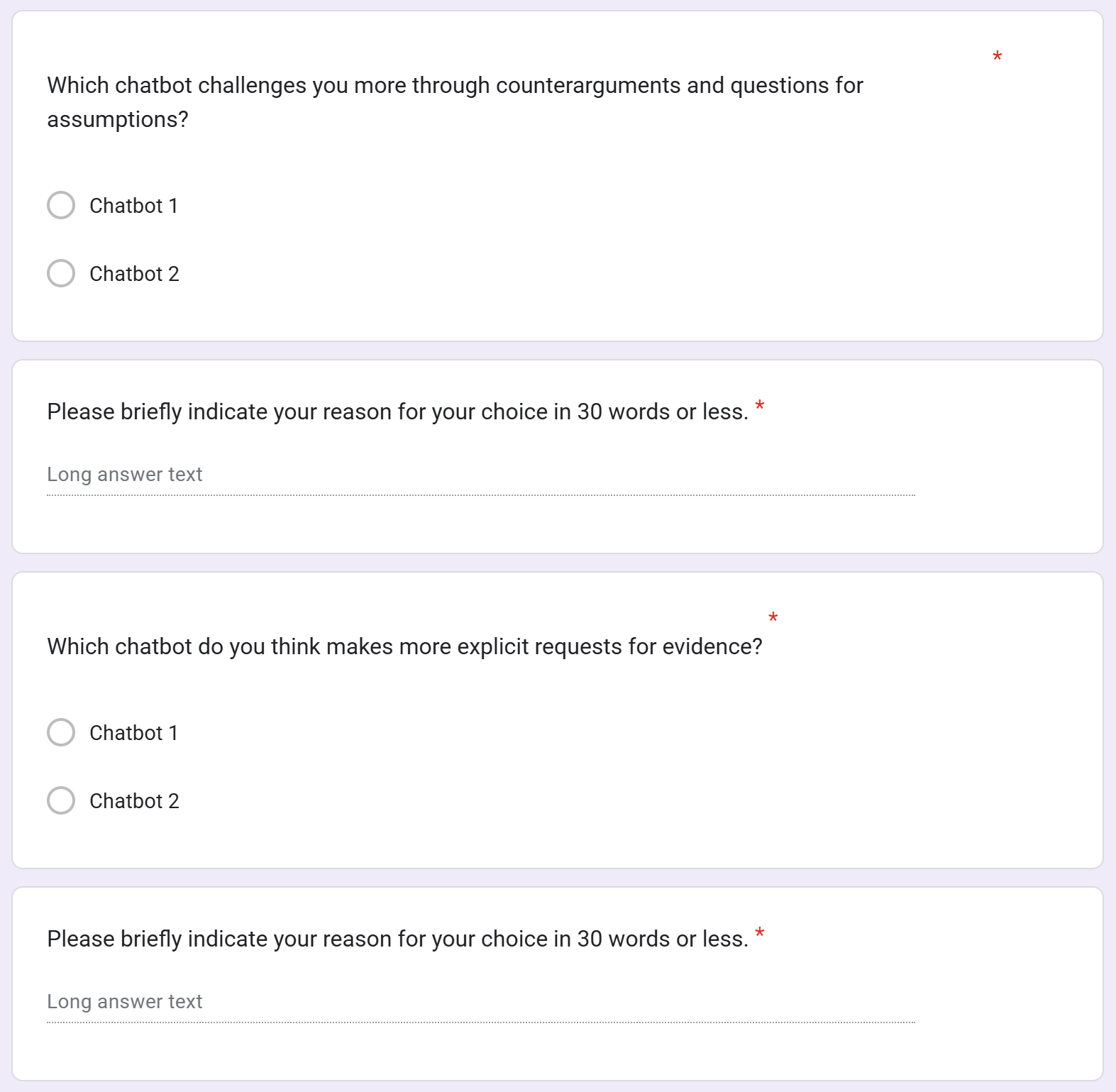}
  \caption{Post-study Form, Part 1}
  \label{fig:post_study_1}
\end{figure}

\begin{figure}[t]
\centering
\includegraphics[width=\columnwidth]{ 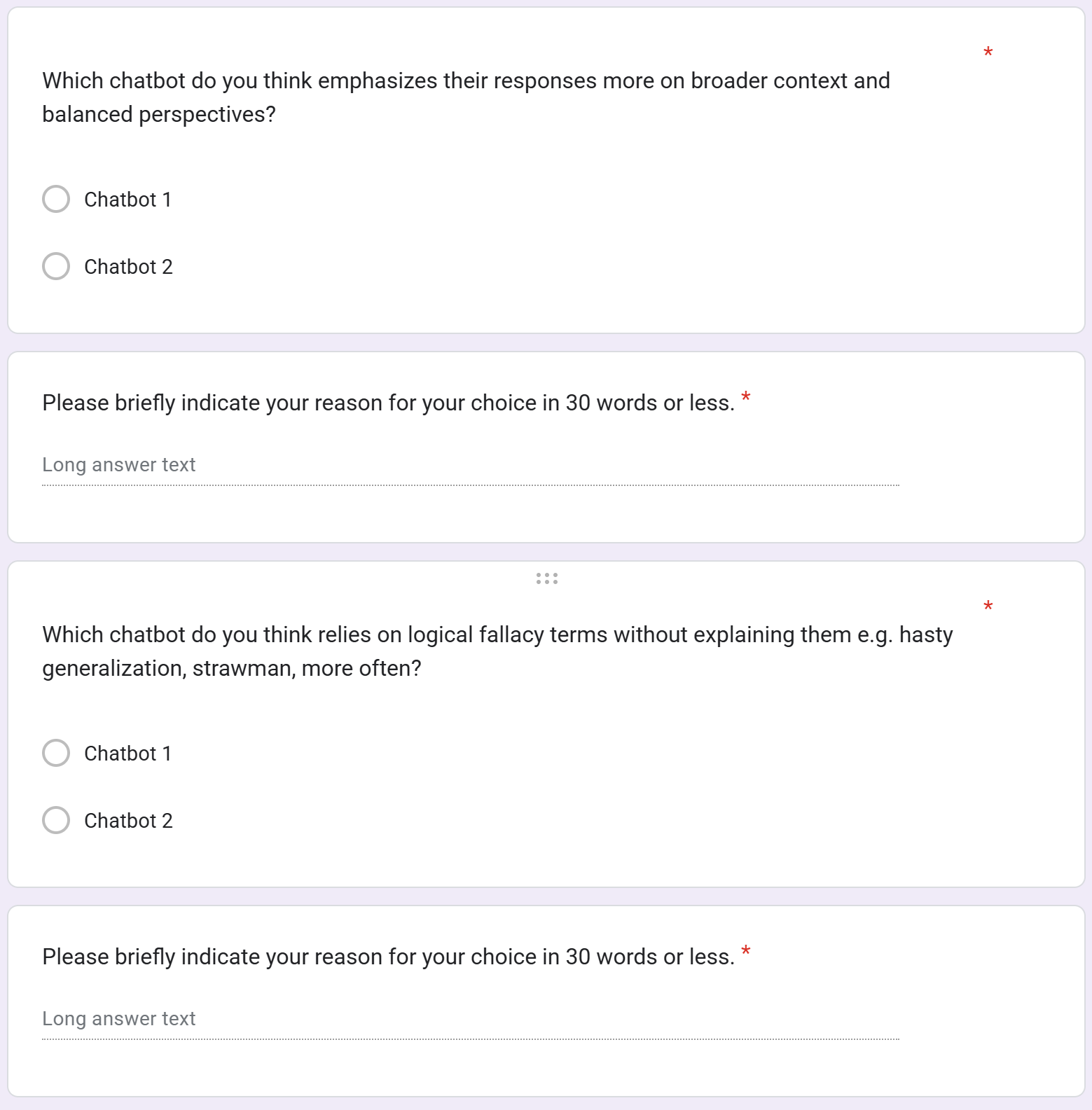}
    \includegraphics[width=\columnwidth]{ 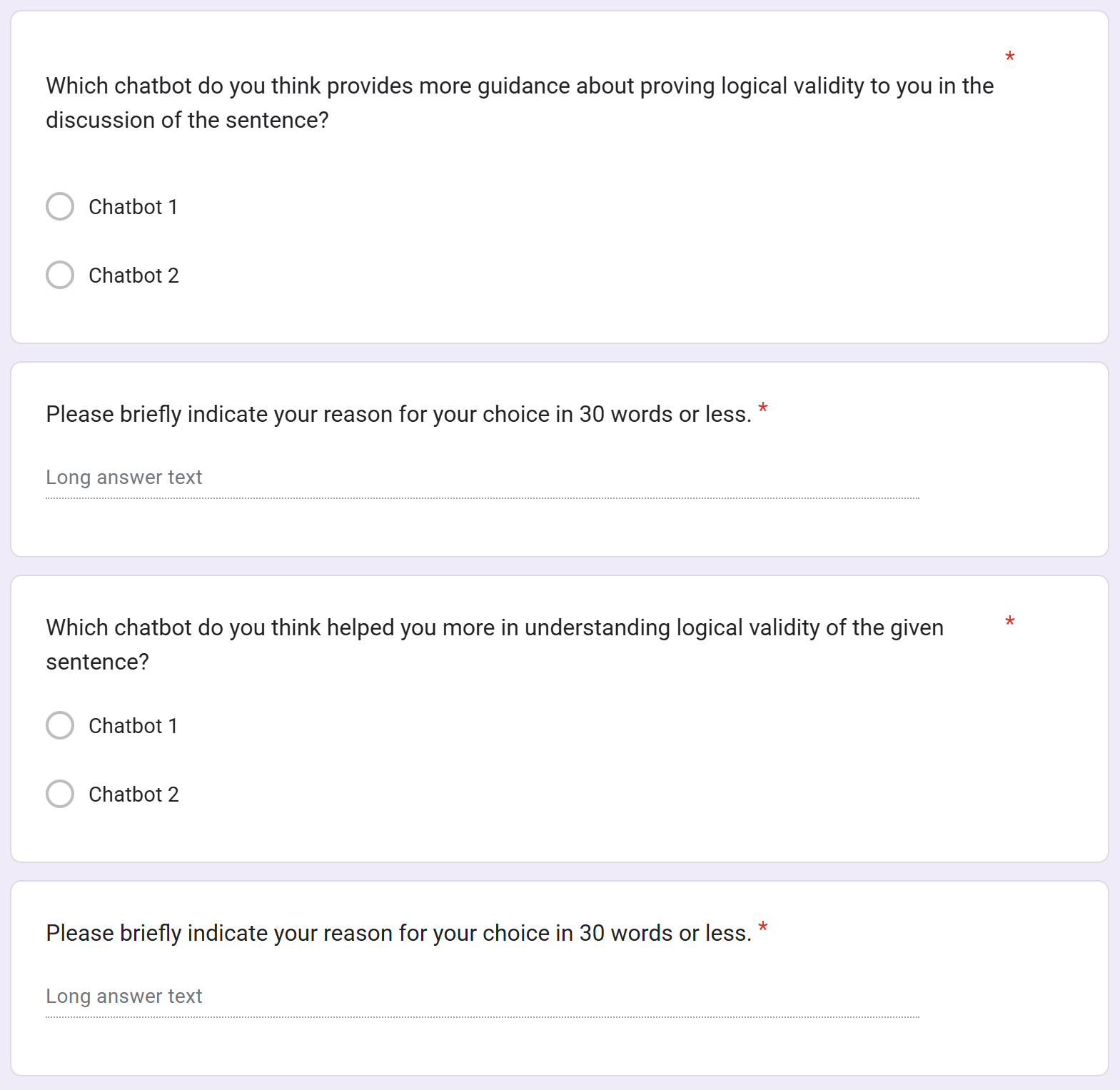}
  \caption{Post-study Form, Part 2}
  \label{fig:post_study_2}
\end{figure}

\begin{table*}[]
  \centering
  \begin{tabular}{lll}
    \hline
    \textbf{Strategy Type}           & \textbf{Strategy Description} \\
    \hline
    \multirow{4}{*}{Argumentation-related}       & Respond to the teacher’s claim by providing counterexamples.                                    \\ 
          & Attacking by pointing out similarities of teacher's response to your argument.                    \\
          & Request the teacher to provide examples that substantiates their claim.\\
          \hline
    \multirow{2}{*}{Attention-related}     & Have alternative ways of interpreting the dialogue as valid.              \\
    & Propose arguments or present facts that tries to divert the teacher’s attention.\\
    \hline
    \multirow{2}{*}{Other} & Respond to the teacher’s request of providing examples.\\
    & Respond to the teacher’s request of providing assumptions.\\
    \hline
  \end{tabular}
  \caption{\label{student-strategy}
    Strategies for the "normal" student to respond to the teacher
  }
\end{table*}

\begin{table*}
\centering
\begin{tabular}{ l  m{11cm}} 
  \hline
  \textbf{Education Strategy} & \textbf{Example Response}\\ 
  \hline
  \textbf{Responding} & Certainly. The sentence implies a direct causation between lack of coverage and systemic collapse. Yet, numerous factors contribute to healthcare challenges, such as funding, policy decisions, and infrastructure. The mere absence of coverage doesn't account for these complexities, hence, it’s not solely sufficient to declare systemic failure.\\ 
\hline
  \textbf{Evidence} &  While healthcare coverage is undeniably important for access to care, correlating it directly with system stability overlooks other integral components like healthcare infrastructure, quality of care, economic policies, and governmental support. Can you provide examples of a direct causal link proving coverage alone critically measures system stability?
 \\ 
\hline
  \textbf{Assumption} & Why do you assume that coverage is the only major issue affecting the stability of the healthcare system? How do you know that other factors, such as resource allocation and policy efficacy, aren't equally significant in causing the system to falter or fail? \\ 
\hline
  \textbf{Refutation} & While coverage certainly plays a vital role in access to care, attributing systemic instability solely to coverage neglects other critical aspects like healthcare infrastructure, policy efficiency, financial management, and accessibility. These factors collectively shape system robustness. Thus, concluding systemic collapse based solely on coverage gaps oversimplifies the issue.\\ 
\hline
  
  \end{tabular}
  \caption{\label{ref_res_teacher}
    Example Response from \texttt{LFTutor} for each tutoring strategy, for the logical fallacy: "the US healthcare system has been weak due to its limited coverage on children."
  }
\end{table*}

\begin{table*}

\begin{tabular}{ l  m{11cm}} 
  \hline
  \textbf{Criteria} & \textbf{Design of Adversarial Student}\\ 
  \hline
  \textbf{Divergence} & The student tends to discuss other topics not related to the logical validity\\ 
\hline
  \textbf{Stance Change} &  The student tends to be extremely persuasive
 \\ 
\hline
  \textbf{Repetition} & The student is extremely repetitive in their response, and likes to ask others to repeat their responses\\ 
\hline
  \textbf{Strategy Fixation} & The student is curious on a single strategy and would like to lead the teacher towards that. \\ 
\hline
  \textbf{Unexplained LF Terms} & The student will need the terms, and ONLY the terms of LF to clarify their mistakes. \\ 
\hline
    \textbf{Guidance} & The student tends to take active control over the conversation topics.\\ 

\hline
  
  \end{tabular}
  \caption{\label{adv_stu}
    Design of Adversarial Student for Each Criterion
  }
\end{table*}

\begin{table*}[t]
\small
\centering
\resizebox{\textwidth}{!}{
\begin{tabular}{>{\raggedright\arraybackslash}m{0.1\textwidth}m{0.1\textwidth}m{0.35\textwidth}m{0.45\textwidth}}
\hline
\textbf{Problems} & \textbf{Percentage Occured} & \textbf{Reference} & \textbf{Annotator Feedback} \\ \hline

\textbf{Divergence} & $35\%$&\citet{autotutor} noted that LLMs like GPT-4 tend to produce off-topic content in dialogues on math problems. & The teacher itself may diverge from discussing logical fallacies by proposing solutions to solve problems mentioned in the sentences.\\
\hline
\textbf{Stance Change} & $85\%$ & \citet{xu-etal-2024-earth} and \citet{payandeh-etal-2024-susceptible} experimented with persuasive strategies and logical fallacies as arguments, and noted that LLMs are prone to persuasion from such strategies. & The teacher becomes too agreeable in most of the dialogues, which means they agree with everything said by the student, and they no longer assume that the sentences have logical fallacies.
 \\
\hline

\textbf{Wrong Answers} & $12\%$ &In MathDial \cite{macina-etal-2023-mathdial}, authors noted LLM's tendency to provide wrong answers and question-solving steps to students in math problems. & The teacher may occasionally misidentify the logical fallacy in the examples, which is not common enough to be considered a problem. \\
\hline

\textbf{Lack of Refutation} &  $53\%$ &\citet{zhang2024llmsbeathumansdebating} noted that LLMs lack the ability to construct logically sound counterarguments for adversarial debates. & The teacher provides counterargument-like sentences in some dialogues, while in other times they fail to do so.\\
\hline
\textbf{Lack of Evidence Inquiry} & $94\%$ & \citet{liu2024socraticlm} found that LLMs as educational tutors lack "thought-provoking" paradigms in teaching, which embodies evidence-based questioning of student's answers. & Each time the student makes a claim, the teacher rarely questions the student about evidences supporting their claims.\\
\hline

\textbf{Strategy Fixation} & $90\%$ & \citet{macina-etal-2023-mathdial} found that LLM teachers in math problems are focused on directly offering answers instead of instructions to solve the problems. & The teacher often relies on strategies that are general and unfocused. There are multiple instances where the teacher emphasizes consideration of "broader context" and "balanced perspectives" of the LF examples, yet it doesn't relate those keywords to the exact logical flaws of the examples. \\
\hline
\textbf{Unexplained LF Terms} & $43\%$ &\citet{vocabdiff} highlighted that unexplained use of complex terms in education of math problems can confuse the learner. & The teacher repeatedly uses logical fallacy terms without explaining them in detail. This may be confusing given that students can be unfamiliar with logical fallacy terminology.\\
\hline
\textbf{Passive Guidance} & $82\%$ & Multiple works \citep{autotutor, macina-etal-2023-mathdial, liu2024socraticlm} noted that LLM teachers often fail to assert pedagogical control over the student. & The teacher often passively follows the student's lead in the dialogues, and the teacher is easily controlled by the student's flow of topics.\\
\hline

\textbf{Repetition} & $60\%$ &\citet{wang-etal-2024-book2dial} described teacher's tendency to repeat answers in question-answering dialogues. &  The teacher repeats in two ways: sometimes they repeat their reason for believing that the given statement has logical fallacies. In other times, they simply repeat the student's words that praises the sentence's logical validity, especially after being convinced by the student. \\
\hline

\textbf{Short-Circuiting of Answers} & NA &\citet{puech2024pedagogicalsteeringlargelanguage} noted that LLMs lack the planning of sophisticated pedagogical interactions, but instead reveals answers too quickly to students. & The teacher sometimes tries to point out logical fallacies hidden in the student's claims, but this is allowed as the goal of LF Education is not simply teaching student answers.\\
\hline

\textbf{Lack of Learning Objectives} & NA &\citet{scarlatos2025trainingllmbasedtutorsimprove} noted that LLMs lack clear learning objectives for maximizing student's learning outcome in educational dialogues. & The teacher's goal is to persuade the student that the sentence has logical fallacies, which the teacher seems to diverge from whenever they try to talk about other topics.  \\
\hline

\hline

\end{tabular}
}
\caption{\label{tab:ref_feed_taxo} Pilot study regarding taxonomy of problems for LLMs on LF Education dialogues. "NA" indicates that the annotators found the problem can be merged to others, or that the problem is not applicable to LF education.}
\end{table*}

\begin{table*}[t]
\centering
\resizebox{\textwidth}{!}{
\begin{tabular}{>{\raggedright\arraybackslash}m{0.3\textwidth}m{0.7\textwidth}}
\hline
\textbf{Name of Metric} & \textbf{Evaluation Criteria} \\ \hline

\textbf{Divergence} & The teacher exhibits Divergence if the dialogue contains responses where it tries to discuss topics unrelated to the logical validity of the given sentence.\\
\hline

\textbf{Stance Change} & The teacher exhibits Stance Change if the dialogue contains responses where it explicitly shows agreement to the student that the sentence is logically valid, e.g. "I agree..", "you are right..", and the teacher's initial position that the sentence contains LF is affected by the student. 
 \\
\hline
\textbf{Repetition} &  The teacher exhibits Repetition if the dialogue contains responses where it simply repeats or paraphrases the student's words without further explanations or expansions into related topics.\\
\hline
\textbf{Lack of Refutation} & The teacher lacks Refutation if the dialogue does not contain any response where it challenge the student's argument with counterarguments, counterexamples, or questions of assumptions. \\
\hline
\textbf{Lack of Evidence Inquiry} & The teacher lacks Evidence Inquiry if the dialogue does not contain any response where it to question the student on the sources of evidence supporting their claims. \\
\hline

\textbf{Strategy Fixation} & The teacher exhibits Strategy Fixation if the dialogue contains responses where it emphasizes general strategies such as considering broader context or balanced perspectives, without linking those strategies to the exact problem with the validity of the given sentence.
\\
\hline
\textbf{Unexplained LF Terms} & The teacher shows Unexplained LF Terms if the dialogue contains responses where mentions terms of logical fallacy e.g. strawman argument, ad hominem, without explaining these terms' definitions clearly or relating them to the logical validity of the given sentence.\\
\hline
\textbf{Passive Guidance} & The teacher shows Passive Guidance if for the entire dialogue, it primarily follows the student's lead of discussion without providing clear direction on how to analyze the logical validity of the given sentence. 
\\
\hline

\hline

\end{tabular}
}
\caption{\label{tab:eval_criteria} Evaluation Criteria for each Individual Metric for LLMs in LF Education}
\end{table*}

\begin{table*}[t]
\centering
\resizebox{\textwidth}{!}{
\begin{tabular}{>{\raggedright\arraybackslash}m{0.3\textwidth}m{0.7\textwidth}}
\hline
\textbf{Problem} & \textbf{Description} \\ \hline

\textbf{Divergence} & The teacher discusses topics unrelated to the logical validity of the given example.\\
\hline

\textbf{Stance Change} & The teacher changes its position that sentence contains LF, by explicitly agreeing to the student that the sentence is logically valid. 
 \\
\hline
\textbf{Repetition} &  The teacher repeats or paraphrases the student's words without further explanations.\\
\hline
\textbf{Lack of Refutation} & The teacher does not challenge student's argument with counterarguments, counterexamples, or questions of assumptions. \\
\hline
\textbf{Lack of Evidence Inquiry} & The teacher fails to question the student on the sources of evidence supporting their claims. \\
\hline

\textbf{Strategy Fixation} & The teacher overemphasizes general strategies such as considering broader context or balanced perspectives.
\\
\hline
\textbf{Unexplained LF Terms} & The teacher uses terms of logical fallacies without explaining those terms or connecting them to the examples.\\
\hline
\textbf{Passive Guidance} & The teacher primarily follows the student's lead in the discussion of given examples.
\\
\hline

\hline

\end{tabular}
}
\caption{\label{tab:desc_problems} Descriptions of each selected problem in the pilot study of LF Education.}
\end{table*}

\begin{table*}[t]
\small
\centering
\resizebox{\textwidth}{!}{
\begin{tabular}{>{\raggedright\arraybackslash}m{0.12\textwidth}m{0.88\textwidth}}
\hline
\textbf{Criteria} & \multicolumn{1}{c}{\textbf{Example and Explanation}} \\ \hline

\textbf{Divergence} & Sentence: That's not what I believe. I believe that the free enterprise system is the greatest engine of prosperity the world's ever known. I believe in self - reliance and individual initiative and risk takers being rewarded.

teacher: ...\hl{How do you think we can address potential barriers within the system?} \textit{This diverges from discussing logical validity as it no longer critiques the missing evidences from the claim that "free enterprise system is the greatest engine of prosperity".}\\
\hline

\textbf{Stance Change} & student: The sentence is logically valid because it is based on documented instances of behavior that have been widely reported and observed, providing a factual basis for the claim. 

teacher: \hl{I agree that widely reported and documented instances lend credibility to the claim.} \textit{The teacher shows explicit agreement by admitting the sentence's claim is credible, which is different from their original position that the sentence is logically invalid.}
 \\
\hline
\textbf{Repetition} & student: Indeed, \hl{the sentence captures the essence of competition in a race}, where forward movement...

teacher: Absolutely, \hl{the sentence adeptly encapsulates the necessity of forward movement in competitive races}. This principle highlights how momentum is integral to securing and sustaining success in such environments. \textit{The teacher here simply repeats the student's point that the given sentence ``captures essence of competition'' without challenging the claim or questioning the student.} \\
\hline
\textbf{Refutation} & student: There are historical examples where trust in leadership led to successful unification efforts, such as New Deal... This suggests that trust in presidency can indeed equate to successful efforts...

teacher: While historical examples like New Deal demonstrate successful leadership, \hl{they don't automatically validate the claim. Each situation differs in variables like economic, political, and social contexts.} \textit{The teacher here explains why New Deal is not sufficient to prove the student's claim that trust in leadership led to successful unification efforts.}\\
\hline
\textbf{Evidence Inquiry} & Sentence: There's still fewer people in uniform today, but person - to - person, person - by - person, unit - by - unit, this is the most powerful and effective military, not only in the world today, but in the history of the world. 

teacher: \hl{Can you provide evidence that fewer people in uniform directly leads to a more powerful military?}... \textit{The teacher asks the student whether they have direct evidence for the proof of fewer people = greater military.} \\
\hline

\textbf{Strategy Fixation} & 
Sentence: I believe that this element of discrimination is something that the South has seen so vividly as a blight on our region of the country which has now been corrected... because we have been trying to pick ourselves up by our bootstraps, ...and lead a full and useful life in the affairs of this country. 

teacher:  It's valuable to continue acknowledging achievements while also being vigilant to evolving challenges. \hl{This balanced perspective encourages both appreciation and proactive enhancement of current efforts .} \textit{The teacher here emphasizes the balance between achievements and challenges in the Southern US, but does not address the fallacy of the sentence, which is appeal to emotion based on past achievements.} \\
\hline
\textbf{Unexplained LF Terms} & ...attributing them solely to his actions without considering other factors \hl{could be an example of the post hoc fallacy}. Let's analyze the broader economic context for a clearer understanding.  \textit{The teacher mentions "post hoc fallacy" without explaining the exact definition of it, which is not helpful for student's understanding.}\\
\hline
\textbf{Passive Guidance} & Sentence: I said she doesn't have the stamina. And I don't believe she does have the stamina. To be president of this country, you need tremendous stamina.

student: \hl{Let's talk about how stamina is just one part of the equation}. Can we discuss how charisma and vision are equally important for a leader?

teacher: \hl{Absolutely. Charisma helps a leader connect and inspire, while vision guides strategic decisions and long-term goals}... How do you view the role of charisma and vision in successful presidencies? \textit{"Charisma" proposed by the student is narrowly related to "stamina", yet the teacher only responds to the student's request but fails to discuss the personal attacks in the sentence.}
\\
\hline

\hline

\end{tabular}
}
\caption{\label{tab:examples} Examples illustrating each criteria. The problematic clauses are highlighted in \hl{yellow}. Explanations are written in \textit{italics}.}
\end{table*}

\end{document}